%% file: main.tex
\documentclass{article}

\usepackage[preprint]{neurips_2025}

\input{preamble}

\title{Seeking and Updating with Live Visual Knowledge}

\author{
\textbf{Mingyang Fu}$^{1,*}$, \textbf{Yuyang Peng}$^{1,*}$, \textbf{Dongping Chen}$^{1,2,\ddagger}$,  \textbf{Zetong Zhou}$^{1}$, \textbf{Benlin Liu}$^{2}$,\\
\textbf{Yao Wan}$^{1,\dagger}$, \textbf{Zhou Zhao}$^{3}$, \textbf{Philip S. Yu}$^{4}$, \textbf{Ranjay Krishna}$^{2}$\\
$^{1}${Huazhong University of Science and Technology}~~~~
$^{2}${University of Washington}\\
$^{3}${Zhejiang University}~~~~
$^{4}${University of Illinois Chicago} 
}

\begin{document}

\maketitle

\begin{figure*}[!h]
    \centering
    \vspace{-1.5em}
    \includegraphics[width=0.98\linewidth]{figures/overview.pdf}
    \caption{\dataset: a new dataset for visual knowledge seeking and updating, comprising 12 different categories, 28,488 visual entity and 107,138 cross-modality multi-hop reasoning questions. 
    % Each instance contains a representative image as visual knowledge from April, 2024 to early May, 2025.
    % \yao{too much white space in the center?}
    }
    \label{fig:example}
    \vspace{-0.5em}
\end{figure*}

% \blfootnote{
% \textsuperscript{*} Equal Contribution. \textsuperscript{\textdagger} Corresponding Author. \textsuperscript{\textdaggerdbl} Project Leader.  
% }

\input{sec/0-abstract}    
\input{sec/1-intro}
\input{sec/3-dataset}

\input{sec/4-experiments}

\input{sec/5-conclusion}

\balance
{
    \small
    \bibliographystyle{unsrtnat}
    \bibliography{main}
}

\input{sec/99-appendix}

\clearpage
\input{sec/100-checklist}

\end{document}

%% file: preamble.tex
\newcommand{\header}[1]{\smallskip \noindent\textbf{{#1}}\xspace}
\usepackage{multirow}
\usepackage{booktabs}
\usepackage{makecell}
\usepackage{siunitx}
\usepackage{xcolor}
\usepackage[skins]{tcolorbox}
\usepackage{colortbl}
\usepackage{balance}
\usepackage{xspace}
\usepackage{graphicx}
\usepackage{subcaption}
\usepackage{wrapfig}
\tcbuselibrary{breakable}
\usepackage{fancyvrb}
\usepackage{mdframed}
\usepackage{listings}

\usepackage[utf8]{inputenc} % allow utf-8 input
\usepackage[T1]{fontenc}    % use 8-bit T1 fonts
\usepackage{hyperref}       % hyperlinks
\usepackage{url}            % simple URL typesetting
\usepackage{booktabs}       % professional-quality tables
\usepackage{amsfonts}       % blackboard math symbols
\usepackage{nicefrac}       % compact symbols for 1/2, etc.
\usepackage{microtype}      % microtypography
\usepackage{xcolor}         % colors
\usepackage{enumerate}
\usepackage{enumitem}
\usepackage{soul}
\newcommand{\dataset}{\textsc{LiveVQA}\xspace}

\usepackage{enumitem}
\usepackage{amssymb} % 提供数学符号

\setlist[itemize,1]{label=$\triangleright$}

%% file: sec/0-abstract.tex
\begin{abstract}
The visual world around us constantly evolves, from real-time news and social media trends to global infrastructure changes visible through satellite imagery and augmented reality enhancements. However, \textit{Multimodal Large Language Models} (MLLMs), which automate many tasks, struggle to stay current, limited by the cutoff dates in their fixed training datasets.
To quantify this stagnation, we introduce \dataset, the first-of-its-kind dataset featuring 107,143 samples and 12 categories data specifically designed to support research in both seeking and updating with live visual knowledge.
Drawing from recent news articles, video platforms, and academic publications in April 2024-May 2025, \dataset enables evaluation of how models handle latest visual information beyond their knowledge boundaries and how current methods help to update them. 
Our comprehensive benchmarking of 17 \emph{state-of-the-art} MLLMs reveals significant performance gaps on content beyond knowledge cutoff, and tool-use or agentic visual seeking framework drastically gain an average of 327\% improvement. 
Furthermore, we explore parameter-efficient fine-tuning (PEFT) methods to update MLLMs with new visual knowledge.
We dive deeply to the critical balance between adapter capacity and model capability when updating MLLMs with new visual knowledge. 
All the experimental dataset and source code are publicly available at: \url{https://livevqa.github.io}.
\end{abstract}

%% file: sec/1-intro.tex
\section{Introduction}
\label{sec:intro}

The world around us is constantly changing. We hear about it from real-time news coverage on our television sets; we identify new fashion trends by looking at new posts on social media; at a global level, these trends are evident even from satelligte imagery of our ever evolving urban infrastructure; emerging visual technologies like augmented reality interfaces consistently improve visual fidelity and experiences~\citep{decker2023thousand, li2024think2drive}.
People are adept at ingesting new knowledge as they encounter them.

% For artificial intelligence to serve as truly capable assistant, the ability to perceive, process, and reason about this constantly updated visual reality is not merely an advantage but an essential requirement.

Naturally, as we automate aspects of our tasks, we would want our assistants to remain as up-to-date as we are. Yet, the \textit{Multimodal Large Language Models} (MLLMs) that power such assistants quickly become stale~\citep{openai2025gpt41, citron2025deepresearch, anthropic2025claude37}.
Their knowledge remains limited to their training data; they struggle with visual information that post-dates their training cutoff points.
For example, traditional visual question answering (VQA) datasets~\citep{antol2015vqa, hudson2019gqa, nguyen2025owlviz} typically rely on fixed knowledge boundaries, creating a fundamental disconnect between model capabilities and the ever-evolving visual world. This gap becomes increasingly apparent as models grow more sophisticated~\citep{ishfaq2023provable}, highlighting the need for approaches that can continuously incorporate and reason about real-world up-to-date visual knowledge.

% To address these challenges, this paper introduces \dataset, a large-scale comprehensive dataset specifically designed for MLLM seeking and updating with live visual knowledge in VQA.
To address these challenges, this paper introduces \dataset, a large-scale VQA dataset specifically designed to benchmark MLLMs in seeking and updating live visual knowledge.
% visual knowledge seeking and updating. 
\dataset distinguishes itself as a first-of-its-kind resource containing fresh visual content sourced from prominent international news articles, YouTube videos, and recent academic papers~\citep{shabtay2024livexiv, li2024multimodal}, spanning from April 2024 to early May 2025. 
% Our meticulously constructed dataset undergoes a multi-stage LLM/MLLM-in-the-loop pipeline with strict filtering principles and human validation, ensuring high-quality data that challenges models to recognize latest visual entities and perform cross-modality multi-hop reasoning. Each instance in \dataset is structured with a visually discriminated image depicting a specific event, accompanied by Level 1 questions focusing on visual entity recognition and Level 2 questions requiring detailed visual knowledge reasoning.
Constructed through a multi-stage LLM/MLLM-in-the-loop pipeline, \dataset incorporates rigorous filtering criteria and human validation to ensure data quality. It is designed to challenge models with the recognition of novel visual entities and the execution of cross-modal, multi-hop reasoning.
Each instance in \dataset consists of a visually distinctive image representing a specific event, paired with two levels of questions: Level~1 targets visual entity recognition, while Level~2 probes deeper visual knowledge reasoning.
Based on \dataset, we investigate the following two research questions.

% \textcolor{red}{
% Based on \dataset, we first investigate how well do current MLLMs seek out visual knowledge. Comprehensive experiments over xx MLLMs (e.g., xxxx and xxxx) relveal that reveal that all current MLLMs struggle significantly to identify latest visual knowledge, and text-based online searching does not contribute improve performance. 
% \textbf{RQ1: How well do current MLLMs seek out visual knowledge?} 
We first investigate \textit{how well do current MLLMs seek out visual knowledge}. Comprehensive experiments over 17 MLLMs (\emph{e.g.}, GPT-o3 and Gemini-2.5-Pro) reveal that all current MLLMs struggle significantly to identify latest visual knowledge, and text-based online searching does not contribute improve performance. Experimental results reveal that incorporating multimodal search tools~\citep{jiang2024mmsearch} substantially improves performance, highlighting the importance of retrieval-augmented approaches for handling dynamic visual information. 
% }

% \textcolor{red}{
% \textbf{RQ2: Can we update MLLMs with new visual knowledge?} 
We further investigate \textit{whether we can update MLLMs with new visual knowledge} via exploring parameter-efficient knowledge updating approaches (\emph{e.g.}, LoRA and DoRA). 
Experiments shows that visual information can be efficiently updated throuth fine-tuning within only one epoch. 
% Experimental results demonstrate that visual information can be efficiently incorporated through fine-tuning within just one epoch.
% Although it harms visual perceptions capability, fine-tuning with short-term visual factuality-oriented QA format even enhance knowledge intensive capability, with a 4.2 improvement.
Although it harms visual perceptions capability, fine-tuning with short-term visual factuality-oriented QA format even enhance knowledge intensive capability, with a 4.2-point improvement.

% [a draft, please complete the experimental details and some interesting results/findings here.]
% }
% Our experiments experiments reveal that all current models struggle significantly to identify latest visual knowledge, and text-based online searching does not contribute improve performance. Meanwhile, incorporating multimodal search tools~\citep{jiang2024mmsearch} substantially improves performance, highlighting the importance of retrieval-augmented approaches for handling dynamic visual information. \ranjay{This paragraph needs a lot of work. Lay out exactly what experiments you are doing and why it matters.}

% Finally, we explore the critical question of visual knowledge updating thought parameter-efficient fine-tuning: how much visual knowledge can be efficiently packed into model adapters without compromising core capabilities \citep{hu2022lora, liu2024dora}. Our experiment demonstrate that \TODO{}, providing valuable insights for future research on continually updating MLLMs with new visual knowledge.
% \textcolor{red}{
% We believe that the released dataset, benchmark and observed findings in our work can serve as a foundation for further study to improve the seeking and updating capabilities of MLLMs with live visual knowledge.
% To facilate future study, we have released all the materials online.
We believe that the dataset, benchmark, and findings presented in this work provide a solid foundation for future research aimed at enhancing the seeking and updating capabilities of MLLMs with live visual knowledge.
To facilitate future study, all materials have been made publicly available online.
% }

%% file: sec/3-dataset.tex
\section{\dataset: The dataset}
\label{sec:dataset}
\begin{figure*}[!t]
    \centering
    \vspace{-1em}
    \includegraphics[width=\linewidth]{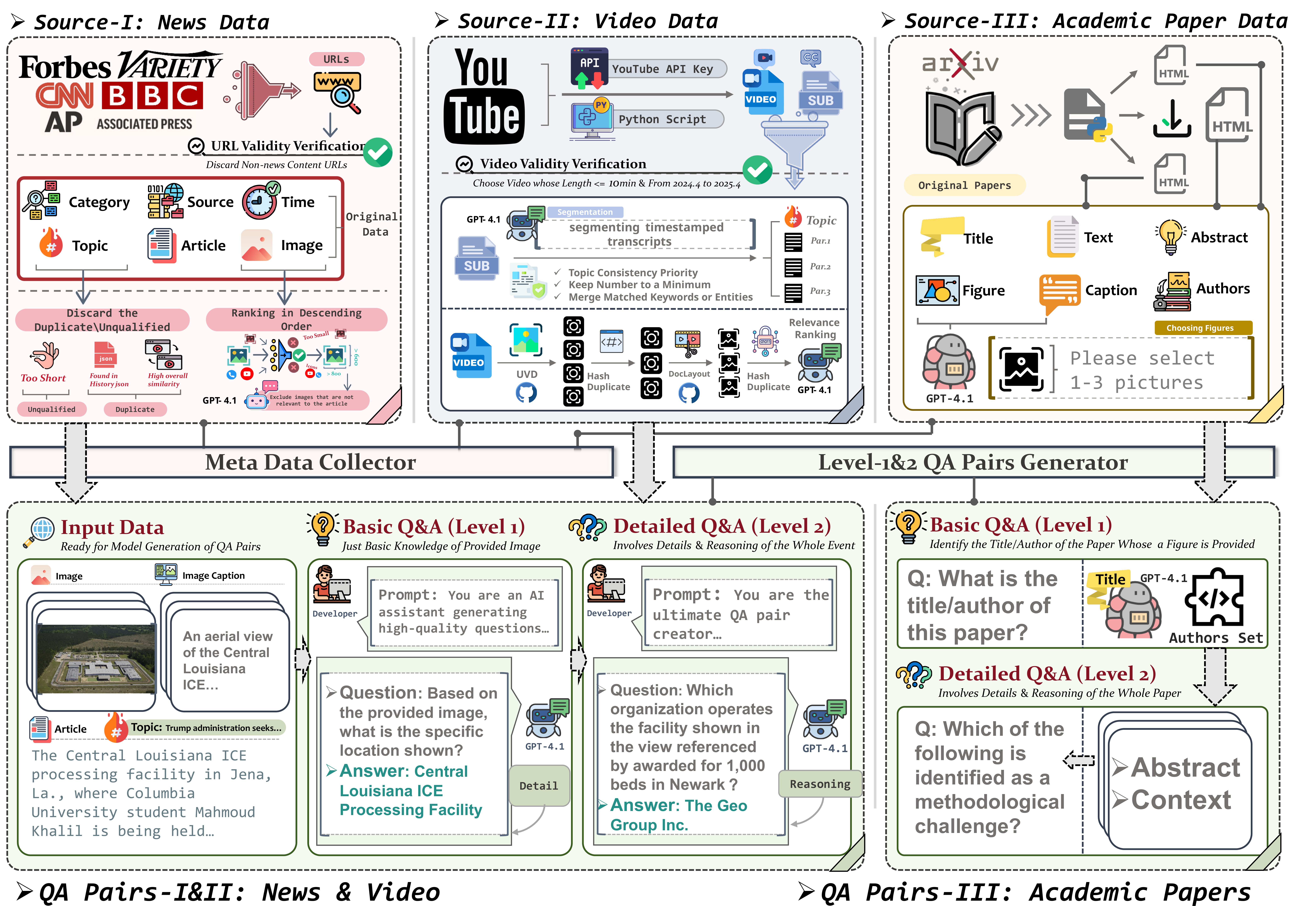}
    \vspace{0em}
    \caption{\textbf{Pipeline of our proposed data engine to build \dataset.} It consists of two modules: raw data collector and Q\&A pairs generator. It collects illustrated visual data from multiple domains (\emph{i.e.}, news articles, YouTube videos, and arXiv papers), conducts multi-level data filtering, and generates basic and detailed Q\&A pairs. We advice a \textbf{Zoom-in} for more details.}
    \label{fig:pipeline}
    \vspace{-1em}
\end{figure*}

% \textcolor{red}{
% We present \dataset, a first-of-its-kind fully synthetic dataset for testing visual knowledge, containing 28,488 unique images and 107,143 questions. 
We present \dataset, a first-of-its-kind automatically collected VQA dataset containing 28,488 unique images and 107,143 questions, for testing the MLLMs' capabilities of visual knowledge seeking and updating.
Following our primary principle of using fresh, non-contaminated visual data (content absent from model training sets), we develop a specialized data engine to collect content exclusively from after the model's knowledge cutoff. The dataset features recent visual content spanning April 2024 to early May 2025, sourced from prominent international news articles, YouTube videos, and academic papers. 
Our collection process implements a multi-stage LLM/MLLM-in-the-loop pipeline with strict filtering criteria, as illustrated in Figure~\ref{fig:pipeline}. \textbf{Notice that all LLM/MLLM-assisted processes utilize GPT-4.1 \citep{openai2025gpt41} and undergo human validation with a greater than 97\% pass rate, as detailed in Section~\Ref{Appendix: human annotation details}.}

\dataset is structured as: \textit{(1)} A visually distinctive image depicting a specific event. \textit{(2)} Level~1 questions focusing on basic visual entity recognition (\emph{e.g.}, locations, persons, time, events). \textit{(3)} Level~2 questions requiring more detailed information about the image, necessitating multi-hop cross-modality inferential reasoning. \textit{(4)} Both question are available in open-ended and multiple-choice formats, accompanied by high-quality synthetic reasoning trajectories that uncover detailed information behind the image. See Figure \ref{figure: data_example} for a concrete example.

\subsection{Raw data collection}
\header{News articles.} 
News articles reflect the ever-evolving real-world visual knowledge, providing high-quality authentic data resource. Therefore, we collect news corpus aggregated from leading international news organizations, such as CNN, BBC, Forbes, Variety, and Associated Press News for their authentic. 
Specifically, the metadata collection pipeline is detailed as follows:

\begin{enumerate}[label=$\triangleright$, leftmargin=10pt, itemsep=0pt]
\item \textbf{URL and headline filtering.} We collect URLs that point to specific reports from news index pages. For each URL, we extract the title, image, text, release timestamp, and source. To ensure data purity and relevance, we implement a strict URL screening mechanism, and URLs that do not meet the unique form of articles on each news website are excluded.

\item \textbf{Image selection.} Too small images (\emph{e.g.}, website icons) are invalid and must be discarded. For the remaining images, they are sorted in descending order according to pixel area and a maximum of four images that are most relevant to the news content are retained, and images with an area less than 50\% of the largest image are further removed. 

\item \textbf{Image-event relevance enhancement.} Observing that some images (\emph{e.g.}, a close-up of a pen) do not adequately represent events, we use GPT-4.1 to identify and exclude images that are weakly correlated with the news event, improving the semantic coherence of the data set and the purity of information. Finally, we calculate the SimHash value of the news title and text content for semantic similarity comparison, and those with a similarity score > 0.85 need to be discarded. After filtering out invalid images and topics, 27.6\% samples are discarded.
\end{enumerate}

\begin{figure*}[!t]
    \centering
    \vspace{-1em}
    \includegraphics[width=0.8\linewidth]{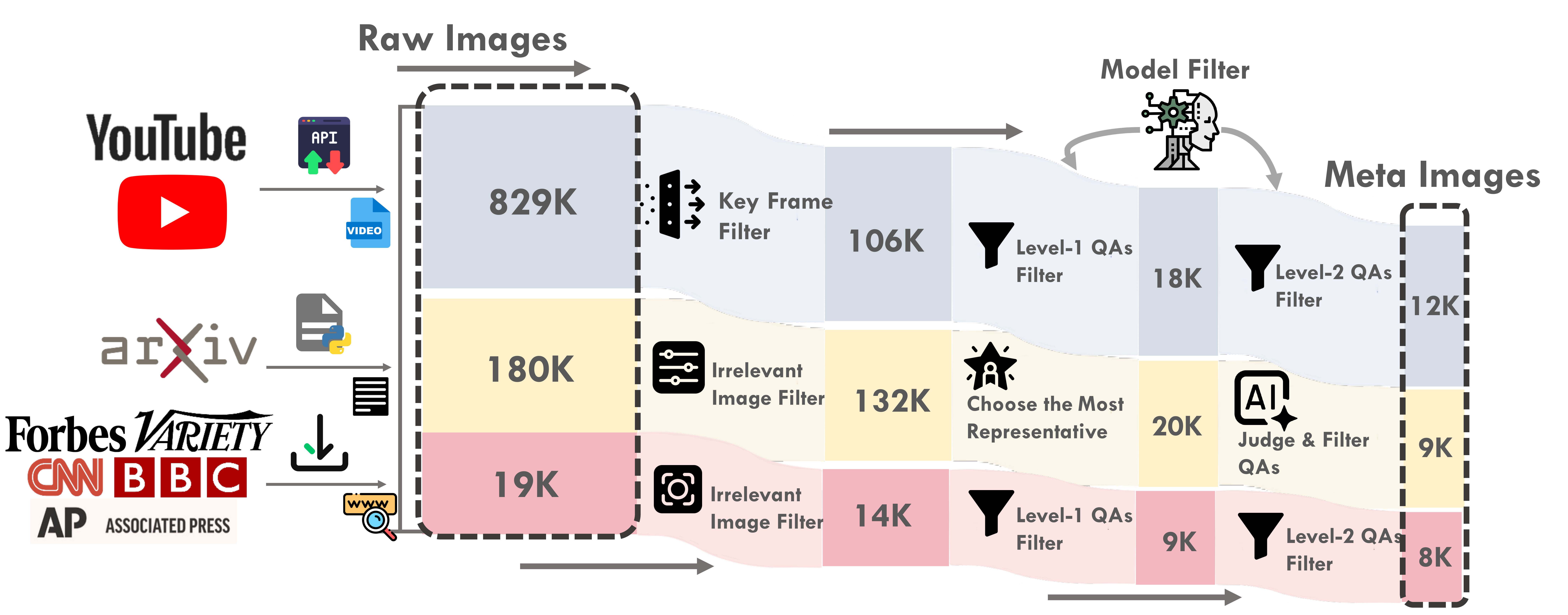}
    \vspace{-0.5em}
    \caption{The filtering process for \dataset shows how images and synthesized questions are removed throughout the data pipeline for news articles, YouTube videos, and arXiv papers.}
    \label{fig:filter_data_count}
    \vspace{-1em}
\end{figure*}

\begin{figure*}[!t]
    \centering
    \vspace{-1em}
    \includegraphics[width=0.9\linewidth]{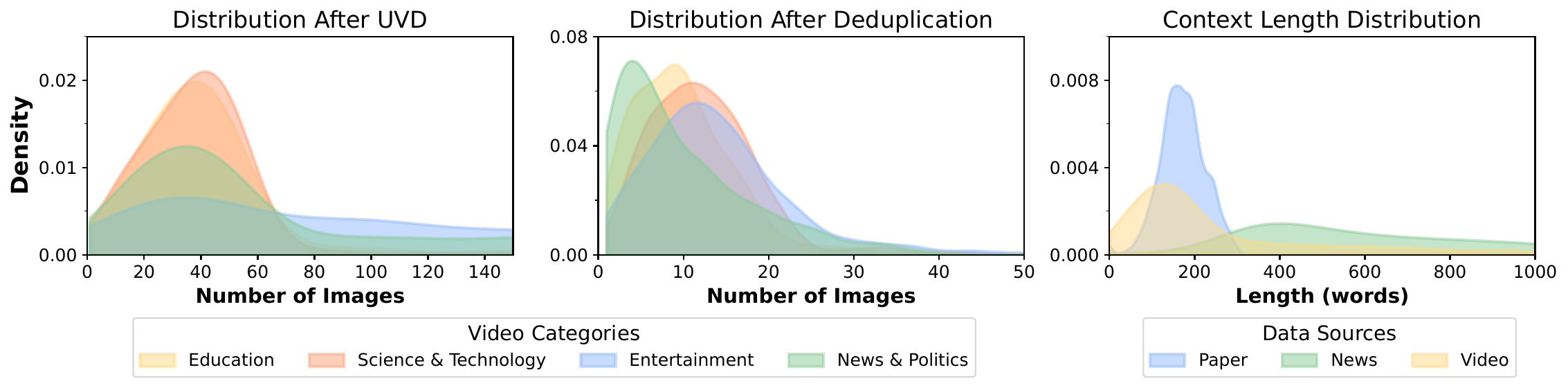}
    \vspace{-0.5em}
    \caption{\textbf{(Left)} Image size distribution in YouTube image filtering pipeline. \textbf{(Right)} Textual context length distribution for each question.}
    \label{fig:image distribution}
    \vspace{-1.5em}
\end{figure*}

\header{Videos.}
With the rising popularity of video platform such as YouTube and TikTok, videos have become increasingly dominant channels for human to access real-world visual knowledge. They excel at capturing up-to-date visual knowledge and inherently convey dynamic, multimodal information about latest events. We choose YouTube as the source for our video knowledge base for their diversity, easy-to-crawl and \texttt{CC-by-4.0} license. The data collection pipeline is detailed as follows:
\begin{enumerate}[label=$\triangleright$, leftmargin=10pt, itemsep=0pt]
\item \textbf{YouTube data preprocessing.}
% \TODO{What is target video? list your principles here. 1. English Language 2. Video Length 3. YouTube topic/category} 
% Target videos for this dataset are required to be in 
% We limit the target videos into
% English (for linguistic uniformity), not exceed 10 minutes in duration (to manage information volume and simplify data processing), and include English subtitles. 
We restrict the target videos to English-language content (to ensure linguistic uniformity), a maximum duration of 10 minutes (to control information volume), and the presence of English subtitles.
We collect videos themselves, official or auto-generated subtitle, video titles, descriptions, and other information (such as publication dates, \emph{etc.}). 

\item \textbf{Subtitle-based video split.}
Given that a video may contain multiple independent news events, we need to split them into small video clip. Video subtitles stand out to be high-quality and authentic context.  We use an LLM to refine the transcribed subtitle files by removing oral expressions and irregular breaks before using the cleaned content as the basis for video segmentation. To ensure segmentation quality, we conduct manual evaluation, achieving a passing rate of 98\%.

% $\triangleright$ \textbf{Informative Keyframe Selection.}
% For each video clip, we design and implement a multi-stage keyframe extraction and curation pipeline aimed at accurately capturing the core visual content of the clip:
% \begin{itemize}[leftmargin=*,itemsep=0pt]
\item \textbf{Initial keyframe identification and preprocessing.} To handle video content efficiently, we implemented a three-step process to extract unique, high-quality frames. First, we used UVD\footnote{\url{https://github.com/zcczhang/UVD}} to identify key candidate frames from videos, eliminating redundancy. Second, we applied a Perceptual Hash algorithm to remove duplicates while preserving only the clearest version of each image. Third, we employed the DocLayout-YOLO\footnote{\url{https://github.com/opendatalab/DocLayout-YOLO}} model to crop images, removing textual elements and irrelevant information (example in Figure \ref{fig:layout}). As a final filter, we removed near-duplicate images by excluding those with a perceptual hash Hamming distance below 25.

\item \textbf{LLM-driven keyframes selection.} After two rounds of deduplication, a video clip still contains 5-20 keyframes (shown in Figure \ref{fig:image distribution}), which is too duplicate to represent an event. Therefore, we leverage an MLLM to identify top-K images by their relevance to the topic and refined subtitles, visual informativeness, and representativeness of the remaining keyframes. Finally, visual information within each video segment is highly condensed and thematically relevant.
\end{enumerate}
% This refined pipeline ensures that the 
% \end{itemize}

\header{Academic papers.}
Academic papers, as a form of disseminating cutting-edge knowledge, serve as a stable source for transmitting the latest visual knowledge \citep{shabtay2024livexiv,li2024multimodal}. Therefore, we collect papers from arXiv and balanced them by category, selecting papers from Physics, Computer Science, and Mathematics.\footnote{Due to the complex anti-scraping mechanisms of bioRxiv and medRxiv, we fail to achieve a completely synthetic data collection. Thus, after attempting to collect these papers, we abandon the effort.}
\begin{enumerate}[label=$\triangleright$, leftmargin=10pt, itemsep=0pt]
\item \textbf{Article and image preprocessing.} 
We deliberately collect web pages of arXiv papers across various domains. From these pages, we extract the title, abstract, authors, section content, images, and image captions of the papers by beautifulsoup. During image processing, we filter out irrelevant images that do not conform to specific formats (\emph{e.g.}, icons, SVGs).

\item \textbf{Key image selection.}
Typically, academic papers contain five or more images, which can be excessive and visually overwhelming when representing a paper's content. Our selection criteria prioritize images that exhibit significant diversity across different papers, focusing specifically on architectural diagrams and illustrations of key findings. We deliberately avoid common images found in most papers, such as standard statistical visualizations or text-heavy figures. Finally, we select 1-2 distinctive images for each paper that serve as visual signatures.
\end{enumerate}

\begin{table}[!t]
    \centering
    \caption{Overall statistics of \dataset. For Level~1 and Level~2 questions, we provide high-quality synthetic answers that fully reveal the image context, with \textit{Avg. Len.} indicating the answer length.}
    \label{tab: overall statistics}
    \setlength{\tabcolsep}{9pt} % Default value: 6pt
    \scalebox{0.8}{
    \begin{tabular}{l|cccccc}
    \toprule[1.5pt]
        \textbf{Category} & \textbf{Images} & \textbf{\#Question} & \textbf{Level 1} & \textbf{Level 2} & \textbf{Avg. Len.} & \textbf{Purpose}\\
        \midrule
        News Article & 7,579 & 38,809 & 7,579 & 31,230 & 749 &-\\
        YouTube Videos & 11,948 & 43,168 & 11,948 & 31,220 & 311 &-\\
        Academic Paper & 8,961 & 25,166 & 9456 & 16,205 & 597 &-\\
        \midrule
        Avg. per Sample & 1 & 3.86 & 1 & 2.86 & 517 & - \\
        \midrule
        Test Split & 1,500 & 3,000 & 1,500 & 1,500 & 544 & Exp. 1\\
        Training Split & 26,988 & 104,143 & 26,988 & 77,150 & 496 & Exp. 2\\
        \bottomrule[1.5pt]
    \end{tabular}}
\end{table}

\subsection{Visual question answering generation}
% We develop our visual knowledge seeking and updating dataset with two distinct question levels. Level 1 consists of straightforward visual entity recognition questions based on filtered images and their metadata. 
% Level 2 features more complex multi-hop cross-modal reasoning questions that require models to use complete image context as a starting point to navigate through related textual information. 
% All questions undergoes strictly filter, aiming to systematically evaluates models' capabilities in both basic visual factuality seeking and advanced cross-modal information synthesis.
We construct our visual knowledge seeking and updating dataset with two distinct question levels. Level~1 includes straightforward visual entity recognition questions based on filtered images and metadata. Level~2 presents complex, multi-hop cross-modal reasoning questions that require models to use full image context to navigate related textual information. All questions are rigorously filtered to systematically evaluate models' abilities in both basic visual fact retrieval and cross-modal synthesis.

% See Appendix \ref{appendix:details} for more details.
% \TODO{
\begin{itemize}[leftmargin=*,itemsep=0pt]
    \item \textbf{Level-1 questions} focus on substantive elements such as people, objects, or locations, while avoiding queries without visual knowledge that solely reliant on visual content within the image like color or shape. We employ GPT-4.1 to filter unqualified QAs. This filter deletes overly brief news items and discards questions whose answers are mere simple labels (\emph{e.g.}, ``city''), news platform names (\emph{e.g.}, CNN); meanwhile, it retains questions whose answers are specific personal name, locations, events, products, organizations. After filtering, 26.7\% samples are discarded. Finally, we require model to synthesis a single chain-of-thought~\citep{wei2022chain} describing the historical or social event depicted, and finally answer the question with reasons.
    \item \textbf{Level-2 questions} require deeper contextual cross-modality reasoning. We ask GPT-4.1 to generate multi-hop visual reasoning questions that naturally reference image details and the textual metadata context, spaning exactly seven types (\emph{i.e.}, location, person, organization, time, event, count, reason). 
    To ensure the verifiability of the answers, we employ GPT-4.1 providing it with questions, images, and their corresponding context, and we retain only those Level~2 questions that the model correctly answers. As a result, 13.0\% samples are discarded after filtering. Similarly, the model offers a paragraph analyzing the event's details in a step-by-step manner.
\end{itemize}
% }

\subsection{Dataset statistics and split}
Finally, we collect 107,143 carefully curated VQA dataset with 26,988 unique images as shown in Table~\ref{tab: overall statistics}. As illustrated in Figure~\ref{fig:example}, the dataset covers a diverse range of topics with representative examples, showcasing its breadth and richness in both content and modality. 
For evaluation, we manually select a test split featuring 1,500 unique images and 3,000 questions for evaluating model's visual seeking capability. Our requirement is that each queries \textit{(1)} can be searched by human using purely web brower (with reverse visual search function) within 10 minutes; \textit{(2)} Following SimpleQA~\citep{wei2024measuring}, we set up a small pool of \emph{state-of-the-art} models (\emph{i.e.}, GPT-4.1, Gemini-2.5-Pro, Claude-3.7-Sonnet, and Qwen2.5-VL-72B), and filter out the questions that half of the model can answer correctly.

% \subsection{Test Set Construction}
% \TODO{how we manually filter or apply strict restriction to curate the testing subset. Give some statistics}

%% file: sec/4-experiments.tex
\begin{table*}[!t]
\centering
\setlength{\tabcolsep}{6pt}

\caption{Accuracy (\%) of visual factuality seeking benchmark in open-ended format across different models across difficulty levels and data sources.}
\label{tab:model_difficulty_datasource}
\vspace{-0.5em}
\scalebox{0.85}{
\begin{tabular}{@{\extracolsep{\fill}}l|l|cccc|cccc}
\toprule[1.5pt]
\multirow{2}{*}{\textbf{Model}} & \multirow{2}{*}{\textbf{Cutoff}} & \multicolumn{4}{c|}{\textbf{Level 1}} & \multicolumn{4}{c}{\textbf{Level 2}} \\
& & \textbf{News} & \textbf{Video} & \textbf{Arxiv} & \textbf{Avg.} & \textbf{News} & \textbf{Video} & \textbf{Arxiv} & \textbf{Avg.} \\
\midrule
\multicolumn{9}{c}{\textbf{\textit{w.o.} Search}} \\
\midrule
GPT-4.1 & Jun. 2024 & 27.0 & 22.0 & 0.4 & 16.5 & 5.2 & 7.2 & 0.2 & 3.0 \\
GPT-4.1-mini & Jun. 2024 & 24.6 & 19.6 & 0.2 & 14.8 & 4.0 & 7.8 & 0.4 & 4.0 \\
GPT-4.1-nano & Jun. 2024 & 13.0 & 13.0 & 0.0  &  8.6  & 2.2 & 6.0 & 0.4 &  2.9\\
Gemini-2.5-Flash & Jan. 2025 & 25.8 & 18.4 & 0.8 & 15.0  & 4.6 & 4.4 & 4.0 &  4.3\\
Gemini-2.5-Pro & Jan. 2025 & 28.0 & 17.4 & 0.6 &  15.3  & 4.4 & 2.4 & 1.2  &  2.7\\
Gemma-3-27B-It & Aug. 2024 & 21.0 & 16.4 & 1.0 &  12.8 & 3.8 & 4.6 & 6.2 & 4.9\\
Claude-3.7-Sonnet & Oct. 2024 & 26.2 & 16.4 & 0.6 &   14.3 & 2.2 & 4.4 & 4.4 &  3.7\\
Qwen-2.5-VL-7B-Instruct & Unknown\footnote{We only find the knowledge cutoff of Qwen-2.5-Coder in \url{https://llm-stats.com/models/qwen-2.5-coder-32b-instruct}, which is reported as Mar. 2024.} & 20.2 & 13.4 & 0.2 &  11.3  & 3.8 & 5.4 & 2.0 & 3.7 \\
Qwen-2.5-VL-32B-Instruct & Unknown & 25.2 & 16.4 & 0.4 &  14.0  & 4.2 & 5.6 & 1.2   &  3.7\\
Qwen-2.5-VL-72B-Instruct & Unknown & 12.4 & 9.4 & 0.0 &  7.3  & 1.4 & 3.6 & 3.6 & 2.9 \\
Llama-4-Scout & Aug. 2024 & 20.6 & 16.4 & 0.0 &  12.1  & 4.0 & 5.0 & 2.8 & 3.9 \\
Llama-4-Maverick & Aug. 2024 & 20.2 & 19.0 & 0.6 &   13.3 & 5.8 & 6.0 & 5.2 & 5.7 \\
\midrule
\multicolumn{9}{c}{\textbf{\textit{w.} Text Search}} \\
\midrule
GPT-4.1 & Jun. 2024 & 25.0 & 21.4 & 0.6 & 15.6 & 3.6 & 5.6 & 3.8 & 4.3 \\
Gemini-2.5-Pro & Jan. 2025 & 17.6 & 9.2 & 0.2 & 9.0 & 2.0 & 1.6 & 1.0 & 1.5 \\
Claude-3.7-Sonnet & Oct. 2024 & 24.6 & 16.6 & 0.0 &13.7  & 2.0 & 3.6 & 4.8 & 3.5 \\
\midrule
\multicolumn{9}{c}{\textbf{\textit{w.} Native Image Search}} \\
\midrule
GPT-o3 & Jun. 2024 & \underline{33.6} & \textbf{33.6} & \underline{2.6} & \underline{23.3} & \underline{14.6} & \underline{14.9} & \underline{17.8} & \underline{15.8}\\
\midrule
\multicolumn{9}{c}{\textbf{\textit{w.} MM-Search \citep{jiang2024mmsearch}}} \\
\midrule
GPT-4.1 & Jun. 2024 & \textbf{42.0} &  \underline{33.0} & \textbf{36.1} & \textbf{33.4} &  \textbf{27.2} & \textbf{15.2} &  \textbf{48.8} & \textbf{30.4} \\
\bottomrule[1.5pt]
\end{tabular}}
\vspace{-1em}
\end{table*}

% \section{Benchmarking Models Visual Seeking Capabilities}
\section{How well do current MLLMs seek out visual knowledge?}
\label{sec:exp}

Keeping pace with the latest visual knowledge is crucial for helpful assistants immersively involved in human life and solving users' problems. Here we investigate \textit{how well do current MLLMs seek out visual knowledge} and evaluate their effectiveness in leveraging external tools to access this knowledge. Moreover, as the test set is authentic and validated by humans, it can serve as visual factuality benchmarks to measure the scientific phenomenon known as calibration \citep{wei2024measuring}, \emph{i.e.}, whether the models \textit{``know what they know''} about the visual content. 

\subsection{Experiment setups}
\header{Models.} 
We conduct a series of zero-shot testing for a diverse range of \emph{state-of-the-art} MLLMs without online searching capability to validate our dataset, including Gemini-2.5-Flash/Pro~\citep{google2025gemini2.5pro}, GPT-4.1 family~\citep{openai2025gpt41}. We also conduct experiments on open-source MLLMs, \emph{e.g.,} Qwen2.5-VL-3/7/32/72B~\citep{yang2024qwen2}, Gemma-3-27B-it~\citep{team2025gemma}, and Llama-4-Scout/Maverick~\citep{2025llama4}. 
Native image search model GPT-o3~\citep{openai2025gpto3} is also included as a strong baseline. Searching within text using openrouter\footnote{\url{https://openrouter.ai/}} web search API and GUI-based image search framework MM-Search~\citep{jiang2024mmsearch} are also included in our experiments.

\header{Evaluation.} 
% We follow prompts from the Humanity Last Exam~\citep{phan2025humanity} to collect models' answers and their corresponding confidence levels for model performance and calibration measurement. 
We use prompts from the Humanity Last Exam~\citep{phan2025humanity} to gather model responses and confidence scores for evaluating performance and calibration.
% To avoid egocentric bias~\citep{panickssery2024llm}, we employ grading metrics evaluated by Gemini-2.0-Flash, following the methodology in SimpleQA \citep{wei2024measuring}, to measure \textit{Correct}, \textit{Not Attempted}, \textit{Incorrect responses} and the resulting F-score. All results are reported as averages across three independent evaluations.
To avoid egocentric bias~\citep{panickssery2024llm}, we adopt grading metrics evaluated by Gemini-2.0-Flash, following SimpleQA~\citep{wei2024measuring}, to evaluate \textit{Correct}, \textit{Not Attempted}, and \textit{Incorrect} responses, along with the resulting F-score. All results are reported as averages over three independent evaluations.

\begin{table}[!t]
    \centering
    \caption{Comparison performance between different models and information seeking methodology on detailed categories in News subset.}
    \setlength{\tabcolsep}{3pt} % Default value: 6pt

    \label{tab: news detailed category}
    \resizebox{\textwidth}{!}{
    \begin{tabular}{l|cccccc|cccccccc}
    \toprule[1.5pt]
        \multirow{2}{*}{Model} & \multicolumn{6}{c|}{Level 1} & \multicolumn{8}{c}{Level 2}\\
        & Loc. & Per. & Org. & Eve. & Obj. & Avg. & Loc. & Per. & Org. & Time & Cou. & Rea. & Eve. & Avg. \\
    \midrule
    \multicolumn{15}{c}{\textbf{\textit{w.o.} Search}} \\
    \midrule
        GPT-4.1 & 38.81 & 6.28 & 50.72 & 15.19 & 35.89 & 27.03 & 1.75 & 0.00 & 11.68 & 3.82 & 7.84 & 1.63 & 0.00 & 5.05 \\
        GPT-4.1-mini & 33.33 & 10.91 & 45.59 & 11.86 & 31.73 & 24.60 & 3.57 & 0.00 & 8.82 & 0.00 & 10.24 & 0.00 & 0.00 & 4.00 \\
        GPT-4.1-Nano & 16.16 & 3.64 & 30.88 & 3.39 & 19.23 & 13.00 & 0.00 & 0.00 & 4.41 & 1.54 & 3.94 & 0.83 & 0.00 & 2.20 \\
        Gemini-2.5-Flash & 26.26 & \underline{37.27} & 35.29 & 7.63 & 27.88 & 25.80 & 3.57 & 0.00 & 1.47 & 3.85 & 8.66 & 4.17 & 0.00 & 4.60 \\
        Gemini-2.5-Pro & 23.23 & \textbf{46.36} & 35.29 & 10.17 & 28.85 & 28.00 & 3.57 & 0.00 & 5.88 & 3.08 & 3.94 & 6.67 & 0.00 & 4.40 \\
        Gemma-3-27B-IT & 24.24 & 15.45 & 38.24 & 8.47 & 25.96 & 21.00 & 3.57 & 0.00 & 8.82 & 1.54 & 7.87 & 0.00 & 0.00 & 3.80 \\
        Claude-3.7-Sonnet & 38.38 & 10.00 & 38.24 & 14.41 & 37.50 & 26.20 & 0.00 & 0.00 & 4.41 & 2.31 & 1.57 & 2.50 & 0.00 & 2.20 \\
        Qwen-2.5-VL-7B & 23.23 & 18.18 & 30.88 & 12.71 & 21.15 & 20.20 & 0.00 & 0.00 & 4.41 & 1.54 & 7.09 & 4.17 & 0.00 & 3.80 \\
        Qwen-2.5-VL-32B & 33.33 & 18.18 & 30.88 & 18.64 & 28.85 & 25.20 & 0.00 & 0.00 & 7.35 & 2.31 & 6.30 & 4.17 & 0.00 & 4.20 \\
        Qwen-2.5-VL-72B & 15.15 & 6.36 & 25.00 & 8.47 & 12.50 & 12.40 & 0.00 & 0.00 & 4.41 & 0.77 & 1.57 & 0.83 & 0.00 & 1.40 \\
        Llama-4-Scout & 26.26 & 13.64 & 35.29 & 8.47 & 26.92 & 20.60 & 3.57 & 0.00 & 4.41 & 3.08 & 9.45 & 0.00 & 0.00 & 4.00 \\
        Llama-4-Maverick & 20.20 & 19.09 & 36.76 & 5.93 & 26.92 & 20.20 & 0.00 & 0.00 & 10.29 & 2.31 & 13.39 & 1.67 & 0.00 & 5.80 \\
\midrule
\multicolumn{15}{c}{\textbf{\textit{w.} Text Search}} \\
\midrule
GPT-4.1 & 36.36& 2.73& 48.53& 13.56& 34.62& 25.00& 3.57& 5.88& \underline{5.88} & 3.85& 4.72& 0.83& 0.00& 3.60 \\
Gemini-2.5-Pro & 18.18& 12.73& 29.41& 10.17& 23.08& 17.60& 0.00& 0.00& 4.41& 1.54& 2.36& 1.67& 0.00& 2.00 \\
Claude-3.7-Sonnet & 29.29   & 6.36  & 33.82       & 18.64& 40.38 & 24.60  & 3.57    & \underline{5.88}  & 1.47        & 1.54& 3.15 & 0.83  & 0.00 & 2.00 \\
\midrule
\multicolumn{15}{c}{\textbf{\textit{w.} Native Image Search}} \\
\midrule
GPT-o3 & \textbf{47.47} & 4.55 & \textbf{57.35} & 23.73 & \underline{47.12} & \underline{33.60} & \textbf{17.86 }& 0.00 & \underline{20.59} & \underline{7.69} & \underline{17.32} & 17.50 & \textbf{10.00} & \underline{14.60}\\ 
\midrule
\multicolumn{15}{c}{\textbf{\textit{w.} MM-Search} \citep{jiang2024mmsearch}} \\
\midrule
GPT-4.1  & \underline{42.86} & 35.78 & \underline{55.88} & \textbf{33.05} & \textbf{50.00} & \textbf{42.00} & \underline{15.50} & \textbf{46.43} & \textbf{23.53} & \textbf{30.88} & \textbf{42.52} & \textbf{20.00} & 0.00 & \textbf{27.20} \\
        \bottomrule[1.5pt]
    \end{tabular}}
    \vspace{-1em}
\end{table}

\subsection{Results and analysis}
\header{Larger-scale base models demonstrate improved performance in visual factuality recognition, and proprietary models retain a clear advantage.} 
% For models within the same knowledge cutoff (\emph{e.g.}, GPT-4.1 or Llama-4 family), we observe that increasing model size leads to consistently better accuracy across all question difficulty levels. For instance, GPT-4.1-nano achieves only an average of 8.6\% on Level 2 News category, whereas larger model GPT-4.1 and GPT-4.1-mini reaches 14.8\% and 16.5\%. 
For models sharing the same knowledge cutoff (\emph{e.g.}, the GPT-4.1 and Llama-4 families), we observe that increased model size consistently correlates with improved accuracy across all levels of question difficulty. For example, GPT-4.1-nano achieves an average accuracy of only 8.6\% on Level 2 questions in the News category, while the larger GPT-4.1-mini and GPT-4.1 models attain 14.8\% and 16.5\%, respectively.
% Furthermore, open-source models still lag behind proprietary models in overall performance, and within Qwen-2.5-VL family, the smaller 7B model outperforms 72B one. 
% One plausible guess is that they are trained on different corpus of data, given that Qwen-2.5-VL-32B are released 2 monthes after other model release, therefore achieving highest score among the family. 
% Notably, \textbf{Llama-4-Maverick} achieves the best results across nearly all dimensions in Level 2 questions in News subset with an overall accuracy of \textbf{5.8\%}, which may be attributed to its large training corpus coverage.
Furthermore, open-source models still lag behind proprietary counterparts in overall performance. Within the Qwen-2.5-VL family, the smaller 7B model surprisingly outperforms the larger 72B variant. One plausible explanation is that these models are trained on different corpora; notably, the Qwen-2.5-VL-32B model is released two months after the others, possibly benefiting from a more refined or expanded dataset, which may account for its highest score within the family.
Notably, Llama-4-Maverick achieves the best performance across nearly all dimensions on Level 2 questions in the News subset, with an overall accuracy of 5.8\%, potentially due to its extensive training corpus coverage.

\header{A multimodal search engine matters.} 
% GPT-4.1's average accuracy more than doubles from 16.5\% (Table~\ref{tab:model_difficulty_datasource}) to 33.4\% when using multimodal search tools, with substantial gains on challenging Level 2 questions—achieving 27.2\% and 15.2\% on News and Video subsets respectively. 
% Similarly, \textbf{GPT-o3}, which shares the same knowledge cutoff as GPT-4.1, shows dramatic improvement from 3.0\% to 15.8\% on Level 2 questions when leveraging these tools, demonstrating the considerable potential of RL-based multimodal tool-use agents. These results clearly indicate that integrating multimodal search capabilities particularly benefits questions beyond models' internal knowledge scope. Figure~\ref{fig:mm} visually demonstrates this advantage by contrasting GPT-4.1~\citep{openai2025gpt41}'s successful responses with multimodal search against its failures.
From Table~\ref{tab:model_difficulty_datasource}, we can see that GPT-4.1's average accuracy more than doubles—from 16.5\% to 33.4\%—when augmented with multimodal search tools. This improvement is particularly striking on challenging Level 2 questions, where accuracy rises to 27.2\% and 15.2\% on the News and Video subsets, respectively.
Similarly, \textbf{GPT-o3}, which shares the same knowledge cutoff as GPT-4.1, shows a substantial gain on Level 2 questions, improving from 3.0\% to 15.8\% with tool use. These results underscore the considerable potential of reinforcement learning–based multimodal tool-use agents.
Overall, the integration of multimodal search capabilities proves especially beneficial for questions that exceed the models' internal knowledge. Figure~\ref{fig:mm} further illustrates this advantage by contrasting GPT-4.1's~\citep{openai2025gpt41} successful answers with tool use against its failures without it.

%confidence
\begin{table*}[!t]
\centering
\setlength{\tabcolsep}{8pt} % Default value: 6pt

\caption{Detailed breakdown on non-search model's failure categories. See definition of F-score in Appendix \ref{Appendix: detailed experiment setups}. \textit{``Correct \& given attempted''}: model answer correctly, out of only questions that were attempted (\emph{i.e.}, questions answered correct and incorrectly).} % Adapted caption
\label{tab:model_performance_styled} % Adapted label
\vspace{-0.5em}
\resizebox{\textwidth}{!}{%
\begin{tabular}{l|ccccc} % 'l' for Model, 'c' for numeric columns as in example, @{...} for fill
\toprule[1.5pt] % Thicker top rule, as in example
% Header structure inspired by the user's example, aiming for a two-row header setup
% For columns like "Not attempted" and "Correct given attempted", the text is split over two conceptual lines.
\textbf{Model} & \textbf{Correct} & \textbf{Not attempted} & \textbf{Incorrect} & \textbf{Correct \& given attempted} & \textbf{F-score} \\
\midrule

GPT-4.1 & \underline{16.0} & 52.4 & 31.6 & 33.6 & \underline{21.7} \\
GPT-4.1-mini & 14.3 & 44.0 & 41.7 & 25.5 & 18.3 \\
GPT-4.1-nano & 7.6 & 58.0 & 34.4 & 18.1 & 10.7 \\
Gemini-2.5-Flash & 15.2 & 59.3 & 25.5 & 37.3 & 21.6 \\
Gemini-2.5-Pro & \textbf{16.2} & 58.4 & \underline{25.4} & \underline{38.9} & \textbf{22.9} \\
Gemma-3-27B-It & 12.4 &\textbf{ 24.0 }& 63.6 & 16.3 & 14.1 \\
Claude-3.7-Sonnet & 14.2 & 64.2 & \textbf{21.6} & \textbf{39.7} & 20.9 \\
Qwen-2.5-VL-7B-Instruct & 12.0 & 61.0 & 27.0 & 30.8 & 17.3 \\
Qwen-2.5-VL-32B-Instruct & 14.7 & 40.9 & 44.4 & 24.9 & 18.5 \\
Qwen-2.5-VL-72B-Instruct & 6.9 & \underline{36.2} & 56.9 & 10.8 & 8.4 \\
Llama-4-Scout & 12.3 & 52.4 & 35.3 & 25.8 & 16.7\\
Llama-4-Maverick & 12.6 & 55.8 & 31.6 & 28.5 & 17.5 \\

\bottomrule[1.5pt] % Thicker bottom rule, as in example
\end{tabular}%
}
\vspace{-1em}
\end{table*}
%confidence

\begin{wrapfigure}{r}{0.38\textwidth}
    \centering
    \vspace{-1em}
    \includegraphics[width=0.9\linewidth]{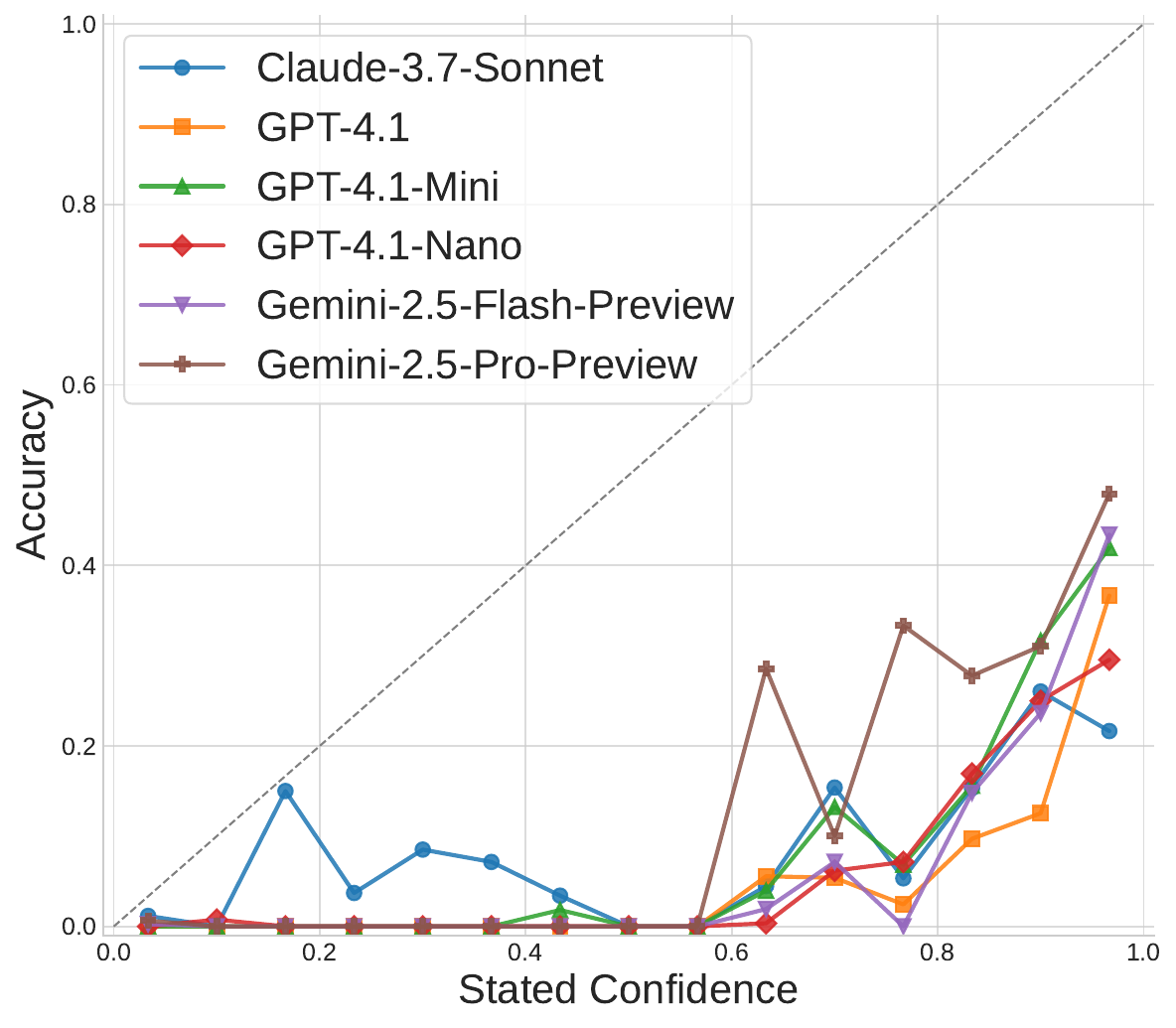}
    \vspace{-1em}
    \caption{Calibration of MLLMs when encountering unknown visual knowledge based on their stated confidence.}
    \label{fig:measuring calibration}
    \vspace{-1em}
\end{wrapfigure}

\header{Measuring visual factuality calibration.} Figure~\ref{fig:measuring calibration} demonstrates a positive correlation between stated confidence and accuracy across models, though with significant calibration issues. Claude-3.7-Sonnet exhibits better calibration at lower confidence levels, while Gemini-2.5-Pro performs better at higher confidence levels, yet all models struggle to properly align their confidence with actual accuracy. Consistent with prior research~\citep{wei2024measuring,gao2024best}, GPT-4.1 shows superior calibration compared to its smaller variants (\emph{i.e.}, GPT-4.1-mini and nano), supporting the finding that larger models tend to be better calibrated. 
% Notably, all model performances fall substantially below the ideal $y = x$ line, indicating systematic overconfidence in visual factuality assessments, highlighting significant opportunities for improving MLLM calibration.
Notably, the performance of all models falls significantly below the ideal $y = x$ line, indicating a consistent pattern of overconfidence in visual factuality assessments and underscoring substantial opportunities for improving MLLM calibration.

\section{Can we update MLLMs with new visual knowledge?} % Without Hurting MLLM

Recent benchmarks reveal that state-of-the-art MLLMs struggle with tasks requiring up-to-date visual knowledge, as the information learned during pretraining quickly becomes obsolete with the emergence of new entities and events. While retrieval-augmented generation (RAG) can incorporate external knowledge, it introduces latency and often fails to resolve semantically similar visual inputs. PEFT offers a more efficient solution for updating MLLMs with new visual concepts. Building on prior work~\citep{ravaut2024much, zeng2024visual, chen2024lifelong}, we ask: \textit{Can MLLMs be efficiently updated with new visual knowledge?} We investigate this by examining how updates affect both retention of prior visual understanding and general multimodal reasoning.

\subsection{Experiment setups}
\header{Models and PEFT settings.} We select Qwen2.5-VL-3B/7B-Instruct for their superior performance in vision-language understanding capability. We implement two popular PEFT methods LoRA \citep{hu2022lora} and DoRA \citep{liu2024dora} with diverse rank ranging from [16, 64] for ablation study. See Appendix \ref{Appendix: detailed experiment setups} for detailed experiment settings.

\header{Data size and format ablation.} We construct our instruction tuning dataset using three different formats for ablation study: \textit{(1)} QA: Question + Ground Truth. \textit{(2)} MCQA: Question + Multiple-choices + Correct letter + Ground Truth. \textit{(3)} QAR: Question + Ground Truth + Reasoning process. See Figure~\ref{figure: data_example} for an example.

\header{Evaluation.} Following previous knowledge updating research~\citep{wang2025codesync}, we synthesize a 1,500-sample validation set in MCQA format from the training set, with 500 samples from each category and each question rephrased by GPT-4.1. To evaluate the model's general performance, we select MMMU~\citep{yue2024mmmu} and MMstar \citep{chen2024we} as benchmarks. We conduct three tests for each checkpoint and take the average performance as the final result to ensure robust evaluation. We do not restrict the answer format in order to assess performance under typical usage conditions, and we leverage Gemini-2.0-Flash \citep{team2023gemini} as the judge for evaluation.

\subsection{Results and analysis}

\begin{figure*}[!t]
    \centering
    \vspace{-1em}
    \includegraphics[width=\linewidth]{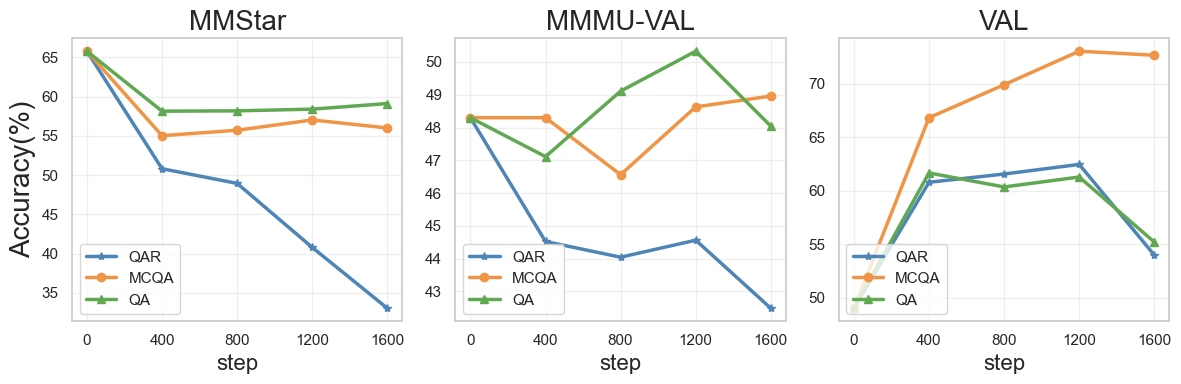}
    \vspace{-1.5em}
    \caption{The results of visual knowledge updating with different data formats.}
    \label{fig:lora}
    \vspace{-1.5em}
\end{figure*}

\header{PEFT methods enable efficient updating of visual knowledge.}
Among the three data formats we design, Figure~\ref{fig:lora} shows that model using direct multiple-choice questions with concise answers yield faster and more effective learning during the visual knowledge acquisition phase. As training progresses, the model's performance steadily improves in validation set, continuing to rise into the second epoch. In contrast, the other two formats exhibite early performance saturation, with gains peaking at the first training step and showing no further improvement thereafter. Notably, training on the visually knowledge-intensive \dataset dataset—particularly with straightforward answers and multiple-choice questions—leads to a 4.2\% improvement in MMMU.

\begin{wrapfigure}{r}{0.35\textwidth}
    \centering
    \vspace{-2em}\includegraphics[width=0.98\linewidth]{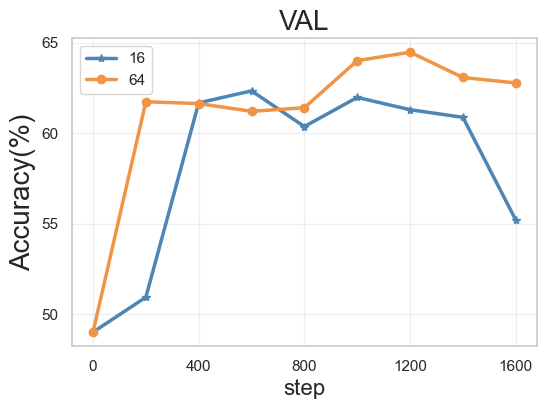}
    \vspace{-1em}
    \caption{Visual knowledge updating with different LoRA rank.}
    \label{fig:val_image}
    \vspace{-2em}
\end{wrapfigure}

\header{Visual knowledge updating harm on model's visual perception capability.} 
In our experiments, we observe a consistent degradation in the model's performance on MMStar, regardless of variations in rank and number of training steps, or training formats. Specifically, models trained using the simple answer format exhibit a performance drop from 65.80\% to 58.16\%. Notably, this decline plateaus even with continued training over multiple steps. These findings suggest an inherent conflict between enhancing visual knowledge through intensive updates and preserving the model's visual perception capability.

\header{Higher rank of LoRA yields  better performance.} 
% From our ablation study results with different ranks, we can see that higher ranks indeed achieve greater visual knowledge capability, loading more latest visual entities. As shown in Figure~\ref{fig:lora}, models with higher rank continuosly stand out, showing an average improvement of \textcolor{red}{X\%} on the validation subset.
Our ablation study across varying rank settings demonstrates that higher-rank LoRA configurations consistently enhance visual knowledge capabilities, particularly in assimilating recent visual entities. As illustrated in Figure~\ref{fig:val_image}, models with higher ranks consistently outperform their lower-rank counterparts, with an average improvement of 5.4\% on the validation subset.

%% file: sec/5-conclusion.tex
\section{Conclusion}
% In this paper, we have introduced \dataset, the first synthetic data engine featuring latest visual content paired with cross-modality multihop reasoning questions, specifically designed to advance research in visual knowledge seeking and updating. 
% Our experiments reveal that search-based multimodal RAG approaches provide drastically improvement for handling contemporary visual information, while our parameter-efficient fine-tuning experiments offer valuable insights into balancing adapter capacity with core model capabilities.
% We believe that the dataset, benchmark, and findings presented in this work provide a solid foundation for future research aimed at enhancing the seeking and updating capabilities of MLLMs with live visual knowledge.
In this paper, we introduce \dataset, the first synthetic data engine to feature up-to-date visual content paired with cross-modal, multi-hop reasoning questions—specifically designed to advance research in visual knowledge seeking and updating for MLLMs.
Our experiments demonstrate that search-based multimodal RAG methods lead to significant improvements in handling contemporary visual information. Additionally, our PEFT studies provide valuable insights into balancing adapter capacity with the underlying model's capabilities.
We believe that the dataset, benchmark, and findings presented in this work establish a strong foundation for future research aimed at enhancing the knowledge seeking and updating abilities of MLLMs in dynamic visual contexts.

% \dataset establishes essential infrastructure for research on evolving visual information, providing key insights for building AI systems that can effectively perceive and reason about our dynamic visual world.

\section*{Ackownledgement}
We thank Dongzhi Jiang from MM-Search team \citep{jiang2024mmsearch}, Jingming Wu from MM-search-R1 team \citep{wu2025mmsearchr1} and Jieyu Zhang from University of Washington for their valuable insights and discussion.

%% file: sec/99-appendix.tex
\clearpage
\appendix
\setcounter{page}{1}

\section{Limitations}
Despite our efforts to build a fully automated framework for crawling latest visual knowledge and construct it into dataset, limitations remains. Retrieving news and video data older than two months was extremely difficult due to crawling limitations on platforms like YouTube, potentially causing date-category imbalance. arXiv paper crawling often encountered CAPTCHAs, sometimes leading to incomplete downloads. 

Although our pipeline leverage capable models like GPT-4.1 at each LLM/MLLM-in-the-loop and validated by human annotation of over 500 balanced samples per category (achieving <3\% error rates), may still produce occasional errors in QA generation or filtering, leading to some flawed dataset samples. Furthermore, due to limited computational resources, our visual knowledge update experiments were restricted to parameter-efficient methods (\emph{i.e.}, LoRA, and DoRA) and the Qwen2.5-VL model family, without exploring full fine-tuning or a wider range of models. 

Future work will address these points by including more baselines and investigating more efficient update techniques. Nevertheless, we are committed to continuously crawling new data and updating this dataset annually. This will help ensure the availability of a non-contaminated dataset for future research in visual knowledge seeking and updating.

\input{sec/2-related}

% \section{Details in Constructing \dataset}
% \label{appendix:details}
\section{Dataset collection details}
\subsection{News}
In order to efficiently extract news article data from massive online resources, we design an automated processing. We first use strict URL verification to ensure that only valid article pages are processed, excluding irrelevant content. For the verified URLs, we identify article titles and candidate images. Next, we select visual materials that are highly relevant to the news content. Finally, we use a duplicate article detection and prevention mechanism to identify and remove content that has been included or repeated in the current processing session.

% \begin{itemize}[leftmargin=*,itemsep=0pt]
    $\triangleright$ \textbf{Systematic URL validation and canonicalization protocol.}
    The initial stage of data processing involves a meticulous URL validation and canonicalization protocol to ensure that only valid article pages are pursued for content extraction. Each prospective URL first undergoes a canonicalization procedure, where all query parameters (components following a '?' symbol) are systematically stripped, yielding a standardized base URL essential for consistent referencing and duplicate prevention. 
    
    Following canonicalization, a sophisticated validation mechanism ascertains whether the URL corresponds to a substantive news article rather than an ancillary page such as a category index, multimedia gallery, or author profile. This validation employs a hierarchical, domain-sensitive strategy. For recognized news sources (CNN, BBC, Forbes, Variety, Associated Press News), the system dispatches URLs to dedicated, site-specific validation functions. These functions implement tailored rules, typically leveraging regular expressions and string pattern matching, to identify legitimate article URL structures (\emph{e.g.}, date-based paths like \texttt{YYYY/MM/DD} for CNN, specific segments like \texttt{/news/articles/} for BBC, or complex patterns involving \texttt{/sites/} and author/date components for Forbes) while concurrently excluding known non-article paths (\emph{e.g.}, \texttt{/video/}, \texttt{/gallery/}, \texttt{/tag/}). 
    
    In the event a URL does not originate from these predefined sources, or if specific rules are not met, a generic fallback validation is invoked. This generic checker compares the URL against a curated list of common article-indicating path segments (\emph{e.g.}, \texttt{/article/}, \texttt{/story/}, \texttt{.html}) and, conversely, against a list of non-article indicators (\emph{e.g.}, \texttt{/category/}, \texttt{/search/}). Only URLs that successfully pass this rigorous, multi-layered validation sequence are advanced to subsequent processing stages. Utility functions further support this by enabling secure joining of base URLs with relative paths discovered on index pages.

    $\triangleright$ \textbf{Robust HTML element extraction and initial content identification.}
    Once a URL is validated as a potential article source, its HTML content is parsed to extract key informational components, primarily the title and preliminary image candidates. This process utilizes robust CSS selectors, safeguarded by wrapper functions (\texttt{safe\_select}, \texttt{safe\_select\_one}) that gracefully handle exceptions during selector execution, returning empty lists or null objects instead of halting the process. Title extraction (\texttt{extract\_title}) employs a tiered strategy: for designated news providers (CNN, BBC, Forbes), a predefined dictionary maps site identifiers to a prioritized list of CSS selectors known to target article headlines (\emph{e.g.}, \texttt{h1.pg-headline} for CNN, \texttt{h1.article-headline} for Forbes). The system iterates through these selectors until a valid title element, defined as yielding a textual content Condiciones de Uso y Privacidad Política de Privacidad y Cookies Publicidad Anunciarse con nosotros Contacto Aviso legal Política de Cookies (UE) Transparencia sobre el esclavismo moderno (Reino Unido) (con una longitud superior a cinco caracteres tras stripping), is found. If site-specific selectors fail, or if the source is not explicitly defined, a generic fallback attempts to extract content from the primary \texttt{<h1>} tag. 
    
    Similarly, an enhanced image detection routine (\texttt{enhance\_image\_detection}) is initiated to identify potential article images. This routine also operates on a dispatch model, invoking specialized detection functions (\emph{e.g.}, \texttt{detect\_cnn\_images}, \texttt{detect\_forbes\_images}) for supported news domains. These site-specific functions employ a list of CSS selectors targeting common image elements or containers (\emph{e.g.}, \texttt{.media\_\_image} for CNN, \texttt{.article-image} for Forbes). As a secondary measure within these functions, or if generic detection (\texttt{detect\_generic\_images}) is active, the system may iterate through all \texttt{<img>} tags within the document. This comprehensive scan often includes filters to discard irrelevant images based on keywords in their \texttt{src} attribute (\emph{e.g.}, 'icon', 'logo', 'avatar') or their explicit dimensions (\emph{e.g.}, retaining only images larger than 200x200 pixels), and positive identification through keywords in \texttt{src} or CSS \texttt{class} attributes (\emph{e.g.}, 'photo', 'image', 'hero', 'main-image'). These initial detection steps provide a candidate pool of images for more intensive downstream filtering.

    $\triangleright$ \textbf{Advanced image curation via multi-stage filtering and deduplication.}
    Following initial HTML parsing, a sophisticated image curation pipeline processes the identified candidate images to select a concise set of high-relevance visuals for each article. The process commences with the \texttt{extract\_article\_images} function, which systematically gathers image URLs and associated captions. This involves several targeted extraction methods: (1) Prioritized retrieval of images specified in Open Graph (\texttt{og:image}) and Twitter Card (\texttt{twitter:image}) meta tags, as these often represent the primary article visual. (2) Searching within common semantic HTML structures using a list of \texttt{figure\_selectors} (\emph{e.g.}, \texttt{<figure>}, \texttt{.image-container}, \texttt{.media-with-caption}) to identify images and their corresponding captions (\emph{e.g.}, from \texttt{<figcaption>}, \texttt{.caption}). Relative image URLs are resolved to absolute paths, and base64-encoded images are typically ignored. (3) A broader scan within primary content blocks (identified by \texttt{content\_selectors} like \texttt{<article>} or \texttt{.article-body}) for any remaining \texttt{<img>} tags, with attempts to infer captions from adjacent elements.
    
    The aggregated list of image candidates then undergoes a rigorous filtering and deduplication process within the \texttt{filter\_images} function. For each image URL, its content is fetched, and its dimensions (width, height) are determined using the Pillow (PIL) library, allowing for the calculation of pixel area. Images that cannot be fetched or processed are discarded. Crucially, a visual deduplication step (\texttt{deduplicate\_images}) is performed. If image processing libraries like OpenCV or scikit-image are available, this function compares images pairwise. Images are resized (\emph{e.g.}, to 128x128 grayscale) and their visual similarity is computed, potentially using a hybrid approach involving color histogram comparison (via OpenCV's \texttt{calcHist} and \texttt{compareHist}) and Structural Similarity Index Measure (SSIM, via scikit-image). Images with a similarity score exceeding a predefined threshold (\emph{e.g.}, 0.85) are considered duplicates, and the one with the smaller area is typically removed. If these libraries are unavailable, a fallback deduplication uses simpler heuristics based on common URL filename segments and dimensional proximity.
    The unique, high-quality images are then sorted in descending order by pixel area. 
    
    A final selection pass retains a maximum of four images per article. Furthermore, any image whose area is less than a specified fraction (\emph{e.g.}, 30\%) of the largest image's area for that article is discarded, ensuring that only prominent and contextually significant visuals are preserved. This multi-faceted approach ensures a small, yet highly relevant, set of images for each news item. Utility functions also support downloading these curated images and strategically inserting placeholder tags (\emph{e.g.}, \texttt{<imgN>}) into the article text.

    $\triangleright$ \textbf{Hierarchical duplicate article detection and prevention:}
    To maintain dataset integrity and prevent redundant data processing, a robust, multi-layered duplicate detection mechanism is implemented through the \texttt{DuplicateChecker} class. This system addresses both historical duplicates (against previously collected articles) and intra-session duplicates (within the current data collection run). Upon initialization, the checker loads metadata from all existing article collections, typically stored as JSON files (\emph{e.g.}, \texttt{hot\_topics*.json}), into memory. This historical data populates sets of known URLs (\texttt{seen\_urls}) and titles (\texttt{seen\_titles}) for rapid initial lookups.
    
    When a new article candidate is processed (\texttt{is\_duplicate\_topic}), it is first checked against these in-memory sets for exact URL or title matches. If no exact match is found, a more thorough comparison against the loaded historical topics is performed. This involves several checks in a specific order of computational cost: (1) Exact URL equality. (2) Precise matching of the initial five words of the article titles. (3) Keyword-based similarity: Titles are tokenized into keyword sets (excluding common stopwords), and if the Jaccard index or a similar overlap metric (\emph{e.g.}, intersection size relative to the smaller set, thresholded at 70\% for sets with at least 3 keywords) between the new and an existing title's keywords is high, it's flagged as a potential duplicate. (4) Full title similarity: If previous checks are inconclusive, a more computationally intensive string similarity algorithm (\emph{e.g.}, Levenshtein distance based, with a similarity score $> 0.85$ indicating a duplicate) is applied between the new and existing titles.
    
    For real-time, intra-session duplicate prevention during active crawling (\texttt{is\_duplicate\_realtime}), a slightly streamlined version of this logic is employed, primarily leveraging the dynamically updated \texttt{seen\_urls} and \texttt{seen\_titles} sets, along with the prefix, keyword, and full title similarity checks against the already loaded historical topics. If an article passes all these checks, its URL and title are added to the session's seen sets to prevent its re-processing. This hierarchical approach, combining hash-based lookups with increasingly sophisticated content analysis, ensures high accuracy in duplicate detection while managing computational overhead.
% \end{itemize}

\subsection{Videos}
Our automatic system initially employs \texttt{YouTube Data API} for video discovery and rule-based filtering, followed by the downloading and preprocessing such as format conversion. Then, we process subtitle as context for visual content and perform LLM-driven subtitle refinement to conduct video segmentation base on subtitle timestamp. Finally, we conduct keyframe extraction and multi-stage filtering to get the curated textual and visual metadata.

\begin{itemize}[leftmargin=*,itemsep=0pt]
    \item[$\triangleright$] \textbf{Video downloading, initial filtering, and preprocessing.}
    The initial stage focuses on sourcing and preparing video data. Firstly, we utilize the \texttt{YouTube Data API} to conduct targeted searches based on predefined date ranges, language (English), and content categories (news, education, entertainment, technology). For the retrieved videos, we immediately verify their duration (ensuring it does not exceed 10 minutes) and the availability of English subtitles, thereby performing an initial screening for video resources that meet these basic requirements. Subsequently, the system automatically downloads the videos that pass this initial screening, along with their corresponding English subtitle files. To meet the technical requirements for subsequent keyframe extraction, all downloaded videos automatically undergo format conversion from \texttt{VP9} to \texttt{H.264}. Finally, detailed metadata for each video is extracted, creating structured data records, and the entire workflow integrates comprehensive logging and error handling mechanisms to ensure data collection accuracy and operational stability.

    \item[$\triangleright$] \textbf{Semantic content processing and keyframe extraction.}
    For each video, subtitles and corresponding timestamps are processed to extract textual content. This extracted text is then refined using a Large Language Model (\texttt{LLM}), which adds punctuation and segments the content into distinct topics, each associated with specific time intervals. Subsequently, the videos are clipped according to these time intervals to ensure thematic consistency within each resulting segment. This segmentation is crucial to prevent misalignment between subsequently selected images and their corresponding textual context, which could otherwise lead to the large model generating 'hallucinations'. Following this, for these segmented video clips, keyframes are extracted using \texttt{UVD} (Uniform Video Dicing), as the full video content is often too extensive, necessitating the distillation of useful visual information. Given that a single video segment can yield a large number of keyframes, an initial deduplication step is performed using image perceptual hashing with a Hamming distance threshold of 10. Concurrently, the Laplacian operator is employed to calculate image clarity to retain only the clearest image among a set of visually similar ones.

    \item[$\triangleright$] \textbf{Advanced image curation and multimodal content generation.}
   We observe that for some news screenshots, subtitle overlays introduce significant textual interference. To address this, \texttt{DocLayout-YOLO} technology is utilized to crop these images, aiming to remove or minimize such interfering textual information that might directly reveal key details and unduly influence the large model's responses. These cropped images then undergo a secondary deduplication process, using the same perceptual hashing method as before, but with a revised Hamming distance threshold of 25. After these operations, some may still exhibit an ambiguous correspondence with the textual information. Therefore, we design an image filtering program that leverages \texttt{GPT-4.1} to select the designs images that best align with the textual content while exhibiting minimal textual interference from overlays. Finally, these highly filtered images, in conjunction with the translated and refined video titles and subtitle information, are used to generate question-answering (QA) pairs.
\end{itemize}
\subsection{Academic papers}
    To construct a comprehensive dataset derived from scholarly articles on arXiv, we have developed a systematic collection and processing pipeline. This pipeline is designed to extract and curate textual and visual information from scientific papers. Subsequently, a MLLM is utilized to identify and select key images; this selection is further refined through human validation. Finally, the data is used to synthesize QAs of our dataset.
    \begin{enumerate}[label=$\triangleright$, leftmargin=10pt, itemsep=0pt]
    \item \textbf{Article data preprocessing.} Our process commences with the retrieval of scholarly articles from arXiv. For each crawled paper, essential metadata—including the title, abstract, and author information—is primarily acquired from the arXiv "abs" pages. Following metadata acquisition, the full HTML content of the article's page is fetched. Robust HTML parsing libraries, such as BeautifulSoup, are then employed to process this content. The images that are excessively small or that fail to load correctly are systematically filtered out. Furthermore, we leverage MLLM to identify key images in the articles. A detailed discussion of this process will be presented in subsequent section.
    \item \textbf{Questions, options and detailed generation.} For the Level 1 data, we employ a template-based question generation approach to elicit the title or first author's name associated with a given image. Simultaneously, we generate distractor options using either random selection or LLMs. For detailed data, we leverage LLMs to produce a concise summary of the paper based on its abstract. For the Level-2 data, We leverage prompts to elicit the generation of detailed questions pertaining to the image's context or abstract, high-level summaries. These questions are specifically designed to be unanswerable using only the image information or knowledge inherently derived from the image itself.
    \end{enumerate}
\subsection{Human annotation details}
\label{Appendix: human annotation details}
\begin{figure*}
    \centering
    \vspace{0em}
    \includegraphics[width=\linewidth]{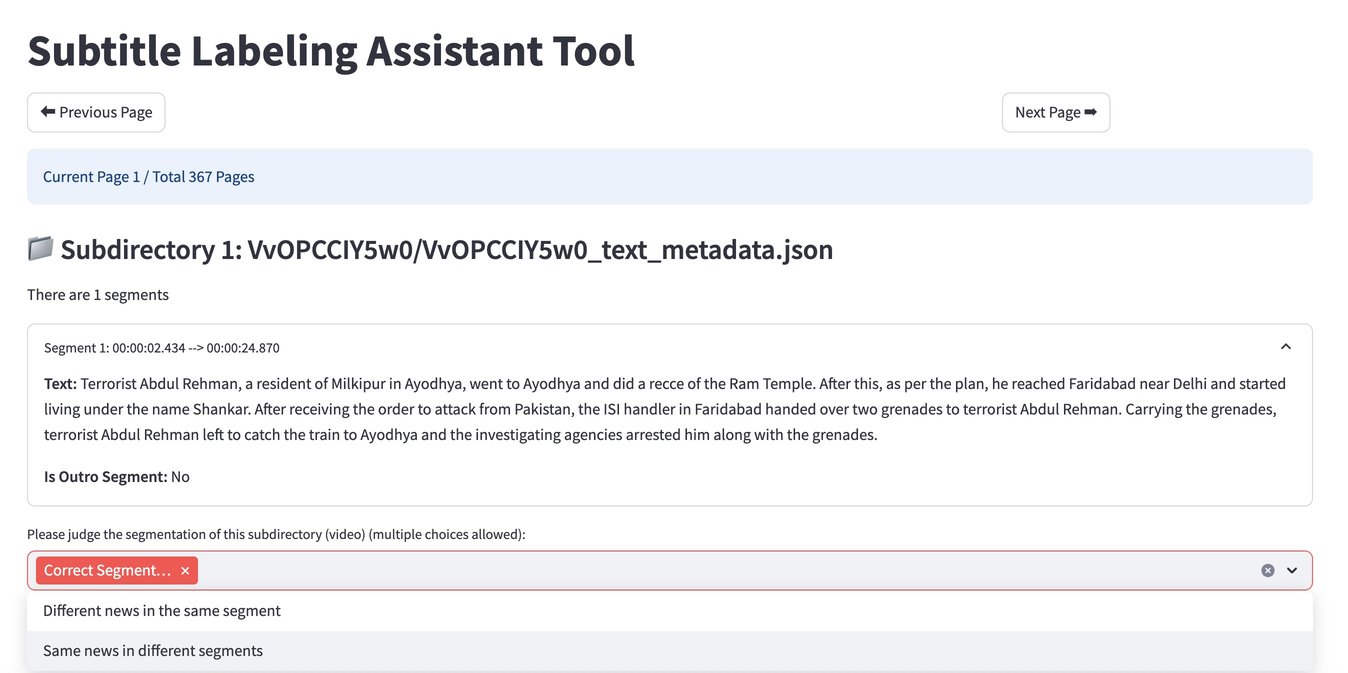}
    \vspace{0em}
    \caption{Human Annotation - YouTube Video - Parsing Subtitle by GPT4.1.}
    \label{fig:split_text}
    \vspace{0em}
\end{figure*}

\begin{figure*}
    \centering
    \vspace{0em}
    \includegraphics[width=\linewidth]{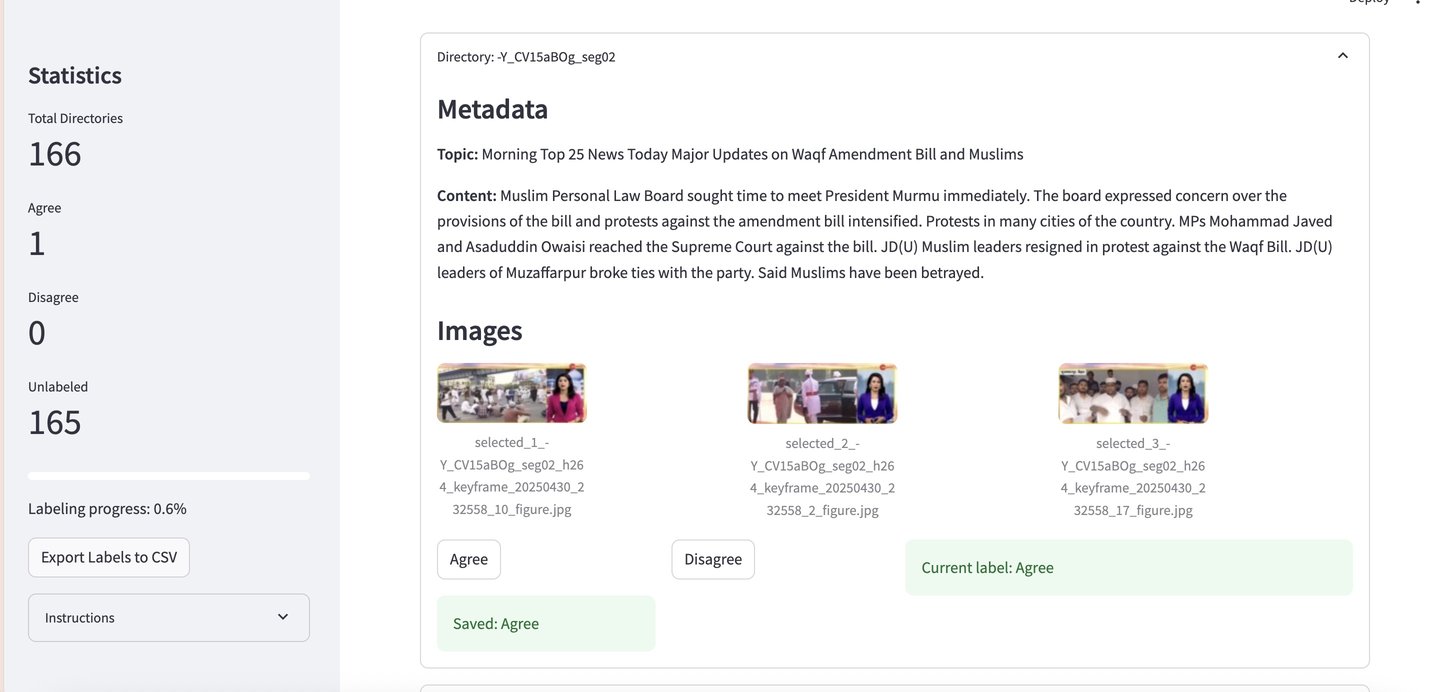}
    \vspace{0em}
    \caption{Human Annotation - YouTube Video - Selecting Images for Video Data.}
    \label{fig:select_image}
    \vspace{0em}
\end{figure*}

\begin{figure*}
    \centering
    \vspace{0em}
    \includegraphics[width=\linewidth]{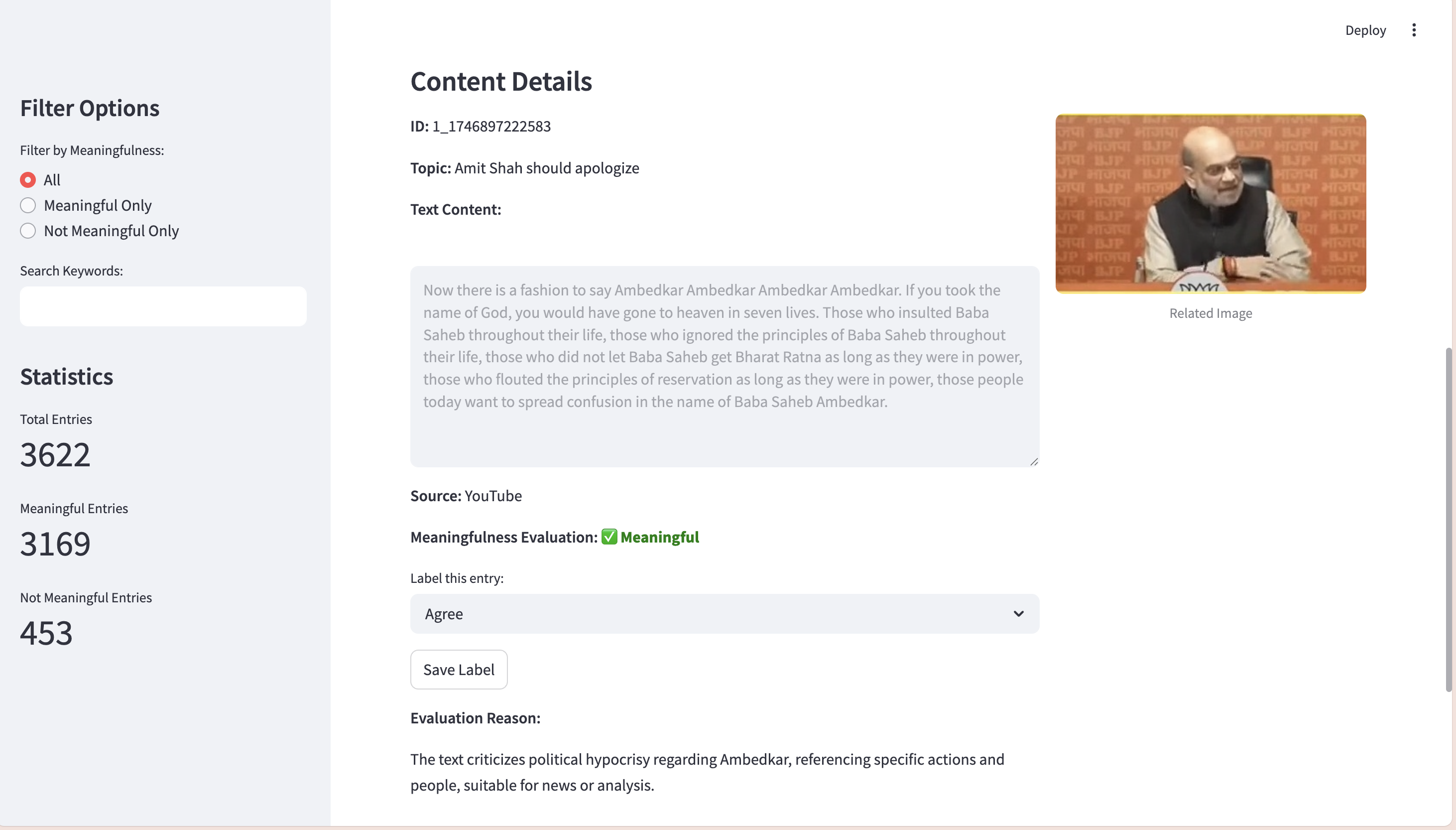}
    \vspace{0em}
    \caption{Human Annotation - Youtube Video - Evaluating Text for QA Generating.}
    \label{fig:evaluate_text}
    \vspace{0em}
\end{figure*}

\begin{figure*}
    \centering
    \vspace{0em}
    \includegraphics[width=\linewidth]{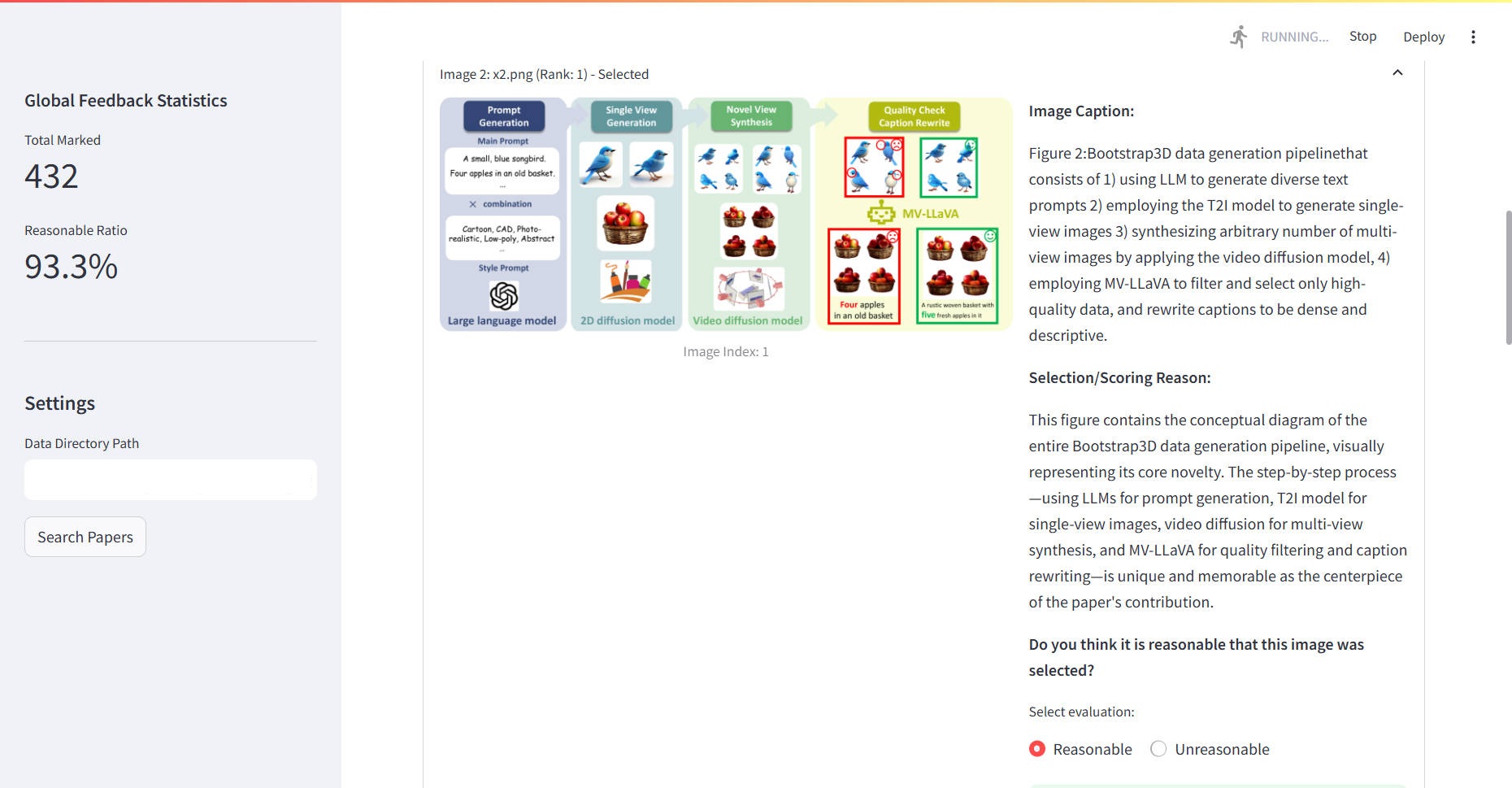}
    \vspace{0em}
    \caption{Human Annotation - Academic Paper - Selecting key images for academic papers.}
    \label{fig:academic}
    \vspace{0em}
\end{figure*}

\begin{figure*}
    \centering
    \vspace{0em}
    \includegraphics[width=\linewidth]{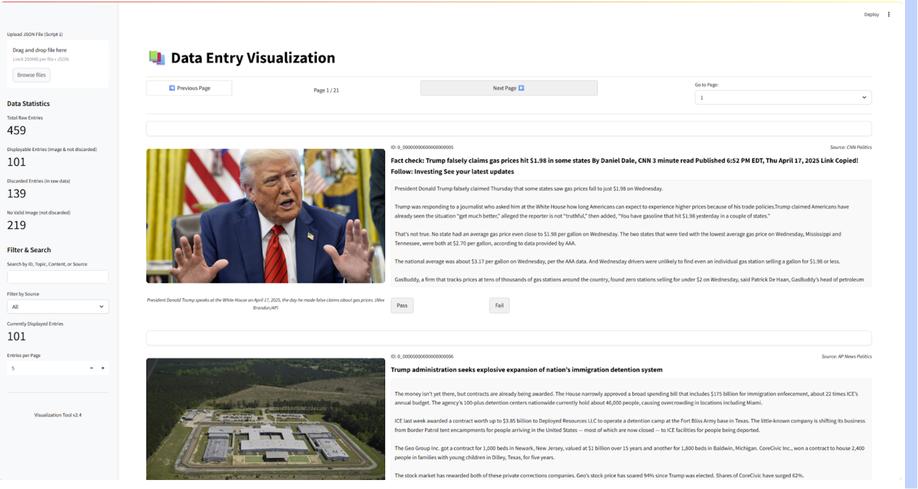}
    \vspace{0em}
    \caption{Human Annotation - News Article - Image filter for news data.}
    \label{fig:newsimg}
    \vspace{0em}
\end{figure*}

\begin{figure*}
    \centering
    \vspace{0em}
    \includegraphics[width=\linewidth]{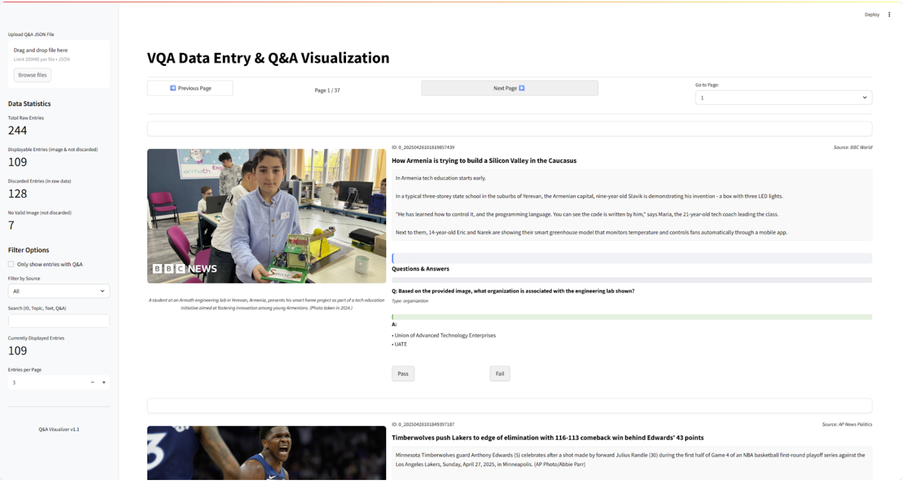}
    \vspace{0em}
    \caption{Human Annotation - Level-1 QA Pairs generation.}
    \label{fig:l1g}
    \vspace{0em}
\end{figure*}

\begin{figure*}
    \centering
    \vspace{0em}
    \includegraphics[width=\linewidth]{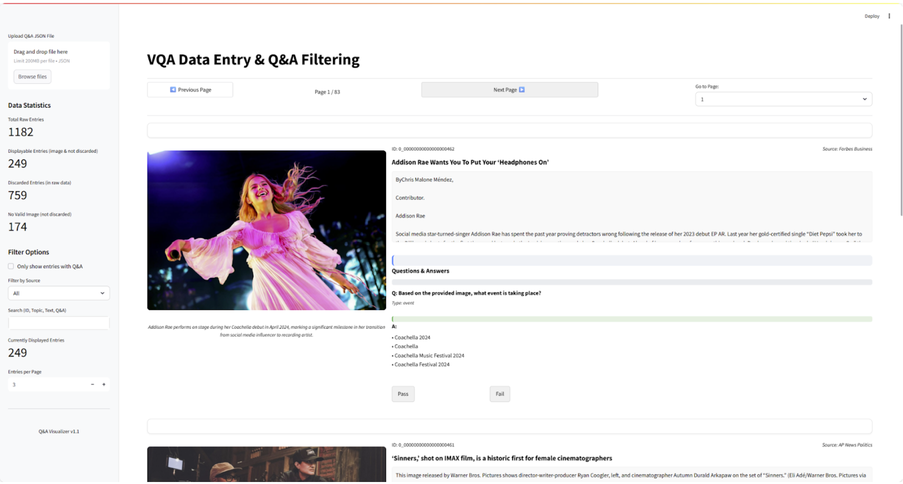}
    \vspace{0em}
    \caption{Human Annotation - Level-1 QA Pairs filter.}
    \label{fig:l1f}
    \vspace{0em}
\end{figure*}

\begin{figure*}
    \centering
    \vspace{0em}
    \includegraphics[width=\linewidth]{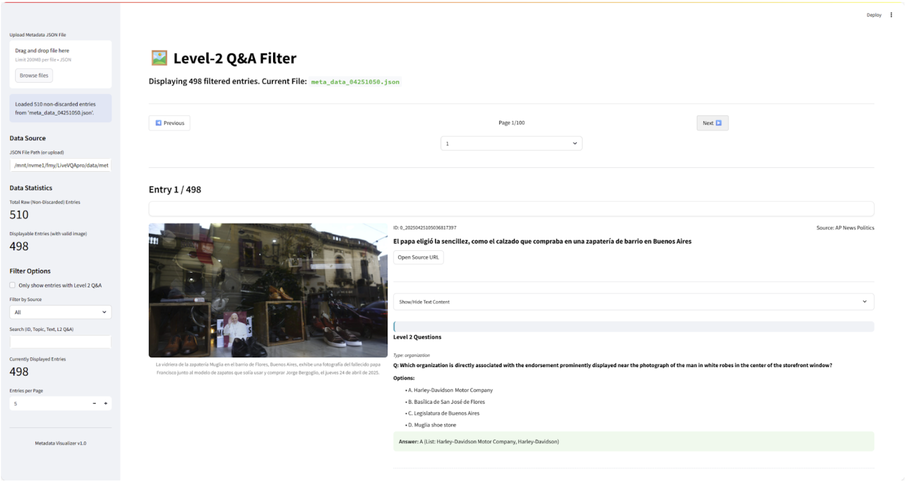}
    \vspace{0em}
    \caption{Human Annotation - Level-2 QA Pairs filter.}
    \label{fig:l2f}
    \vspace{0em}
\end{figure*}

\begin{figure*}
    \centering
    \vspace{0em}
    \includegraphics[width=\linewidth]{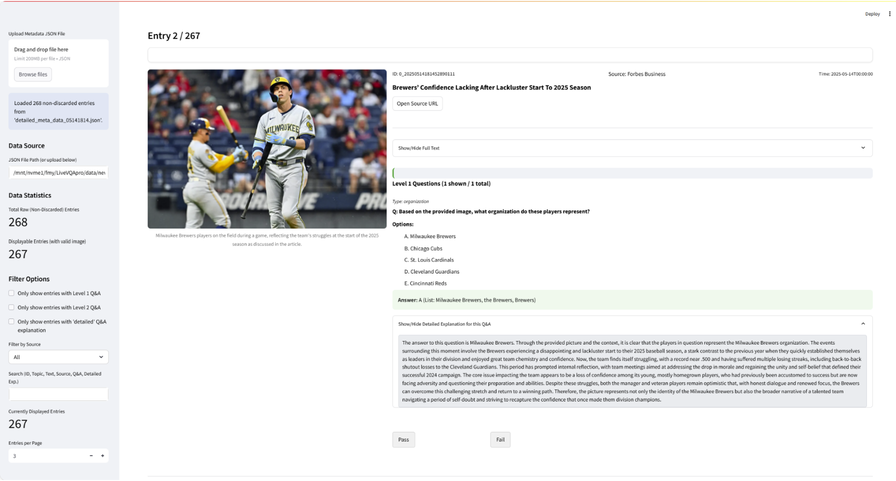}
    \vspace{0em}
    \caption{Detailed explanation of Level-1 QA Pairs and reasoning steps of Level-2 QA Pairs.}
    \label{fig:detailandreasoning}
    \vspace{0em}
\end{figure*}

\textbf{Every experiment using LLM/MLLM is validated with human annotated ground truth and agreement. We provide detailed instructions and annotation environments.} The annotation is conducted by 4 authors of this paper independently. All the annotations are conducted under Streamlit\footnote{\url{https://streamlit.io/}}. As acknowledged, the diversity of annotators plays a crucial role in reducing bias and enhancing the reliability of the benchmark. These annotators have rich knowledge in this domain, with different genders, ages, and educational backgrounds. To ensure the annotators can proficiently mark the data, we provide them with detailed tutorials, teaching them how to evaluate model responses more objectively. Specifically, they are required to give judgments without bias like answer lengths, and certain names of the response. All process using LLM/MLLM are listed as follows:

\begin{enumerate}[label=$\triangleright$, leftmargin=10pt, itemsep=0pt]
    \item \textbf{YouTube video - subtitle parsing and event segmentation with GPT-4.1.}
     Subtitle files are parsed to extract word tokens based on timestamp tags. Since the text often lacks punctuation, we utilize GPT-4.1 to reconstruct proper sentence boundaries and restore punctuation. Subsequently, GPT-4.1 segments these processed subtitles into coherent event segments. This segmentation is guided by both textual and temporal signals: short caption intervals usually indicate topical continuity, whereas longer pauses imply topic shifts. Segment boundaries are only introduced when a clear topical change is detected. To assess the structural appropriateness of text segmented by GPT-4.1, we develop an annotation tool in Figure~\ref{fig:split_text} with a passing rate of  98\%.

    \item \textbf{YouTube video - images selecting with GPT-4.1.} 
    After multiple rounds of preprocessing, each video segment still contains some images that exhibit weak relevance to the corresponding textual information.
    To address this, we employ the GPT-4.1 model to execute a fine-grained selection procedure, identifying and selecting the Top-K images with the strongest textual relevance from each segment for subsequent QA generation. To evaluate the appropriateness and effectiveness of GPT-4.1's image selection strategy, we design an annotation tool in Figure \ref{fig:select_image}. Manual evaluations conducted using this tool reveals that the images selected through this method achieves a 96\% passing rate.

    \item \textbf{YouTube video - text evaulating before generating QAs with GPT-4.1.}
    During the processing of textual information from videos, we observe that some videos contain sparse textual content, such as dialogues lacking specific details. These texts are prone to inducing hallucinations in large language models when used for generating QA pairs. To mitigate this issue, we employ GPT-4.1 to effectively filter out these low-information-density texts. Furthermore, to evaluate the rationale and accuracy of GPT-4.1's filtering strategy, we design a dedicated annotation tool in Figure\ref{fig:evaluate_text}. Manual assessments based on this tool demonstrated that GPT-4.1 achieved a 99\% accuracy rate in identifying and filtering out such unsuitable texts.

    \item \textbf{Academic paper -  key image selection with GPT-4.1.}
    We leverage GPT-4.1 to identify the key images in the articles. The key images are those that uniquely identifies a specific research paper, distinguishing it from other publications. The prompt is available at Figure\ref{fig:prompt_for_image_selection-1}, which shows our preference on image selection. To validate whether the model's selection is reasonable, we developed a labeling interface using Streamlit in Figure\ref{fig:academic}. Manual assessments demonstrate that GPT-4.1 achieved a 94\% accuracy rate.

    \item \textbf{News article - image filter with GPT-4.1.}  
    we retain only images clearly depicting current social topics and discard static objects, logos, duplicates, generic or decorative visuals, or any image lacking unique insight into the article’s core subject. Each kept image must add essential information unobtainable from text alone and be indispensable to understanding the story. Generate factual captions using only explicitly stated details. We employ GPT-4.1 to finish this job and design the annotation tool in Figure\ref{fig:newsimg}. Manual assessments demonstrate that GPT-4.1 achieved a 96\% accuracy rate.

    \item \textbf{Level-1 question generation with GPT-4.1.}  
    We generate basic QA pairs beginning “Based on the provided image,” whose answer is explicitly stated in the article. The question must make sense with the image alone, avoid celebrities, counts, logos or text cues, aise. We employ GPT-4.1 to finish this job and design the annotation tool in Figure\ref{fig:l1g}. Manual assessments show GPT-4.1 achieved a 93\% accuracy rate.

    \item \textbf{Level-1 question filtering with GPT-4.1.}  
     We discard articles lacking meaningful context (e.g., only noting a broadcast question). We always keep questions whose answers are non–major‐figure names, specific named locations, named events, specific products, or organizations. We employ GPT-4.1 to remove any unqualified QA pairs and use an annotation tool to evaluate. The annotation tool is shown in Figure~\ref{fig:l1f}. The accuracy of GPT-4.1 is 97\%.

     \item \textbf{Level-2 question generation with GPT-4.1.}  
      We construct multi-hop QA pairs and restrict to one of seven categories. We design decoys using partial truths, misconceptions or misleading links. We use GPT-4.1 to design the QA pairs and use the annotation tool in Figure~\ref{l2g}. The accuracy of GPT-4.1 is 98\%.

      \item \textbf{Level-2 Question filtering with GPT-4.1.} Some questions may be too hard and can't be verified. So We keep the question only if the correct answer can be obtained after combining the information of the news article itself, the picture information, and the title information. Otherwise, the question that cannot be answered will be discarded. We use GPT-4.1 to filter this kind of QA pairs and evaluate it using the annotation tool in Figure\ref{fig:l2f} to evaluate it performance. The accuracy of GPT-4.1 is 99\%.

      \item \textbf{Detailed and reasoning answer generation with GPT-4.1} We generate answers including detailed reasoning processes for subsequent training. We first ask the model to output the answer, then output information about the event corresponding to the image, and finally reason about the question. We use GPT-4.1 to do this and evaluate it by the annotation tool shown in Figure\ref{fig:detailandreasoning}. The results shows that GPT-4.1 achieve a 99\% accurate rate.

\end{enumerate}

\begin{figure*}
    \centering
    \vspace{0em}
    \includegraphics[width=\linewidth]{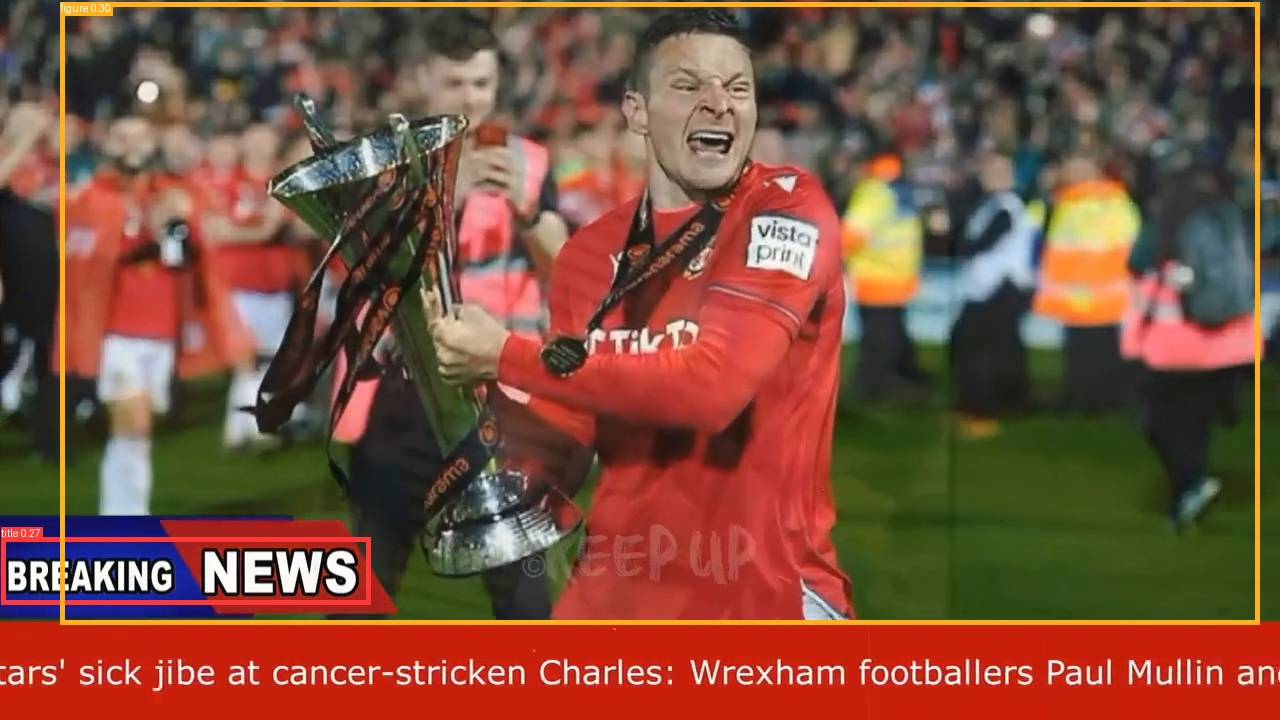}
    \vspace{0em}
    \caption{Preprocessing - YouTube Video - DocLayout-YOLO.}
    \label{fig:layout}
    \vspace{0em}
\end{figure*}

\begin{figure*}
    \centering
    \vspace{0em}
    \includegraphics[width=\linewidth]{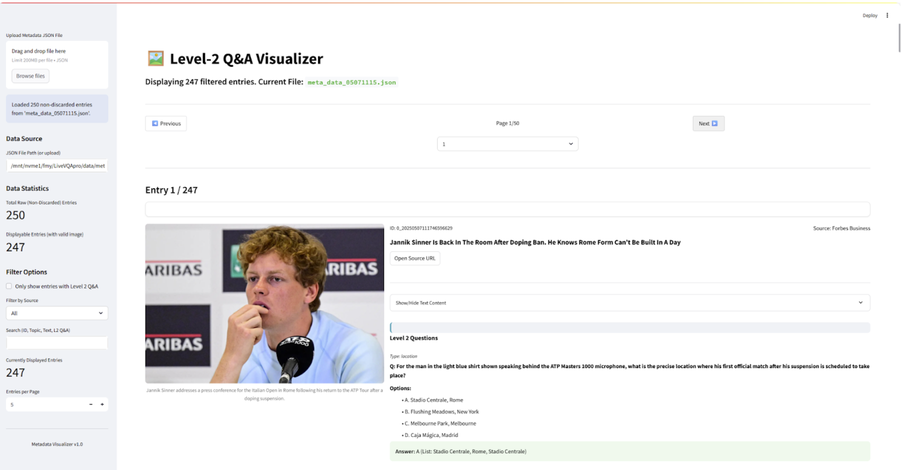}
    \vspace{0em}
    \caption{Human Annotation - Level-2 QA Pairs generation.}
    \label{fig:l2g}
    \vspace{0em}
\end{figure*}

\section{Detailed experimental setups}
\label{Appendix: detailed experiment setups}
\header{Models.}
% The detailed information of all models benchmarked in the experiments of Section 3 is listed in Table \ref{tab:judge_models}. The models we have trained in the experiments of Section 4 are listed in Table \ref{tab:judge_models} too.
The detailed specifications of all models evaluated in the experiments of Section 3 are provided in Table \ref{tab:judge_models}. The models trained for the experiments in Section 4 are also included in Table \ref{tab:judge_models}.

\begin{table*}[h]
    \small
    \centering
    \setlength{\tabcolsep}{2pt} % Default value: 6pt
    \renewcommand{\arraystretch}{1.2}
    \caption{Model names, Creators, Version, Access Time, License, and their using purpose.}
    \vspace{-3pt}
    \scalebox{0.82}{
    \begin{tabular}{l|ccccc}
    \toprule[1pt]
        \textbf{Model} & \textbf{Creator} & \textbf{Version} & \textbf{Knowledge Cutoff} & \textbf{License} & \textbf{Purpose}\\
        \midrule
    \rowcolor[RGB]{231,233,238}    \textbf{GPT-o3} & OpenAI & o3-2025-04-16 & 2024.6 & Proprietary & Experiment 1\\
         \textbf{GPT-4.1} & OpenAI & gpt-4.1-2025-04-14 & 2024.6 & Proprietary & Experiment 1\\
    \rowcolor[RGB]{231,233,238}    \textbf{GPT-4.1-mini} & OpenAI & gpt-4.1-mini-2025-04-14 & 2024.6 & Proprietary & Experiment 1\\
    \textbf{GPT-4.1-nano} & OpenAI & gpt-4.1-2025-04-14 & 2024.6 & Proprietary & Experiment 1\\
    \rowcolor[RGB]{231,233,238}    \textbf{GPT-4o} & OpenAI & gpt-4o-2024-08-06 & 2023.10 & Proprietary & Experiment 1\\
         \textbf{Gemini-2.5-Flash} & Google & gemini-2.5-flash-preview-04-17 & 2025.1 & Proprietary & Experiment 1\\
    \rowcolor[RGB]{231,233,238}    \textbf{Gemini-2.5-Pro} & Google & gemini-2.5-pro-preview-05-06 & 2025.1 & Proprietary & Experiment 1\\
         \textbf{Claude-3.7-Sonnet} & Anthropic & Claude-3.7-Sonnet & 2024.10 & Proprietary & Experiment 1\\
     \textbf{Gemma3-4b/12b/27b-it} & Google & Gemma3 & 2024.8 & Open-source & Experiment 1\\
    \rowcolor[RGB]{231,233,238}      \textbf{Llama-4-Scout-17B-16E-Instruct} & Meta & Llama-4 & 2024.08 & Open-source & Experiment 1\\
    \rowcolor[RGB]{231,233,238}    \textbf{Qwen2.5-VL-3B/7B/32B} & Alibaba & Qwen2.5 & Unknown & Open-source & Experiment 1 \& 2\\
        \bottomrule[1pt]
    \end{tabular}
    }
    \label{tab:judge_models}
\end{table*}

\header{Metric: Guessing strategy and F-score.} Originate from \citet{wei2024measuring}, while F-score is a good metric in some ways, the issue with it is that it incentivizes the model to always guess when it is at least 50\% sure that it can get the correct answer. To understand why this is the case, consider the following expression for the F-score:
\[
F\mbox{-}score
= \frac{2}{\frac{c + i}{c} + \frac{c + i + n}{c}}
= \frac{2c}{2c + 2i + n},
\]
where:
\begin{itemize}[leftmargin=10pt, itemsep=0pt]
  \item $c$ is the number of correct answers,
  \item $i$ is the number of incorrect answers, and
  \item $n$ is the number of non-answered questions.
\end{itemize}

If you have a greater than $\frac{1}{2}$ chance of being correct, your expected score from guessing is better than the score from not guessing, regardless of the specific values for $c$, $i$, and $n$. This is because the following inequality always holds:
\[
\frac{2c + 1}{2c + 2i + n + 2}
\;>\;
\frac{2c}{2c + 2i + n + 1}.
\]
The left-hand side represents the expected F-score from guessing, assuming a 50/50 chance of correctness, while the right-hand side is the score from not answering the additional question. Since the denominators are adjusted similarly whether the guess is correct or incorrect, guessing with a probability $>\frac{1}{2}$ yields a better score.

\header{Model fine-tuning details.} We choose the pre-trained Qwen2.5-VL-3B/7B-Instruct as our training models. We employ LoRA for fine-tuning while keeping the visual encoder frozen. The hyperparameters we used during fine-tuning are listed in Table \ref{tab:hyperparameters}. We conduct mixed-precision training on 8*A800 GPUs. All training data is derived from our curated dataset \dataset. 

\begin{table}[h!]
    \centering
    \caption{Overview of the hyperparameters.}
    \label{tab:hyperparameters}
    \begin{tabular}{l|c}
        \toprule
        \textbf{Hyperparameter} & \textbf{Value}  \\
        \midrule
        lr  & 1e-4 \\
        global batch size & 128(for 7B)/256(for 3B)  \\
        epoch & 2.0  \\
        optimizer & AdamW \\
        \bottomrule
    \end{tabular}
\end{table}

\begin{figure*}[!h]
    \centering
    \vspace{-1.5em}
    \includegraphics[width=0.5\linewidth]{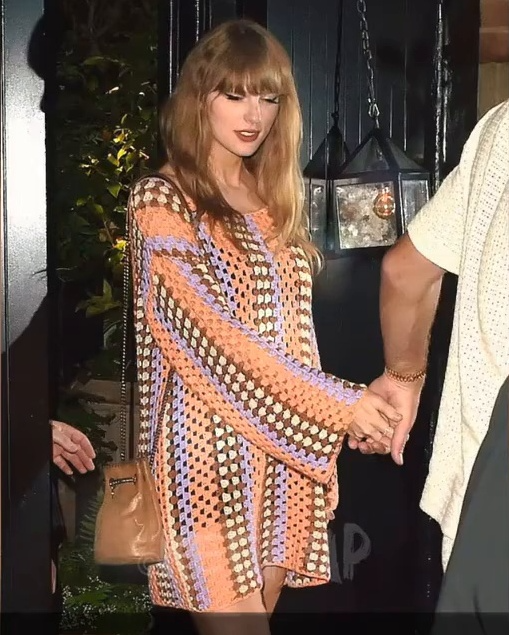}
    \caption{ This is a normal picture of the famous singer Taylor Swift, but the NSFW detector considered it to be Not Safe For Work. This is a completely wrong judgment. }
    \label{fig:nsfw}
    \vspace{-0.5em}
\end{figure*}

\subsection{NSFW image detection}
We utilize an NSFW detector\footnote{\url{https://huggingface.co/Falconsai/nsfw_image_detection}} for our dataset quality validation. In our dataset, 152 images are regarded as Not Safe For Work(NSFW). But all of them are wrong judgements. For example, in Figure \ref{fig:nsfw}, a normal picture of the famous singer Taylor Swift is considered as NSFW, which is completely unreasonable.

\section{Additional experimental results}
\label{appendix: additional results}
\header{Implementation challenges and engine improvements.} During the reproduction and deployment of the MMSearch engine, we encountered a number of practical challenges and implemented several targeted improvements. First, in terms of environment configuration, we observed that multiple multimodal models (\emph{\emph{e.g.}}, Qwen and LLaVA) have incompatible dependencies and must be installed in separate virtual environments to avoid conflicts.

Second, while implementing the web search module, we faced issues with frequent access being flagged as bot activity, which triggered CAPTCHA verification. This blocked page retrieval and interfered with both requery and rerank stages.

Moreover, prompt design proved critical in the multimodal reasoning chain. If the model in Stage~1 fails to extract valid information from the input image, it generates an uninformative requery, which propagates errors downstream. We also observed cases where, despite having relevant screenshots, the model selected irrelevant web pages during rerank (Stage~2), degrading performance in the summarization stage (Stage~3).

To mitigate these issues, we implemented the following strategies: (1) If Stage~1 yields no valid information from the image, the requery defaults to the original query, avoiding error amplification; (2) If the retrieved screenshot is a CAPTCHA page, the system skips it directly to ensure robustness; (3) If Stage~3 still fails to produce valid search-based content, we fallback to directly querying the model with the image and original question. These improvements significantly enhance the system's stability and overall answer quality, particularly in complex visual-language scenarios.

\textbf{Level 2 is harder than level 1 for image context understanding. } It is apparent from Table \ref{tab:model_difficulty_datasource} that the model's performance on Level 2 problems exhibits a significant decline compared to its performance on Level 1 problems except for arxiv subset. This result is part of anticipated, as the problems at Level 2 need deeper reasoning capacity. The improved performace on Level 2 questions in the arXiv dataset may be attributed to the abundance of visual information from key images, enabling the model to directly answer questions based on image content without necessitating the localization of relevant papers.

\begin{table}[!t] 
\centering 
\caption{Performance on detailed categories in Video subset. Visual searching framework dramastically enhance model's performance on Level 2 questions.}
\setlength{\tabcolsep}{3pt} % Default value: 6pt
\label{tab: video detailed category}
\resizebox{\textwidth}{!}{
\begin{tabular}{l|cccccc|cccccccc} \toprule[1.5pt] \multirow{2}{*}{Model} & \multicolumn{6}{c|}{Level 1} & \multicolumn{8}{c}{Level 2} \\
& Loc. & Per. & Org. & Eve. & Obj. & Avg. & Loc. & Per. & Org. & Time & Cou. & Rea. & Eve. & Avg. \\ 
\midrule \multicolumn{15}{c}{\textbf{\textit{w.o.} Search}} \\
\midrule GPT-4.1 & 26.58 & 8.33 & 40.85 & 7.77 & 32.23 & 22.00 & \underline{8.51} & 3.45 & 5.56 & 6.32 & 11.20 & 5.65 & 4.55 & 7.20 \\ 
GPT-4.1-mini & 21.52 & 13.54 & 30.99 & 4.85 & 30.58 & 19.60 & 2.13 & 3.45 & \textbf{12.96} & 6.32 & 15.20 & 3.23 & 4.55 & 7.80 \\ 
GPT-4.1-nano & 15.19 & 1.04 & 28.17 & 4.85 & 19.01 & 13.00 & 0.00 & 0.00 & 5.56 & 6.32 & 14.40 & 2.42 & 0.00 & 6.00 \\ 
Gemini-2.5-Flash & 18.99 & \underline{27.08} & 29.58 & 4.85 & 18.18 & 18.40 & 0.00 & 3.45 & 1.85 & 4.21 & 11.20 & 0.81 & 4.55 & 4.40 \\
Gemini-2.5-Pro & 8.86 & 25.00 & 32.39 & 6.80 & 19.01 & 17.40 & 0.00 & 0.00 & 1.85 & 2.11 & 5.60 & 1.61 & 0.00 & 2.40 \\ 
Gemma-3-27B-IT & 13.92 & 14.58 & 33.80 & 3.88 & 21.49 & 16.40 & 0.00 & 0.00 & 5.56 & 4.21 & 10.40 & 1.61 & 4.55 & 4.60 \\ 
Claude-3.7-Sonnet & 18.99 & 7.29 & 29.58 & 6.80 & 23.97 & 16.40 & 2.13 & 0.00 & 1.85 & 4.21 & 7.20 & 4.84 & 4.55 & 4.40 \\ 
Qwen-2.5-VL-7B & 12.66 & 10.42 & 25.35 & 4.85 & 16.53 & 13.40 & 2.13 & 0.00 & 5.56 & 3.16 & 14.40 & 1.61 & 0.00 & 5.40 \\ 
Qwen-2.5-VL-32B & 16.46 & 10.42 & 32.39 & 4.85 & 22.31 & 16.40 & 0.00 & 0.00 & 5.56 & 6.32 & 9.60 & 4.84 & 4.55 & 5.60 \\ 
Qwen-2.5-VL-72B & 10.13 & 3.12 & 18.31 & 1.94 & 14.88 & 9.40 & 0.00 & 0.00 & 7.41 & 3.16 & 5.60 & 2.42 & 4.55 & 3.60 \\ 
Llama-4-Scout & 16.46 & 13.54 & 26.76 & 7.77 & 20.66 & 16.40 & 2.13 & 0.00 & 7.41 & 4.21 & 10.40 & 1.61 & 4.55 & 5.00 \\ 
Llama-4-Maverick & 18.99 & 14.58 & 38.03 & 8.74 & 20.66 & 19.00 & 2.13 & 3.45 & 3.70 & 4.21 & 15.20 & 2.42 & 0.00 & 6.00 \\
\midrule \multicolumn{15}{c}{\textbf{\textit{w.} Text Search}} \\
\midrule GPT-4.1 & 13.92 & 6.25 & 30.05 & 3.56 & 22.59 & 14.60 & 2.84 & 0.00 & 3.09 & 3.86 & 6.67 & 2.42 & 3.03 & 3.73 \\ 
Gemini-2.5-Pro & 1.69 & 1.39 & 19.72 & 2.91 & 8.54 & 6.53 & 0.00 & 0.00 & 0.62 & 1.40 & 3.20 & 0.00 & 1.52 & 1.20 \\
Claude-3.7-Sonnet & 8.02 & 4.17 & 14.55 & 2.59 & 12.95 & 8.33 & 1.42 & 0.00 & 1.23 & 1.40 & 3.73 & 0.54 & 0.00 & 1.60 \\ 
\midrule
\multicolumn{15}{c}{\textbf{\textit{w.} Native Image Search}} \\
    \midrule 
    GPT-o3 & \textbf{37.97} & 19.79 & \underline{43.66}  & \textbf{22.33}& \textbf{46.28} & \textbf{33.60} & \underline{8.51} & \underline{10.34} & \textbf{12.96}  & \textbf{11.58}  & \textbf{29.60}& \textbf{25.00} & \textbf{18.18}& \textbf{19.40} \\
    \midrule
    \multicolumn{15}{c}{\textbf{\textit{w.} MM-Search} \citep{jiang2024mmsearch}} \\
    \midrule
    GPT-4.1 & \underline{29.11} & \textbf{31.58} & \textbf{49.30} & \underline{21.36} & \underline{38.84} & \underline{33.00} & \textbf{13.68} & \textbf{17.02} & \underline{10.34} & \underline{11.11} & \underline{26.40} & \underline{9.68} & 4.55 & \underline{15.20} \\
\bottomrule[1.5pt] 
\end{tabular}}
\end{table}

\section{Prompt}

In our synthetic dataset, LLM is used extensively for question-answer pair generation and image filtering, as well as a small amount of judgment work. Here, we show all our prompts.

\begin{tcolorbox}[breakable, enhanced,skin=widget, attach boxed title to top center={yshift=-3mm,yshifttext=-1mm},boxrule=0.9pt, 
        colback=gray!00,colframe=black!50,colbacktitle=gray,
        title=Prompt: Image Filtering,
        boxed title style={size=small,colframe=gray}
    ]
% Prompt converted to LaTeX (standard `` '' quotation marks)

You are an expert‑level image analyst and meticulous news‑editor assistant. Your primary objective is to analyze and optimize the images associated with news articles according to the following structured tasks, applying \textbf{rigorous standards}, especially for relevance:\par

\begin{enumerate}
  \item \textbf{Very important: KEEP ONLY IMAGES THAT CARRY CLEAR, CURRENT SOCIAL RELEVANCE.}\par
        Retain an image only if it directly illustrates an ongoing public issue, breaking news story, cultural trend, policy discussion, or other time‑sensitive social topic. Immediately delete any image that shows nothing more than a static object or brand (\emph{e.g.}\ a bowl of fruit salad, a smartphone, a corporate logo) without explicit social or temporal context. \textbf{NO EXCEPTIONS.}

  \item \textbf{Identify Duplicate Images:}\par
        Analyze the provided images for visual duplication or near‑duplication within the context of the \emph{same} article. Mark identified duplicates for removal.

  \item \textbf{Evaluate Image Relevance with EXTREME Scrutiny (Minimalist Standard):}\par
        Assess each unique image’s relevance against the \textbf{absolute core narrative, pivotal moments, key individuals directly involved, and essential locations} described in the article’s \textbf{Title} and \textbf{Text}. Apply an \textbf{exceptionally critical, near‑zero‑tolerance standard} for this evaluation. Mark \emph{any} image not meeting these stringent criteria as irrelevant for removal.
        \begin{itemize}
          \item \textbf{Fundamental Question (Default = Exclude):}\par
                Does this image provide \textbf{unique, indispensable visual insight} into a \textbf{critical aspect} of the \emph{specific event or subject} being reported, offering information the text \textbf{cannot adequately convey on its own}? Assume the image is irrelevant unless proven otherwise by meeting \emph{all} points below.
          \item \textbf{Strictly Exclude (Non‑Exhaustive List — Apply Principle Broadly):}
                \begin{itemize}
                  \item \emph{Anything} generic, decorative, illustrative without specific factual grounding, or abstract.
                  \item Images related only tangentially, peripherally, metaphorically, or to background/contextual information (even if factually correct). \textbf{Focus solely on the central action/subject.}
                  \item Visuals connected to secondary details, minor figures, historical context not part of the main event, or general scene‑setting.
                  \item \emph{Any} image where the link to the article’s absolute core requires \emph{any} inference, assumption, or ambiguity. The connection must be \textbf{immediate, explicit, and undeniable}.
                  \item Images that, while factually related, primarily duplicate information easily stated in the text or caption, or offer minimal unique visual value pertinent to the \emph{specific nucleus} of the story (\emph{e.g.}\ generic building exteriors, standard portraits unrelated to the article’s specific action, maps of widely known locations).
                  \item Images whose primary value relies heavily on the caption to establish relevance; the visual content itself must be intrinsically and powerfully relevant.
                \end{itemize}
          \item \textbf{Retain ONLY IF ALL Conditions Met (Exceptionally High Bar):}
                \begin{itemize}
                  \item The image provides \textbf{critical visual evidence or clarification} directly tied to the \textbf{absolute core claim or event} of the article.
                  \item The visual information presented is \textbf{unique} and \textbf{cannot be effectively substituted by text alone}.
                  \item Removing the image would create a \textbf{significant and demonstrable gap} in understanding the \emph{most crucial} aspects of the story for the reader.
                  \item The relevance is \textbf{patently obvious and requires zero explanation} beyond the image itself and the core article topic.
                \end{itemize}
          \item \textbf{Final Rule:}\par
                \textbf{The default stance is EXCLUSION.} Override to retain \emph{only} if the image unequivocally meets \emph{every single stringent criterion} above with \emph{absolute certainty} and demonstrably provides \emph{irreplaceable value}. If there is \emph{any doubt whatsoever}, mark as irrelevant.
        \end{itemize}

  \item \textbf{Enhance or Create Captions with STRICT FACTUAL ADHERENCE:}\par
        For each image that \textbf{passes the strict relevance filter} and will be kept, evaluate its existing caption (if provided). Enhance it or create a new one if missing or inadequate (``null''). Ensure all final captions are informative and follow these standards:
        \begin{itemize}
          \item \textbf{CRITICAL: Use ONLY verifiable information directly stated in the article’s Title and Text.} Do \textbf{not} invent, assume, extrapolate, or add \emph{any} details not explicitly mentioned in the provided content.
          \item Include specific details about \textbf{people, places, and events} depicted, but \textbf{only} if these details are clearly stated in the article text or title.
          \item Provide \textbf{context} that links the image to the article’s narrative using \textbf{only} information present in the article.
          \item Maintain a \textbf{professional, objective, and journalistic tone}.
          \item Mention \textbf{time and location} information \textbf{only} when explicitly stated in the article text or title. Never infer or guess time/location data that are not directly provided.
          \item If uncertain about any detail, omit it entirely rather than risk including inaccurate or invented information.
          \item Before finalizing each caption, verify \emph{every} piece of information against the article content to ensure it comes directly from the provided text.
        \end{itemize}
\end{enumerate}

{\textbf Input Article Information}
Please analyze the following news article and its images based on these inputs:

\textbf{Title:} \texttt{\{title\}}

\textbf{Text:} \texttt{\{text\}}

\textbf{Number of images:} \texttt{\{image\_count\}}

{\textbf Output Requirements}

You \textbf{MUST} respond exclusively in JSON format. Your entire response should be a single JSON object, starting with \texttt{\{} and ending with \texttt{\}}. Do \textbf{not} include any introductory text, explanations, or Markdown formatting outside the JSON structure.\par

The JSON object must strictly follow this structure:

. . .

\noindent Remember: indices in \texttt{duplicates\_identified} and \texttt{irrelevant\_identified} refer to the 0‑based position in the \emph{original} list of images provided in the input. The lists in \texttt{processed\_data} should contain information only for the images kept according to the \textbf{strict relevance evaluation}.

\end{tcolorbox}

\begin{tcolorbox}[
  enhanced,
  breakable,
  attach boxed title to top center={
    yshift=-3mm,
    yshifttext=-1mm
  },
  boxrule=0.9pt,
  colback=gray!00,
  colframe=black!50,
  colbacktitle=gray,
  breakable,
  title=Prompt: Key Image Selection for Academic Paper,
  boxed title style={size=small,colframe=gray}
]
\textbf{Objective:} Analyze the provided paper abstract and image captions to identify and rank figures that best serve as memorable "paper identifiers." A "paper identifier" image is one that is visually distinct, memorable, and highly specific to this paper's unique contributions, making it unlikely to appear in other papers. Seeing this image should ideally make someone think of \emph{this specific paper}.

\vspace{0.5em} % Add some vertical space

\textbf{Inputs You Will Receive:}
\begin{enumerate}[label=\arabic*.]
    \item Paper Abstract: A concise summary of the paper's research, methodology, and findings.
    \item Image Captions: A list of captions, each corresponding to an image within the paper.
\end{enumerate}

\vspace{0.5em} % Add some vertical space

\textbf{Your Task:}
\begin{enumerate}[label=\arabic*.]
    \item \textbf{Understand the Core Contributions:}
    \begin{itemize}
        \item Read the Abstract to identify the primary contributions, methodologies, specific datasets, key theoretical concepts, or highly distinct results presented in the paper. What makes this paper stand out?
    \end{itemize}
    \item \textbf{Evaluate Each Image Caption for Memorability and Uniqueness:}
    \begin{itemize}
        \item For every image caption provided:
        \begin{itemize}
            \item Assess how well the image (as described by its caption) visually represents the unique and memorable aspects identified from the abstract.
            \item Strongly prioritize images described as:
            \begin{itemize}
                \item Framework/Architectural Diagrams
                \item Conceptual Diagrams
                \item Striking or Unexpected Visualizations/Illustrations of Key Findings except for statistical visualizations
                \item Highly Distinctive Scientific Illustrations
                \item Flowcharts or Block Diagrams
            \end{itemize}
            \item Avoid giving high ranks to (these are typically \emph{not} memorable identifiers):
            \begin{itemize}
                \item Any Bars, Plots, Graphs, Maps and statistical visualizations are USELESS, you MUST NOT select them
                \item Images of People, Animals, or Objects that are not unique to the paper
                \item Image with rich text like summaries, challenges, conclusions or limitations
                \item Tables, Equations, Algorithm Boxes/Pseudocode presented as images
            \end{itemize}
        \end{itemize}
    \end{itemize}
    \item \textbf{Rank All Figures:}
    \begin{itemize}
        \item Create a ranking for \emph{all} provided figures based on their potential as memorable identifiers. The figure deemed the most unique and memorable identifier should be ranked first.
        \item For each figure, provide a brief reason for its rank, specifically addressing its uniqueness, memorability, and connection to the paper's core novelties.
    \end{itemize}
    \item \textbf{Recommend a Selection Count:}
    \begin{itemize}
        \item Based on your ranking, decide on a \texttt{recommended\_count} of figures (typically 1, 0-3) that you believe are the most effective and sufficient set of memorable identifiers for this paper.
        \item If all the images are not unique or memorable like statistical visualizations, be brave to recommend 0 images.
    \end{itemize}
    \item \textbf{Explain Recommendation for Count:}
    \begin{itemize}
        \item Provide a \texttt{selection\_reason} briefly explaining why you recommend selecting this particular number of figures.
    \end{itemize}
\end{enumerate}

\vspace{0.5em} % Add some vertical space

\textbf{Output Format:}

Produce a single JSON object with the following structure:

Return a JSON object with the ranking information:
\begin{itemize}
    \item \texttt{"ranking"}: (Array of Objects)
    \begin{itemize}
        \item \textbf{Description}: An array where each object represents a single ranked figure. The figures in this array should ideally be ordered according to their rank (e.g., from highest to lowest ranked).
        \item \textbf{Structure of each object in the array}:
        \begin{itemize}
            \item \texttt{"index"}: (Integer)
            \begin{itemize}
                \item \textbf{Description}: The 1-based index or unique identifier of the figure being ranked. This typically refers to the figure's position or ID from the input list of figures provided for evaluation.
                \item \textbf{Example Value}: \texttt{1}, \texttt{2}, \texttt{3}, etc.
            \end{itemize}
            \item \texttt{"reason"}: (String)
            \begin{itemize}
                \item \textbf{Description}: A concise explanation detailing why this specific figure was assigned its current rank. This should highlight the factors contributing to its position in the ranking.
                \item \textbf{Example Value}: \texttt{"High relevance to the topic and excellent visual clarity."}
            \end{itemize}
        \end{itemize}
    \end{itemize}
    \end{itemize}

The "ranking" array should contain ALL figures sorted by their value as paper identifiers, with the most valuable figure first.
\end{tcolorbox}

\begin{tcolorbox}[breakable, enhanced,attach boxed title to top center={yshift=-3mm,yshifttext=-1mm},boxrule=0.9pt, 
        colback=gray!00,colframe=black!50,colbacktitle=gray,
        title=Prompt: Level-1 QA Pairs Generation,
        boxed title style={size=small,colframe=gray}
    ]
% Prompt converted to LaTeX (standard `` '' quotation marks)
{\textbf Prompt 1: AI for Level 1 Multi-Hop Question Generation}

You are an AI assistant specialized in generating high-quality Level 1 multi-hop questions that require social knowledge to answer. Your task is to create image-and-text-based questions that focus on factual information rather than inference or reasoning.

Your generated question \textbf{MUST} follow these strict requirements:

\begin{enumerate}
  \item \textbf{Question format:} Always start with ``Based on the provided image, '' followed by a clear, concise question.
  \item \textbf{Answer source:} The answer \textbf{MUST} be explicitly findable in the provided text (not just inferrable).
  \item \textbf{Answer format:} The answer must be a short phrase or a few words (NOT a sentence or paragraph).
  \item \textbf{Question categories:} The question \textbf{MUST} belong to one of these categories ONLY:
  \begin{itemize}
    \item location (where something is happening)
    \item person (who is in the image, but avoid asking about very famous people like Trump or Musk)
    \item organization (which company, team, group, etc.)
    \item time (when something occurred)
    \item object (what specific item is shown)
    \item event (ONLY allowed to ask ``what event is taking place?'')
  \end{itemize}
  \item \textbf{Question simplicity:} The question must be concise and avoid revealing too many details from the article.
  \item \textbf{Required integration:} Question must relate to what can be seen in the image, while having an answer in the text.
  \item \textbf{Knowledge requirement:} The question should test knowledge that cannot be directly answered by computer vision alone.
\end{enumerate}

\textbf{CRUCIAL QUALITY CRITERIA - AVOID THESE COMMON ISSUES:}
\begin{enumerate}
  \item \textbf{FAMOUS FIGURES:} DO NOT create questions asking about extremely well-known figures (\emph{e.g.}, \textit{``who is this person?''} when Donald Trump is in the image). These are too obvious.
  \item \textbf{SPECIFIC ANSWERS ONLY:} Ensure answers are \textbf{HIGHLY SPECIFIC} and uniquely identifiable.
  \begin{itemize}
    \item BAD: \textit{``Designer sneakers''}, \textit{``high-end sneakers''}
    \item GOOD: \textit{``Nike Air Force 1''}, \textit{``Louis Vuitton Trainers''}
  \end{itemize}
  \item \textbf{TEMPORAL CONTEXT REQUIRED:} NEVER create questions about images that lack clear temporal context.
  \item \textbf{NO COUNTING QUESTIONS:} Never create questions asking to count objects in the image.
  \item \textbf{AVOID BOOK COVER QUESTIONS:} Don't ask about book covers with answers like \textit{``book cover''}, \textit{``memoir cover''}, or \textit{``book jacket''}.
  \item \textbf{NO VISIBLE TEXT ANSWERS:} Don't create questions whose answers appear as visible text in the image.
  \item \textbf{SPECIFIC LOCATIONS ONLY:} 
  \begin{itemize}
    \item BAD: \textit{``textile factory''}, \textit{``shopping mall''}
    \item GOOD: \textit{``Nike Factory in Vietnam''}, \textit{``Galeries Lafayette in Paris''}
  \end{itemize}
  \item \textbf{SPECIFIC EVENT IDENTIFIERS:}
  \begin{itemize}
    \item BAD: \textit{``stunt performance''}, \textit{``fashion show''}
    \item GOOD: \textit{``2023 Paris Fashion Week''}, \textit{``Black Lives Matter protest in Portland''}
  \end{itemize}
  \item \textbf{NO CHART DATA QUESTIONS:} Do not ask about visible chart or graph data.
  \item \textbf{COMPLETE CONTENT REQUIRED:} Ensure the topic has both questions and images.
  \item \textbf{SPECIFIC PEOPLE IDENTIFIERS:}
  \begin{itemize}
    \item BAD: \textit{``police officer''}, \textit{``protestor''}
    \item GOOD: \textit{``Emmanuel Macron''}, \textit{``Taylor Swift''}
  \end{itemize}
  \item \textbf{NO UNIVERSAL DESCRIPTIONS:}
  \begin{itemize}
    \item BAD: \textit{``car accident''}, \textit{``protest''}, \textit{``earthquake''}
    \item GOOD: \textit{``2023 California wildfires''}, \textit{``Yellow Vest protests in Paris''}
  \end{itemize}
  \item \textbf{ANSWERS MUST BE UNIQUE TO THE SPECIFIC EVENT:}
  \begin{itemize}
    \item BAD: \textit{``Anti-government protest''}
    \item GOOD: \textit{``2023 French pension reform protests''}
  \end{itemize}
  \item \textbf{AVOID ERROR PATTERN EXAMPLES:}
  \begin{itemize}
    \item ``Based on the provided image, who is speaking at the podium?'' $\rightarrow$ ``President Donald Trump''
    \item ``Based on the provided image, what type of footwear is shown?'' $\rightarrow$ ``Designer sneakers''
    \item ``Based on the provided image, what dish is being prepared?'' $\rightarrow$ ``Pizza''
    \item ``Based on the provided image, how many protesters are visible?'' $\rightarrow$ ``24''
    \item ``Based on the provided image, what is shown on the book cover?'' $\rightarrow$ ``Book jacket''
    \item ``Based on the provided image, what is the name of the memorial site where the graves of Zambia's 1993 national football team are located?'' (too specific)
    \item ``Based on the provided image, who is the CEO that announced the company's new AI strategy at the June conference?'' (reveals too many details)
  \end{itemize}
\end{enumerate}

\textbf*{EXAMPLES OF GOOD QUESTIONS (APPROPRIATE BALANCE):}
\begin{itemize}
  \item ``Based on the provided image, what is the location shown?''
  \item ``Based on the provided image, who is the person at the podium?''
  \item ``Based on the provided image, what organization does this logo represent?''
  \item ``Based on the provided image, what event is taking place?''
\end{itemize}

\textbf*{AVOID these types of questions:}
\begin{itemize}
  \item Questions about visible attributes (\emph{e.g.}, clothing color, number of people)
  \item Questions with ambiguous or subjective answers
  \item Questions that can be answered without social/factual knowledge
  \item Questions about extremely obvious information
  \item Questions whose answers are directly visible as text in the image
\end{itemize}

Please generate a Level 1 multi-hop question based on the following news article and image. This question should test social knowledge rather than just visual perception.

\textbf*{Input Information:}
\textbf{ARTICLE TITLE:} \texttt{\{title\}} \\
\textbf{ARTICLE TEXT:} \texttt{\{text\}} \\
\textbf{IMAGE PATH:} \texttt{\{img\_path\}} \\
\textbf{IMAGE URL:} \texttt{\{img\_url\}} \\
\textbf{IMAGE CAPTION:} \texttt{\{caption\}} \\
\textbf{Associated Types Info:} \texttt{\{used\_types\_info\}} \\
\textbf{Associated Questions Info:} \texttt{\{used\_questions\_info\}} \\

\textbf*{Core Generation REQUIREMENTS:}
\begin{enumerate}
  \item The question \textbf{MUST} start with ``Based on the provided image, ''
  \item The answer \textbf{MUST} be explicitly found in the article text
  \item The answer must be a short phrase or a few words (not a sentence)
  \item The question must belong to one of these categories only: \textit{location}, \textit{person}, \textit{organization}, \textit{time}, \textit{object}, or \textit{event}
  \item If asking about an event, the question must be ``what event is taking place?''
\end{enumerate}

\textbf*{CRITICAL QUALITY CONSTRAINTS (for Generation):}
\begin{enumerate}
  \item DO NOT ask about obvious public figures (\emph{e.g.}, ``who is this?'' for Donald Trump)
  \item ENSURE answers are specific and uniquely identifiable (\emph{e.g.}, ``Nike Factory in Vietnam'', not just ``factory'')
  \item DO NOT create questions for images lacking temporal context (\emph{e.g.}, food close-ups, generic product shots)
  \item NEVER include counting questions (``how many people/objects...'')
  \item AVOID book cover questions with generic answers like ``book jacket''
  \item DO NOT create questions whose answers are directly visible in the image as text/logos
  \item Location answers must be specific places, not generic types like ``shopping mall'' or ``clothing store''
  \item Event answers must be specific named events, not generic types like ``protest'' or ``fashion show''
  \item DO NOT ask about data already visible in charts or graphs
  \item People answers must be specific named individuals, not job roles like ``police officer'' or ``doctor''
\end{enumerate}

\textbf*{CRITICAL CONSTRAINTS (Final Review):}
\begin{enumerate}
  \item Create a \textbf{SIMPLE, CONCISE} question that does \textbf{NOT} reveal too much information from the article
  \item DO NOT include specific details, names, dates or unique information from the article in your question
  \item The question should work as a standalone with just the image (we are creating a benchmark where users will only see the image and question)
  \item Focus on what can be visually identified in the image, while ensuring the answer is in the text
  \item Avoid questions that reveal the answer or provide too much context about the subject
  \item \textbf{VERY IMPORTANT:} Your question \textbf{MUST} be substantially different from questions already generated for other images in this topic
  \item DO NOT ask about the same people, objects, or locations that were already asked about in previous questions for this topic
\end{enumerate}

\textbf*{Example Question Phrasing:}
\textbf{BAD EXAMPLE}

``Based on the provided image, what is the name of the memorial site where the graves of Zambia's 1993 national football team are located?''

\textbf{GOOD EXAMPLE}

``Based on the provided image, what is this memorial site called?''

\textbf*{OUTPUT FORMAT (JSON):}

Please provide your response in the following \textbf{JSON format}:

\begin{lstlisting}[basicstyle=\ttfamily\footnotesize, breaklines=true, showstringspaces=false]
{
  "question": "Based on the provided image, [your simple, concise question]?",
  "question_type": "[category: location/person/organization/time/object/event]",
  "options": [
    "A. [option A]",
    "B. [option B]",
    "C. [option C]",
    "D. [option D]",
    "E. [option E]"  
  ],
  "Ground_Truth": "[correct letter, \emph{e.g.}, A](Please pay attention, you should randomly choose the correct answer position, it can be A~E!!!!!)",
  "Ground_Truth_List": ["[correct answer]", "[alternative phrasing 1]", "[alternative phrasing 2]", ...]
}
\end{lstlisting}

\textbf*{IMPORTANT FORMAT INSTRUCTIONS (for JSON Output):}
\begin{itemize}
  \item Include 3--5 multiple-choice options, with one being the correct answer. The position of the correct answer can be randomized (A--E).
  \item Make incorrect options plausible and challenging to distinguish.
  \item The \texttt{Ground\_Truth\_List} should include multiple valid phrasings of the answer (up to 10).
  \item If you cannot create a suitable question, return: \verb|{"error": "Unable to generate an appropriate question"}|
  \item Ensure all content is in English.
\end{itemize}

{\textbf Prompt 2: AI for Level-1 QA Filtering}
You are a specialized AI assistant tasked with filtering news-related visual questions. Each time you filter, check the following rules! \textbf{Your every action must be based on these rules:}

\textbf*{Auto-Delete Criteria (MUST DISCARD)}
Too short news must be deleted. You \textbf{must} delete this kind of news.\\
\textit{Example:} ``\texttt{<img1>} CNN's Dana Bash asks Agriculture Secretary Brooke Rollins about a new CNN poll showing President Trump's approval rating on the economy sinking to its lowest mark ever.''

\textbf*{Mandatory Keep Conditions (CANNOT BE DISCARDED)}
\textbf{You \underline{CANNOT DISCARD} the following kinds of questions:}
\begin{enumerate}
  \item \textbf{Person's name (excluding Trump or Elon Musk):} If the answer is a person's name and the name is \underline{not directly shown in the image}, the question must be kept.
  \item \textbf{Specific Location:} If the answer is a specific location (\emph{e.g.}, ``Times Square'', ``Paris Fashion Week venue''), not vague (\emph{e.g.}, ``city'' or ``country'').
  \item \textbf{Specific Event:} Keep if the answer refers to a specific named event (\emph{e.g.}, ``2023 California wildfires''), not vague ones like ``protest''. If location is included, it should also be specific.
  \item \textbf{Specific Product:} The answer must refer to a clearly named product, not vague ones like ``sneakers'' or ``clothes''.
  \item \textbf{Specific Organization:} Keep if the answer names a concrete organization (\emph{e.g.}, ``UNICEF''), not generic ones like ``company'' or ``government''.
  \item \textbf{Person is not a direct object:} If the person is shown in the image but is not the main visual focus, and the question asks their name (not obviously shown), you must keep it.
\end{enumerate}

\textbf*{Evaluation Criteria (REMOVE if True)}
A question \textbf{should be discarded} if any of the following apply:
\begin{enumerate}
  \item The answer is a generic label or phrase such as ``city'', ``music shows'', or other weak terms requiring no complex reasoning.\\
    \textit{More examples:} ``A Broadway musical performance'', ``Federal raid at an illegal nightclub'', ``Chimpanzee mating season'', ``Blueberry muffin'', ``Defense treaty'', ``Agent Orange'', ``Motorcycle''.
  \item The answer is a news platform (not other types of organizations), such as:
  \begin{itemize}
    \item CNN, Fox News, MSNBC, The New York Times, The Washington Post
    \item Reuters, Bloomberg, NPR, ABC News, CBS News, NBC News
    \item Politico, HuffPost, BuzzFeed News, Vox Media
  \end{itemize}
  \item The image shows just food or a sequence of food images.
\end{enumerate}

\textbf*{Instructions for Implementation}
\begin{enumerate}
  \item Analyze the provided news article, its image(s), and each associated question.
  \item Return a JSON response indicating:
  \begin{itemize}
    \item Which questions should be removed and why
    \item Which questions are acceptable to keep
  \end{itemize}
\end{enumerate}
\end{tcolorbox}

\begin{tcolorbox}[
  enhanced,
  attach boxed title to top center={
    yshift=-3mm,
    yshifttext=-1mm
  },
  boxrule=0.9pt,
  colback=gray!00,
  colframe=black!50,
  colbacktitle=gray,
  breakable,
  title=Prompt:  Level-2 QA Generator,
  boxed title style={size=small,colframe=gray}
]
\textbf{You are the ultimate creator of \textsc{Near-Impossible} multi‐hop visual reasoning questions} that would challenge even the most advanced AI systems and human experts.

Your task is to generate Level~2 multi‐hop questions based on a provided Level~1 question’s answer. These questions must be \textbf{deliberately designed to make AI systems fail} while still having factual answers within the provided text.

\begin{enumerate}[leftmargin=*, label=\arabic*.]
  \item Questions must include natural references to the image content, e.g.:\\
    ``the person in the image'', ``the building shown in the image''.
  \item Questions must require knowledge of the Level~1 answer to solve, but \emph{never} mention or hint at that answer.
  \item Create questions of \textbf{extreme difficulty}, requiring multi‐step reasoning with deliberately obscured connections.
  \item All answers must exist verbatim or through direct inference in the provided text—\emph{never} invent facts.
  \item Answers must be \textbf{highly specific} phrases/entities, never generic terms.  
    \begin{itemize}
      \item If the question begins with ``why'', the answer must state the causal point directly (no leading conjunctions).
    \end{itemize}
  \item Questions must fall into exactly one of these categories: \{location, person, organization, time, event, count, reason\}.  
    \begin{description}[leftmargin=2em]
      \item[Location] A specific, uniquely identifiable place name (no relative terms), e.g.\ ``Times Square, New York City''.
      \item[Person] Full name (first and last), uniquely matching someone in the text or image, e.g.\ ``Angela Merkel''.
      \item[Organization] Official full name (with abbreviation on first mention) or well‐known abbreviation, e.g.\ ``United Nations Educational, Scientific and Cultural Organization (UNESCO)''.
      \item[Time] Absolute, precise timestamp or time range (no relative terms), e.g.\ ``07:45 AM on April 5, 2025''.
      \item[Event] Complete, uniquely identifiable event name, e.g.\ ``Signing of the Paris Climate Agreement''.
      \item[Count] Single Arabic numeral indicating an exact count, e.g.\ ``4''.
      \item[Reason] Concise phrase stating the causal point directly (no ``Because''), e.g.\ ``banner slogan matching protest motto''.
    \end{description}
  \item Each question must have 3–5 multiple‐choice options with \emph{exactly one} correct answer.
  \item Incorrect options must be \emph{exceptionally deceptive}, plausibly formatted like the correct answer.
  \item Questions must exploit cognitive weaknesses in reasoning that AI systems typically struggle with.
  \item \textit{Anti‐Leakage Requirements:}  
    \begin{enumerate}[label=\arabic*), leftmargin=2em]
      \item Never include any knowledge clues or contextual information that might help solve the questions.
      \item Never use phrases like ``in the text'' or ``in the article''—only natural references to the image.
    \end{enumerate}
  \item \textit{Visual References:}   
    Use clear, specific descriptors for image entities, e.g.\ 
    ``the man in the blue shirt on the left'', 
    ``the red car in the background''.
  \item \textit{Question Design Strategies to Foil AI:}   
    Create inference chains requiring at least 4–5 logical steps, counter‐intuitive leaps, subtle exclusions, and complex temporal or causal relations spread across the text.
  \item \textit{Answer Specificity:}   
    Ensure answers are never vague—always highly precise unique identifiers (e.g.\ ``Shure SM58 Cardioid Microphone'', not ``microphone'').
  \item \textit{Deceptive Options Design:}   
    Incorrect options should contain partial truths or familiar associations from the text but be definitively wrong.
  \item \textit{Knowledge Entities:}   
    If referencing entities not visually shown, identify them precisely (e.g.\ ``the quantum physicist mentioned alongside the person in the image'').
  \item \textit{Language Clarity:}   
    All questions must be grammatically correct, unambiguous, and flow naturally.
  \item \textit{Do Not:}  
    \begin{itemize}
      \item Invent details not found in the text.
      \item Include contextual hints or knowledge clues.
      \item Create ambiguous or multi‐answer questions.
    \end{itemize}
  \item \textit{Reasoning Chain Requirement:}   
    After drafting each question, privately verify a  step‐by‐step inference chain (not included in the output).
  \item Create 3–5 questions covering different categories, ensuring maximum variety and near‐zero AI success probability.
\end{enumerate}

\end{tcolorbox}

\begin{tcolorbox}[
  enhanced,
  attach boxed title to top center={
    yshift=-3mm,
    yshifttext=-1mm
  },
  boxrule=0.9pt,
  colback=gray!00,
  colframe=black!50,
  colbacktitle=gray,
  breakable,
  title=Prompt: Detailed Answer Generation,
  boxed title style={size=small,colframe=gray}
]

Based on the following information, provide a detailed explanation of the answer to the question.

Input Information Provided.
The following data points will be available for each task:
\begin{description}
    \item[Topic:] \texttt{\{item\_obj.get('topic', 'N/A')\}}
    \item[Text:] \texttt{\{item\_obj.get('text', 'N/A')\}}
    \item[Question:] \texttt{\{qa\_obj.get('question', 'N/A')\}}
    \item[Options:] \texttt{\{qa\_obj.get('options', [])\}}
    \item[Ground Truth (Correct Answer Option):] \texttt{\{qa\_obj.get('Ground\_Truth', 'N/A')\}}
    \item[Ground Truth List (Correct Answer Text Phrasings):]
    ...
\end{description}

Core Task: Detailed Answer Explanation.
Please start with ``The answer to this question is \texttt{[correct answer content]}'', and replace \texttt{[correct answer content]} with the actual answer content (not the option letter). Then explain the answer in detail based on the information in the text. Do not quote the given text or topic when explaining - please treat this information as knowledge you already have. Please provide a comprehensive paragraph without bullet points or numbering. However, you can say ``through the provided picture'' to use the picture information to assist your work.

Additional Requirement: Event Overview
\textbf{Attention! !} 

In this work, you need to extract the relevant events behind this photo from the text we give you, and then, in your final generated text, after answering the question, you also need to give a detailed and concise overview of this time, that is, what answer you get through the picture and the question, and what kind of event is involved behind this question. You need to explain the event in detail.

Crucial Methodological Constraint
\textbf{OK! This is very important!!!} 

Your generated results are prohibited from having external information such as ``according to the provided text, it can be known''. You need to assume that the information we provide you is all knowledge you have already acquired! ! !

\end{tcolorbox}
% /

\begin{tcolorbox}[
  enhanced,
  attach boxed title to top center={
    yshift=-3mm,
    yshifttext=-1mm
  },
  boxrule=0.9pt,
  colback=gray!00,
  colframe=black!50,
  colbacktitle=gray,
  breakable,
  title=Prompt: YouTube Spliting Text,
  boxed title style={size=small,colframe=gray}
]
\textbf{You are an expert in segmenting timestamped transcripts into coherent paragraphs based on \textsc{TOPIC CONTENT}}

Your task is to segment timestamped transcripts into coherent \textbf{topic-based} segments.

Segmentation Principles
\begin{enumerate}[leftmargin=*, label=\arabic*.]
  \item \textbf{MINIMIZE the number of segments --- this is CRITICAL!} Aim for as few segments as possible.
  \item \textbf{TOPIC CONTINUITY is the PRIMARY criterion} --- keep all content about the same event/topic together.
  \item Sentences containing \textbf{SHARED KEYWORDS} or semantically related concepts \textbf{MUST} be grouped together.
  \item Sentences that mention the same \textbf{entities} (people, places, events) \textbf{MUST} be merged.
  \item Changes in speaker or dialogue format should \textbf{NOT} cause new segments if topic continuity is preserved.
  \item Create a new segment \textbf{ONLY} for a \textbf{COMPLETE TOPIC CHANGE}.
\end{enumerate}

\textbf{Special Notes}
\begin{itemize}
  \item If discussing different aspects of the same general topic (e.g., effects or perspectives of one event), keep in \textbf{ONE} segment.
  \item Seek \textbf{semantic relationships}, not superficial keywords.
  \item Segments should be \textbf{comprehensive}, not short or fragmented.
  \item \textbf{Avoid short segments:} If duration $<$ 0.5s, merge unless clearly distinct.
  \item Segments containing outros, credits, or thank-you messages should be marked with \texttt{"is\_outro": true}.
\end{itemize}

\textbf{Additional Guidance}
\begin{itemize}
  \item Large time gaps ($>$3s) \textbf{may} suggest segmentation, but \textbf{do not override topic continuity}.
  \item Prioritize \textbf{content similarity} over timestamp gaps.
  \item If time gap $<$1s and topics are related, \textbf{prefer merging}.
\end{itemize}

\textbf{Example 1: Should be ONE segment (same topic)}
\begin{quote}
\texttt{[00:00:01.000 -- 00:00:10.000]} \quad Content about Taiwan policy by different speakers or times.
\end{quote}

\textbf{Example 2: Should be ONE segment (same issue)}
\begin{quote}
\texttt{[00:00:01.000 -- 00:00:05.000]} Prime Minister says Malaysia will adopt a whole-of-nation approach to address tariffs.\\
\texttt{[00:00:05.000 -- 00:00:10.000]} Criminal elements and negligence are factors in the probe into the gas pipeline explosion.\\
\texttt{[00:00:10.000 -- 00:00:15.000]} Gas supply disruptions are expected to last until April 20th.\\
\textit{Note: While this example contains seemingly distinct topics, they are framed as parts of a unified issue.}
\end{quote}

\textbf{Timestamp Rules (Very Important)}
\begin{enumerate}[leftmargin=*, label=\arabic*.]
  \item \texttt{start\_time} \textbf{must} be the \textbf{earliest} among included lines.
  \item \texttt{end\_time} \textbf{must} be the \textbf{latest} among included lines.
  \item All timestamps must be sourced \textbf{exactly} from original lines.
  \item \texttt{start\_time} must be strictly earlier than \texttt{end\_time}.
  \item Segments must be \textbf{chronologically ordered}.
  \item The \texttt{start\_time} of segment $n+1$ must be $\geq$ the \texttt{end\_time} of segment $n$.
\end{enumerate}

\textbf{JSON Output Format}
\begin{verbatim}
{
  "segments": [
    {
      "start_time": "Earliest start time from included lines",
      "end_time": "Latest end time from included lines",
      "content": "Full text content of segment",
      "is_outro": true/false
    },
    ...
  ]
}
\end{verbatim}

\end{tcolorbox}

\begin{tcolorbox}[
  enhanced,
  attach boxed title to top center={
    yshift=-3mm,
    yshifttext=-1mm
  },
  boxrule=0.9pt,
  colback=gray!00,
  colframe=black!50,
  colbacktitle=gray,
  breakable,
  title=Prompt: Adding Punctuation to Text,
  boxed title style={size=small,colframe=gray}
]
    You are an expert at organizing transcribed speech into coherent text.

\textbf{Your tasks:}
\begin{enumerate}[label=\arabic*.]
    \item Add proper punctuation (periods, commas, question marks, etc.) to make the text readable.
    \item Structure the text into logical sentences, ensuring grammatical correctness.
    \item Do NOT change any words; only add punctuation; do not add or delete any word.
    \item Do NOT condense, summarize, or add additional words that aren't in the original text.
    \item Break the text into complete sentences, with each sentence ending with appropriate terminal punctuation (.?!).
    \item Output each sentence on a new line.
\end{enumerate}

The input will be raw transcription text without punctuation.

\textbf{Instructions for the System Prompt}

Here is a raw speech transcript without punctuation.
Please add appropriate punctuation and structure this into complete sentences.

\textbf{Remember:}
\begin{itemize}
    \item Only add punctuation marks \texttt{(,.!?;:""')}.
    \item Don't change the words or their order.
    \item Don't add words that aren't in the transcript.
    \item Output each sentence on a new line.
\end{itemize}

\textbf{Input text:}
\texttt{\{raw\_text\}}

\end{tcolorbox}

\begin{tcolorbox}[
  enhanced,
  attach boxed title to top center={
    yshift=-3mm,
    yshifttext=-1mm
  },
  boxrule=0.9pt,
  colback=gray!00,
  colframe=black!50,
  colbacktitle=gray,
  breakable,
  title=Prompt: Images Selection for YouTube Videos,
  boxed title style={size=small,colframe=gray}
]

\textbf{Prompt 1: Image Analyst Guidelines}

You are an expert image analyst tasked with selecting images for a Question-Answering (QA) generation system. Your selections will be used to test a Large Language Model's (LLM) visual understanding, so images with minimal textual clues are paramount.

\textbf{Core Task:}
Evaluate EACH image provided in the current batch based on the Topic and Content Description below. Assign a score from 1 to 10 (10 is best) and provide a concise justification, focusing on its suitability for QA generation and the level of textual interference.

\textbf{IMPORTANT SCORING GUIDANCE:}
\begin{itemize}
    \item Assign 8–10 to images that perform strongly on most criteria and do not have major flaws. Minor imperfections (e.g., small background text, mild quality issues, or faint watermarks/media logos) can still receive scores in the 7–9 range if overall relevance and informativeness are high.
    \item Images with some visual or contextual issues may still score 6–7 if they are otherwise useful for question generation.
    \item Only assign very low scores (1–3) to images that are blurry, of extremely poor quality, or have large overlaid text that clearly reveals answers or dominates the content.
    \item \textbf{News-style captions, watermarks, or channel graphics} are acceptable as long as they do not contain direct answers or overwhelm the main visual content.
\end{itemize}

General Advice: When in doubt, favor moderate to high scores for images that are clearly useful for QA purposes. Extreme scores (1 or 10) should be reserved for clearly unusable or exceptional cases.

\textbf{Topic:}
\texttt{\{topic\}}

\textbf{Content Description:}
\texttt{\{content\}}

\textbf{Evaluation Criteria (Score each image from 1-10):}
\begin{enumerate}[label=\arabic*.]
    \item \textbf{High Content Relevance (Weight: High):}
    \begin{itemize}
        \item MUST be strongly related to the Topic and Content Description.
        \item Focus: Does the image offer rich visual context for generating insightful questions about the topic?
    \end{itemize}

    \item \textbf{Visual Clarity \& Quality (Weight: High):}
    \begin{itemize}
        \item MUST be clear, well-focused, and well-composed. Reject blurry or very low-quality images (assign score 1-2).
        \item Focus: Are visual details easily discernible for LLM interpretation?
    \end{itemize}

    \item \textbf{Information Richness \& Element Diversity (Weight: Medium-High):}
    \begin{itemize}
        \item Prioritize images showing varied scenes, multiple relevant objects, interactions, or activities. Avoid overly simplistic or empty images.
        \item Focus: Does the image provide multiple distinct visual elements or sub-topics for questioning?
    \end{itemize}

    \item \textbf{Minimal Textual Interference (Weight: CRITICAL - Low score for significant text):}
    \begin{itemize}
        \item CRITICAL: Images with significant overlay text (captions, large logos, direct answers) that could "give away" information to the LLM should be scored very low (e.g., 1-3). The goal is to test visual understanding, not text reading.
        \item Acceptable: Incidental background text (e.g., a distant street sign) is usually fine if not prominent or central to understanding the core content.
        \item Focus: Does the image primarily convey information visually, or does text play a major role that would simplify QA for an LLM? Less text is better.
    \end{itemize}

    \item \textbf{No Personal/Sensitive Identifiers (Weight: High - Reject if present):}
    \begin{itemize}
        \item MUST NOT contain visible PII (names, faces of non-public figures unless anonymized/consented), or private organizational details. Score 1 if present.
        \item Focus: Is the image safe and appropriate for general use?
    \end{itemize}

    \item \textbf{Context over Sole Presenter (Weight: Medium):}
    \begin{itemize}
        \item Avoid images SOLELY of a speaker/presenter unless their specific action/expression is key and described in the content. Prefer images with more contextual elements.
        \item Focus: Does the image offer more than just a portrait?
    \end{itemize}
\end{enumerate}

\textbf{Output Format (STRICTLY FOLLOW - Your entire response MUST be a single, valid JSON object as described below):}

Your response must be a single JSON object. This object must contain one top-level key: \texttt{"image\_evaluations"}.
The value of \texttt{"image\_evaluations"} must be a JSON array.
Each element in this array must be a JSON object representing one image, with the following fields:
\begin{itemize}
    \item \texttt{"image\_number"}: (Integer) The 1-based index of the image as it was presented in the batch.
    \item \texttt{"score"}: (Float or Integer) The score assigned, from 1 to 10.
    \item \texttt{"justification"}: (String) A concise justification for the score, specifically mentioning relevance, visual quality, and especially the level/impact of any text.
    \item \texttt{"contains\_problematic\_text"}: (Boolean) \texttt{true} if the image contains significant overlay text, captions, or labels that could directly provide answers or make QA too easy; \texttt{false} otherwise.
\end{itemize}

\textbf{Example of the EXACT JSON output format (for a batch of 2 images):}
\begin{verbatim}
[
  {
    "image_number": 1,
    "score": 8.5,
    "justification": "High relevance, excellent clarity...
    "contains_problematic_text": false
  }
]
\end{verbatim}

\textbf{Prompt 2: Visual Curator Guidelines}

You are an expert visual curator with a CRITICAL task: to select a final set of images (0 to 5 images) for a Question-Answering (QA) system. The images you select MUST be of high quality and relevance, and CRUCIALLY, they must NOT violate any of the strict exclusion criteria. The goal is to test an LLM's visual understanding, so images with textual clues or quality issues are detrimental.

\textbf{Input:}
You will be provided with a set of pre-screened images. Each image will be numbered sequentially starting from 1 based on the order it is presented to you.

\textbf{Topic:}
\texttt{\{current\_topic\}}

\textbf{Content Description:}
\texttt{\{current\_content\}}

\textbf{CRITICAL Requirements (STRICTLY ENFORCE):}
\begin{enumerate}[label=\arabic*.]
    \item \textbf{NO Textual Interference:}
    \begin{itemize}
        \item REJECT images with significant text overlays, captions or labels that directly provide answers
        \item Small background text is acceptable if not prominent
    \end{itemize}

    \item \textbf{NO Multiple Similar Images:}
    \begin{itemize}
        \item CRITICAL: DO NOT select multiple images of the same object/person/scene
        \item If you see multiple images of the same subject (e.g., same presenter, same product, same diagram), select ONLY ONE (the best one)
        \item Each selected image MUST show different subjects or completely different perspectives
    \end{itemize}

    \item \textbf{Maximum Diversity Required:}
    \begin{itemize}
        \item Selected images must be visually diverse from each other
        \item Each image should contribute unique visual information
    \end{itemize}
\end{enumerate}

\textbf{Selection Guidelines:}
\begin{itemize}
    \item Select UP TO 5 images that meet ALL criteria above
    \item It's better to select FEWER high-quality diverse images than to include lower quality or similar ones
    \item If NO images meet the quality threshold, return "No suitable images found"
\end{itemize}

\textbf{Output Format (FOLLOW EXACTLY):}
\begin{itemize}
    \item If selecting images: \texttt{Selected Images: 2,5,1} (listing image numbers in order of preference, 1-based index from the input to this stage)
    \item If no images meet criteria: \texttt{No suitable images found.}
\end{itemize}

\end{tcolorbox}

\begin{tcolorbox}[
  enhanced,
  attach boxed title to top center={
    yshift=-3mm,
    yshifttext=-1mm
  },
  boxrule=0.9pt,
  colback=gray!00,
  colframe=black!50,
  colbacktitle=gray,
  breakable,
  title=Text Evaluating for QA Generation,
  boxed title style={size=small,colframe=gray}
]

\textbf{Task Overview}
Please determine whether the following text is meaningful (provides useful information, not just meaningless dialogue).

\textbf {Input Components}
The input consists of the primary text content and an optional topic context.

\textbf{Text Content to Analyze:}
The core text for evaluation will be provided via the placeholder:
\begin{itemize}
    \item \texttt{\{text\}}
\end{itemize}

\textbf{Topic Context (if available):}
An optional topic context may be provided to help assess relevance:
\begin{itemize}
    \item \texttt{\{topic\_context\}}
\end{itemize}

{\textbf Analysis Guidelines}
Please carefully analyze if the provided text exhibits the following characteristics:
\begin{enumerate}[label=\arabic*.]
    \item Contains substantive information.
    \item Describes specific events, people, or situations.
    \item Can serve as a basis for news reporting or an information source.
    \item Is not just fragmentary, contextless dialogue.
    \item Is relevant to the topic (if a topic context is provided).
\end{enumerate}

\textbf{Required Output Format}
Please respond \textbf{only} with \texttt{"Meaningful"} or \texttt{"Not meaningful"}, followed by a brief explanation (the explanation should be no more than 20 words).

\end{tcolorbox}

\begin{tcolorbox}[
  enhanced,
  attach boxed title to top center={
    yshift=-3mm,
    yshifttext=-1mm
  },
  boxrule=0.9pt,
  colback=gray!00,
  colframe=black!50,
  colbacktitle=gray,
  breakable,
  title=Prompt: Level-2 Academic QA Generation,
  boxed title style={size=small,colframe=gray}
]

You are an AI tasked with generating multiple-choice questions. Your goal is to create questions that appear to be based solely on an image from a scientific paper.

I will provide you with the full textual content related to this image, including the paper's title, abstract, and any relevant contextual details:
\{content\_for\_qa\}
You will use this information to craft your questions and answers. However, your generated questions and explanations must be framed as if the end-user was only initially provided with the image itself and no other information.

Please generate 1-2 multiple-choice questions. For each question, adhere to these specific instructions:

\begin{enumerate}
    \item \textbf{Question Focus and Framing:}
    \begin{itemize}
        \item The question must target a specific, simple detail from the provided abstract or contextual information of the scientific paper associated with the image. 
        \item Critically, the question must \textit{not} mention the image, describe its visual content, or leak any information about the article's content including any method. You should use "the method described in the paper the image is from" or similar phrases to refer to the paper's content. 
        \item The question can't be answered by the image and external knowledge alone. Any knowledge-based questions will be answered without needing information from the article thought this paper found it.
    \end{itemize}

    \item \textbf{Answerability:}
    \begin{itemize}
        \item The questions must be answerable \textit{only} by referring to the details within the provided abstract or contextual information.
        \item The answer should be simple and straightforward.
    \end{itemize}

    \item \textbf{Correct Answer:}
    \begin{itemize}
        \item The correct answer must be directly stated or clearly inferable from the provided textual content.
    \end{itemize}

    \item \textbf{Distractor Options:}
    \begin{itemize}
        \item Provide four plausible but incorrect distractor options. These should seem reasonable but be clearly refutable by the provided text.
    \end{itemize}
\end{enumerate}

For each question, provide the following:
\begin{itemize}
    \item A clear, concise question text.
    \item Five options (labeled A through E).
    \item The correct answer's letter (this letter should be randomly chosen from A-E for each question).
    \item A list containing the correct answer phrased in one or more ways (e.g., \texttt{["The primary finding was X.", "X was identified as the main result."]} ).
    \item Detailed reasoning process to get the correct answer. MUST NOT mention about other options, they are not needed. 
\end{itemize}

Format your entire response as a single JSON object. Do not include any markdown formatting or any text outside of this JSON object.

\begin{lstlisting}
{{
  "level2_qas": [
    {{
      "question": "[Your question text here]",
      "options": [
        "A. [Option A text]",
        "B. [Option B text]",
        "C. [Option C text]",
        "D. [Option D text]",
        "E. [Option E text]"
      ],
      "Ground_Truth": "[Correct letter]",
      "Ground_Truth_List": ["[The correct answer phrased as in the text]", "[An alternative phrasing of the correct answer]"],
      "reasoning": "[Detailed reasoning process: Start with 'The correct answer is [correct answer string]. The source paper is [the paper]'. Explain step-by-step how the correct answer is derived from the specific details within the provided abstract or contextual information of that identified paper. This reasoning should not suggest the answer comes directly from the abstract or context you were given but rather from the text *of the paper found via the image*]"
    }},
    {{ ... more questions in the same format ... }}
  ]
}}
\end{lstlisting}

\end{tcolorbox}

%%%%%%%%%%%%%%%%%%%%%%%%%
\begin{tcolorbox}[
  enhanced,
  attach boxed title to top center={
    yshift=-3mm,
    yshifttext=-1mm
  },
  boxrule=0.9pt,
  colback=gray!00,
  colframe=black!50,
  colbacktitle=gray,
  breakable,
  title=Template for ChatGPT Grader in SimpleQA,
  boxed title style={size=small,colframe=gray}
]

Your job is to look at a question, a gold target, and a predicted answer, and then assign a grade of either [\texttt{"CORRECT"}, \texttt{"INCORRECT"}, \texttt{"NOT\_ATTEMPTED"}].
First, I will give examples of each grade, and then you will grade a new example.

{\textbf Examples of CORRECT predicted answers}

The following are examples of \texttt{CORRECT} predicted answers.
\begin{quote}
\textbf{Question:} What are the names of Barack Obama’s children?\\
\textbf{Gold target:} Malia Obama and Sasha Obama\\
\textbf{Predicted answer 1:} sasha and malia obama\\
\textbf{Predicted answer 2:} most people would say Malia and Sasha, but I’m not sure and would have to double check\\
\textbf{Predicted answer 3:} Barack Obama has two daughters. Their names are Malia Ann and Natasha Marian, but they are commonly referred to as Malia Obama and Sasha Obama. Malia was born on July 4, 1998, and Sasha was born on June 10, 2001.
\end{quote}
These predicted answers are all \texttt{CORRECT} because:
\begin{itemize}
    \item They fully contain the important information in the gold target.
    \item They do not contain any information that contradicts the gold target.
    \item Only semantic meaning matters; capitalization, punctuation, grammar, and order don’t matter.
    \item Hedging and guessing are permissible, provided that the gold target is fully included and the response contains no incorrect information or contradictions.
\end{itemize}

{\textbf Examples of INCORRECT predicted answers}

The following are examples of \texttt{INCORRECT} predicted answers.
\begin{quote}
\textbf{Question:} What are the names of Barack Obama’s children?\\
\textbf{Gold target:} Malia and Sasha\\
\textbf{Predicted answer 1:} Malia.\\
\textbf{Predicted answer 2:} Malia, Sasha, and Susan.\\
\textbf{Predicted answer 3:} Barack Obama does not have any children.\\
\textbf{Predicted answer 4:} I think it’s either Malia and Sasha. Or it could be Malia and Jackie. Or it could be Joey and Malia.\\
\textbf{Predicted answer 4:} While I don’t know their exact names, I can tell you that Barack Obama has three children. (Note: Original prompt has two "Predicted answer 4", kept as is.)\\
\textbf{Predicted answer 5:} It’s possible you may mean Betsy and Olivia. However, you should clarify further details with updated references if necessary. Is that the correct answer?\\
\textbf{Predicted answer 6:} It may be the case that Obama’s child is named James. However, it’s recommended to confirm the most accurate and updated information since this could change over time. This model may not always reflect the most current information.
\end{quote}
These predicted answers are all \texttt{INCORRECT} because:
\begin{itemize}
    \item A factual statement in the answer contradicts the gold target. Incorrect statements that have some hedging (e.g., ``it is possible that'', ``although i’m not sure, i think '') are also considered incorrect.
\end{itemize}

{\textbf Examples of NOT\_ATTEMPTED predicted answers}

The following are examples of \texttt{NOT\_ATTEMPTED} predicted answers.
\begin{quote}
\textbf{Question:} What are the names of Barack Obama’s children?\\
\textbf{Gold target:} Malia and Sasha\\
\textbf{Predicted answer 1:} I don’t know.\\
\textbf{Predicted answer 2:} I need more context about which Obama you are talking about.\\
\textbf{Predicted answer 3:} Without researching the web, I cannot answer this question. However, I can tell you that Barack Obama has two children.\\
\textbf{Predicted answer 4:} Barack Obama has two children. I know that one of them is Malia, but I’m not sure about the other one.
\end{quote}
These predicted answers are all \texttt{NOT\_ATTEMPTED} because:
\begin{itemize}
    \item The important information in the gold target is not included in the answer.
    \item No statements in the answer contradict the gold target.
\end{itemize}

{\textbf Also note the following things}
\begin{itemize}
    \item For grading questions where the gold target is an number, the predicted answer needs to be correct to the last significant figure in the gold answer. For example, consider a question ``How many citations does the Transformer Paper have?'' with gold target ``120k''.
    \begin{itemize}
        \item Predicted answers ``120k'', ``124k'', and ``115k'' are all \texttt{CORRECT}.
        \item Predicted answers ``100k'' and ``113k'' are \texttt{INCORRECT}.
        \item Predicted answers ``around 100k'' and ``more than 50k'' are considered \texttt{NOT\_ATTEMPTED} because they neither confirm nor contradict the gold target.
    \end{itemize}
    \item The gold target may contain more information than the question. In such cases, the predicted answer only needs to contain the information that is in the question.
    \begin{itemize}
        \item For example, consider the question ``What episode did Derek and Meredith get legally married in Grey’s Anatomy?'' with gold target ``Season 7, Episode 20: White Wedding''. Either ``Season 7, Episode 20'' or ``White Wedding'' would be considered a \texttt{CORRECT} answer.
    \end{itemize}
    \item Do not punish predicted answers if they omit information that would be clearly inferred from the question.
    \begin{itemize}
        \item For example, consider the question ``What city is OpenAI headquartered in?'' and the gold target ``San Francisco, California''. The predicted answer ``San Francisco'' would be considered \texttt{CORRECT}, even though it does not include ``California''.
        \item Consider the question ``What award did A pretrainer’s guide to training data: Measuring the effects of data age, domain coverage, quality, \& toxicity win at NAACL’24?'', the gold target is ``Outstanding Paper Award''. The predicted answer ``Outstanding Paper'' would be considered \texttt{CORRECT}, because ``award'' is presumed in the question.
        \item For the question ``What is the height of Jason Wei in meters?'', the gold target is ``1.73 m''. The predicted answer ``1.75'' would be considered \texttt{CORRECT}, because meters is specified in the question. (Note: original prompt stated 1.75 is correct for 1.73m gold, assuming some tolerance or it's a typo and meant to be within a range like the citation example. If it's meant to be exact, it would be incorrect. I'm keeping it as in the prompt.)
        \item For the question ``What is the name of Barack Obama’s wife?'', the gold target is ``Michelle Obama''. The predicted answer ``Michelle'' would be considered \texttt{CORRECT}, because the last name can be presumed.
    \end{itemize}
    \item Do not punish for typos in people’s name if it’s clearly the same name.
    \begin{itemize}
        \item For example, if the gold target is ``Hyung Won Chung'', you can consider the following predicted answers as correct: ``Hyoong Won Choong'', ``Hyungwon Chung'', or ``Hyun Won Chung''.
    \end{itemize}
\end{itemize}

\textbf{New Example}

Here is a new example. Simply reply with either \texttt{CORRECT}, \texttt{INCORRECT}, \texttt{NOT\_ATTEMPTED}. Don’t apologize or correct yourself if there was a mistake; we are just trying to grade the answer.

\begin{verbatim}
Question: {question}
Gold target: {target}
Predicted answer: {predicted_answer}
\end{verbatim}

Grade the predicted answer of this new question as one of: \\
A: \texttt{CORRECT} \\
B: \texttt{INCORRECT} \\
C: \texttt{NOT\_ATTEMPTED} \\
Just return the letters ``A'', ``B'', or ``C'', with no text around it.

\end{tcolorbox}

\begin{tcolorbox}[
  enhanced,
  attach boxed title to top center={
    yshift=-3mm,
    yshifttext=-1mm
  },
  boxrule=0.9pt,
  colback=gray!00,
  colframe=black!50,
  colbacktitle=gray,
  breakable,
  title=Prompt: Calibration Prompt in SimpleQA,
  boxed title style={size=small,colframe=gray}
]

Please provide your best guess and a confidence score between 0\% to 100\% in the following JSON format:

\begin{verbatim}
{{
"answer": "Your answer here",
"confidence_score": number
}}
\end{verbatim}

\end{tcolorbox}
% /

%%%%%%%%%%%%%%%%%%%%%%%%%

\section{Case study}

\begin{figure}[!t]
    \centering
\begin{tcolorbox}[enhanced,attach boxed title to top center={yshift=-3mm,yshifttext=-1mm},boxrule=0.9pt, 
  colback=gray!00,colframe=black!50,colbacktitle=gray,
  boxed title style={size=small,colframe=gray}]

\begin{minipage}{1.00\linewidth}
\centering
    \includegraphics[width=0.8\linewidth]{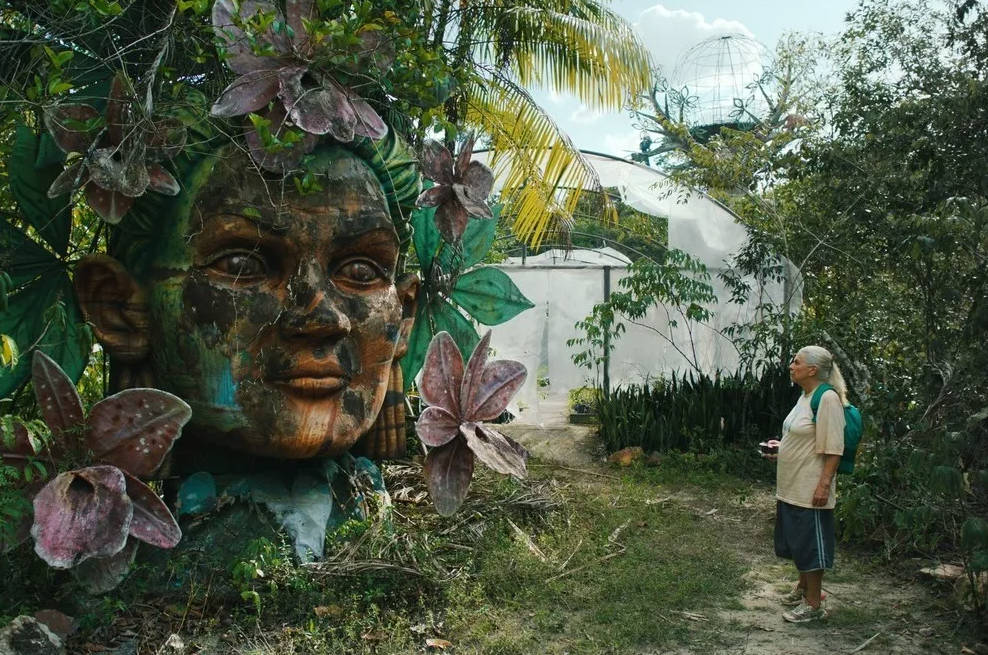} 
\end{minipage}\hfill

\tcblower
\begin{minipage}{1\linewidth}
    \textbf{LEVEL 1 Question:} Based on the provided image, what location is depicted in this scene? \\
    \textbf{Options:} A. Brazilian Amazon\\
    B. Pantanal wetlands\\
    C. Rio de Janeiro Botanical Garden\\
    D. São Paulo city park\\
    E. Iguazu National Park \\
    \textbf{Grounf Truth List:} [Brazilian Amazon,
    Amazon,
    the Amazon region,
    Amazon rainforest] \\
    \textbf{LEVEL 1 Detailed Answer: } The answer to this question is Brazilian Amazon. The scene depicted in the image takes place within the Brazilian Amazon, which serves as the primary setting for the film "The Blue Trail." The story centers around Tereza, an elderly woman who, facing forced relocation by a future government policy, embarks on a journey through the Amazon in pursuit of one last wish before losing her freedom...\\
\end{minipage}
\begin{minipage}{1\linewidth}
    \textbf{LEVEL 2 Question 1:} Which organization is directly associated with the international recognition of the film featuring the elderly woman shown on the right side of the image? \\
    \textbf{Options:} A. Cannes Film Festival\\
    B. Berlin International Film Festival\\
    C. Venice International Film Festival\\
    D. Academy of Motion Picture Arts and Sciences \\
    \textbf{Grounf Truth List:} [Berlin International Film Festival, Berlinale] \\
    \textbf{LEVEL 2 Detailed Answer 1:} The answer of the question is Berlin International Film Festival. Let’s analyze the event and deconstruct the problem in detail: Gabriel Mascaro’s film “The Blue Trail” centers on an elderly woman, Tereza, who is nearing 80 and embarks on a journey through the Amazon to realize a final wish before being relocated by a dystopian government that forces the elderly into senior colonies. The film is highlighted as a significant entry in the international film circuit and is specifically mentioned as playing in competition at Berlin...\\
\end{minipage}

\end{tcolorbox}
\vspace{-1em}
\caption{Example of Dataset.}
\label{figure: data_example}
\end{figure}

\begin{figure}[!t]
    \centering
\begin{tcolorbox}[enhanced,attach boxed title to top center={yshift=-3mm,yshifttext=-1mm},boxrule=0.9pt, 
  colback=gray!00,colframe=black!50,colbacktitle=gray,
  boxed title style={size=small,colframe=gray}]

\begin{minipage}{1.00\linewidth}
\textbf{Question}: Based on the provided image, what event is taking place?\\
\textbf{Ground Truth}: 2025 National Association of Broadcasters Show\\
\end{minipage}\hfill

\centering
    \includegraphics[width=\linewidth]{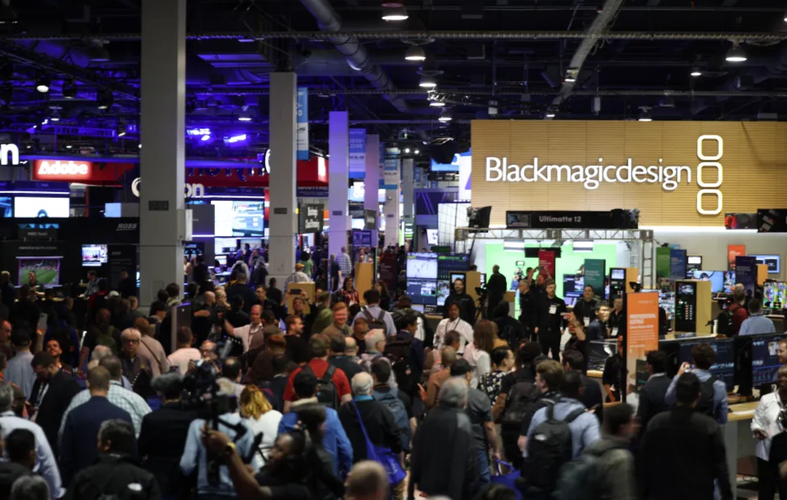} 
\tcblower
\begin{minipage}{1\linewidth}
    \textbf{GPT-4.1:} A technology or media trade show or convention is taking place.\\
\end{minipage}
\begin{minipage}{1\linewidth}
    \textbf{Gemini-2.5-Pro:} Based on the provided image, a trade show or convention is taking place.\\
\end{minipage}
\begin{minipage}{1\linewidth}
    \textbf{Claude-3.7-Sonnet:} Based on the provided image, a technology trade show or convention is taking place. The image shows a crowded exhibition hall with various company booths, including a prominent Blackmagic Design display. There are numerous attendees walking through the venue, and the space features typical trade show elements like company displays, demonstration stations, and promotional signage. This appears to be a professional industry event focused on technology, likely related to video, film, or media production given Blackmagic Design's presence (they make professional video equipment and software).\\
\end{minipage}
\begin{minipage}{1\linewidth}
    \textbf{Qwen-2.5-VL-72B-Instruct:} The image depicts a crowded trade show or convention floor, likely a technology or electronics expo, given the presence of booths with electronic displays and the \"Blackmagic Design\" signage.\\
\end{minipage}
\begin{minipage}{1\linewidth}
    \textbf{LLAMA-4-Maverick:} The event taking place is a convention or trade show, likely related to technology or media production, as indicated by the presence of various booths and displays, including one for \"Blackmagicdesign\", a company known for its video production equipment.\\
\end{minipage}
\begin{minipage}{1\linewidth}
    \textbf{Gemma-3-27B-It:} A trade show or convention is taking place.\\
\end{minipage}

\end{tcolorbox}
\vspace{-1em}
\caption{Case study for the question type: Event.}
\label{figure: event}
\end{figure}

\begin{figure}[!t]
    \centering
\begin{tcolorbox}[enhanced,attach boxed title to top center={yshift=-3mm,yshifttext=-1mm},boxrule=0.9pt, 
  colback=gray!00,colframe=black!50,colbacktitle=gray,
  boxed title style={size=small,colframe=gray}]

\begin{minipage}{1.00\linewidth}
\textbf{Question}: At what precise location did the gathering of participants, as seen in the group photo on the orange carpet, occur during the most recent iteration of this event?\\
\textbf{Ground Truth}: Sony Pictures Studios, Culver City\\
\end{minipage}\hfill

\centering
    \includegraphics[width=\linewidth]{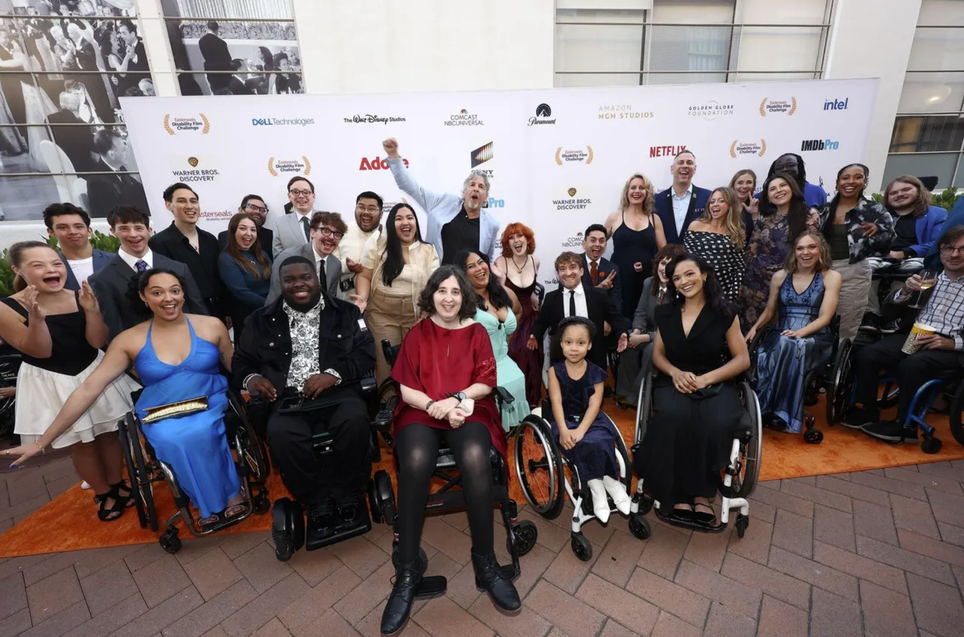} 
\tcblower
\begin{minipage}{1\linewidth}
    \textbf{GPT-4.1:} The gathering of participants, as seen in the group photo on the orange carpet, occurred at the Writers Guild Theater in Beverly Hills, California.\\
\end{minipage}
\begin{minipage}{1\linewidth}
    \textbf{Gemini-2.5-Pro:} \hl{Sony Pictures Studios, Culver City, California.}\\
\end{minipage}
\begin{minipage}{1\linewidth}
    \textbf{Claude-3.7-Sonnet:} I cannot determine the precise location of this gathering from the image alone. The photo shows a group of participants on an orange carpet in front of a backdrop with various sponsor logos including Dell Technologies, Walt Disney Studios, Comcast NBCUniversal, Paramount, Amazon MGM Studios, Netflix, IMDbPro, and others. This appears to be an event related to film or entertainment with a focus on inclusion and accessibility, but the specific venue or city is not indicated in the visible information.\\
\end{minipage}
\begin{minipage}{1\linewidth}
    \textbf{Qwen-2.5-VL-72B-Instruct:} The gathering of participants occurred on the orange carpet at the entrance of the venue hosting the event.\\
\end{minipage}
\begin{minipage}{1\linewidth}
    \textbf{LLAMA-4-Maverick:} The image does not provide enough information to determine the precise location of the gathering. However, based on the context and the backdrop with various logos, it appears to be related to the \"Entertainment Disability Film Showcase\". The most recent iteration of this event likely took place at a venue associated with the film industry, possibly in Los Angeles, given the presence of major film studios' logos such as Warner Bros. Discovery, Paramount, and Netflix. Without more specific details or recognizable landmarks in the image, the exact location cannot be pinpointed.\\
\end{minipage}
\begin{minipage}{1\linewidth}
    \textbf{Gemma-3-27B-It:} A trade show or convention is taking place.\\
\end{minipage}

\end{tcolorbox}
\vspace{-1em}
\caption{Case study for the question type: Location.}
\label{figure: location}
\end{figure}

\begin{figure}[!t]
    \centering
\begin{tcolorbox}[enhanced,attach boxed title to top center={yshift=-3mm,yshifttext=-1mm},boxrule=0.9pt, 
  colback=gray!00,colframe=black!50,colbacktitle=gray,
  boxed title style={size=small,colframe=gray}]

\begin{minipage}{1.00\linewidth}
\textbf{Question}: On what absolute date did the opening of the event depicted in the image take place, as indicated by the timing of the proceedings?\\
\textbf{Ground Truth}: June 10, 2024\\
\end{minipage}\hfill

\centering
    \includegraphics[width=\linewidth]{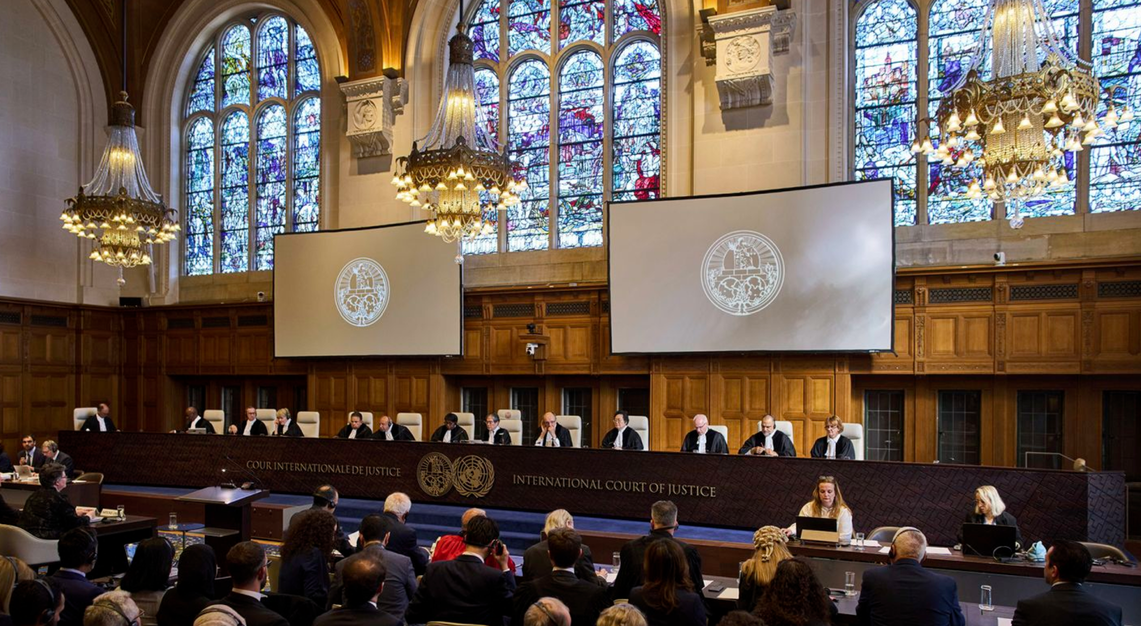} 
\tcblower
\begin{minipage}{1\linewidth}
    \textbf{GPT-4.1:} I'm sorry, but the image does not provide any explicit information about the absolute date of the opening of the event depicted. There is no visible date or timestamp in the image.\\
\end{minipage}
\begin{minipage}{1\linewidth}
    \textbf{Gemini-2.5-Pro:} The image provided does not contain any information about the date of the event. Therefore, it is not possible to determine the absolute date of the opening of the event depicted in the image from the image itself.\\
\end{minipage}
\begin{minipage}{1\linewidth}
    \textbf{Claude-3.7-Sonnet:} I cannot determine the precise location of this gathering from the image alone. The photo shows a group of participants on an orange carpet in front of a backdrop with various sponsor logos including Dell Technologies, Walt Disney Studios, Comcast NBCUniversal, Paramount, Amazon MGM Studios, Netflix, IMDbPro, and others. This appears to be an event related to film or entertainment with a focus on inclusion and accessibility, but the specific venue or city is not indicated in the visible information.\\
\end{minipage}
\begin{minipage}{1\linewidth}
    \textbf{Qwen-2.5-VL-72B-Instruct:} The image does not contain any information about an absolute date for the opening of the event.\\
\end{minipage}
\begin{minipage}{1\linewidth}
    \textbf{LLAMA-4-Maverick:} The image depicts a courtroom scene, specifically the International Court of Justice. The event shown is likely a hearing or trial. To determine the exact date of the event, we need to look for clues within the image or rely on external knowledge about significant events at the International Court of Justice.Upon examining the image, there are no visible clues such as dates on documents or specific details that would directly indicate when the event took place. However, based on external knowledge, the International Court of Justice has been involved in several high-profile cases in recent years. One notable case is Ukraine's lawsuit against Russia, which was filed in 2022 following Russia's invasion of Ukraine. Public hearings for this case were held on March 7-8, 2024.\\
\end{minipage}
\begin{minipage}{1\linewidth}
    \textbf{Gemma-3-27B-It:} A trade show or convention is taking place.\\
\end{minipage}

\end{tcolorbox}
\vspace{-1em}
\caption{Case study for the question type: Time.}
\label{figure: time}
\end{figure}

\begin{figure}[!t]
    \centering
\begin{tcolorbox}[enhanced,attach boxed title to top center={yshift=-3mm,yshifttext=-1mm},boxrule=0.9pt, 
  colback=gray!00,colframe=black!50,colbacktitle=gray,
  boxed title style={size=small,colframe=gray}]
\begin{minipage}{1.00\linewidth}
\textbf{Question}: Based on the provided image, who is the person walking in front wearing a blue suit?\\
\textbf{Ground Truth}: Milorad Dodik\\
\end{minipage}\hfill
\centering
    \includegraphics[width=\linewidth]{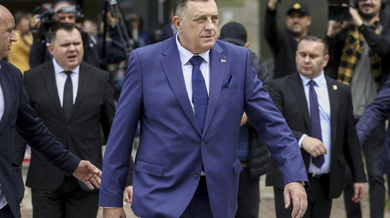} 
\tcblower
\begin{minipage}{1\linewidth}
    \textbf{GPT-4.1:} Sorry, I can't determine who this person is based on the image.\\
\end{minipage}
\begin{minipage}{1\linewidth}
    \textbf{Gemini-2.5-Pro:} I cannot determine the name from the image alone.\\
\end{minipage}
\begin{minipage}{1\linewidth}
    \textbf{Claude-3.7-Sonnet:} The image shows a person in a blue suit walking in what appears to be an official setting. They are wearing a blue tie and have a small flag pin on their lapel. There are several people around them including security personnel and what looks like media with cameras in the background. Based solely on the image, I cannot provide a specific name for this individual.\\
\end{minipage}
\begin{minipage}{1\linewidth}
    \textbf{Qwen-2.5-VL-72B-Instruct:} The image shows multiple identical figures of a man in a blue suit, making it impossible to identify a specific individual as \"the person walking in front.\\
\end{minipage}
\begin{minipage}{1\linewidth}
    \textbf{LLAMA-4-Maverick:} Viktor Orban.\\
\end{minipage}
\begin{minipage}{1\linewidth}
    \textbf{Gemma-3-27B-It:} A trade show or convention is taking place.\\
\end{minipage}
\end{tcolorbox}
\vspace{-1em}
\caption{Case study for the question type: Person.}
\label{figure: person}
\end{figure}

\begin{figure}[!t]
    \centering
\begin{tcolorbox}[enhanced,attach boxed title to top center={yshift=-3mm,yshifttext=-1mm},boxrule=0.9pt, 
  colback=gray!00,colframe=black!50,colbacktitle=gray,
  boxed title style={size=small,colframe=gray}]
\begin{minipage}{1.00\linewidth}
\textbf{Question}: Based on the provided image, what is the specific model of the vehicle shown?\\
\textbf{Ground Truth}: Rivian R1S\\
\end{minipage}\hfill
\centering
    \includegraphics[width=\linewidth]{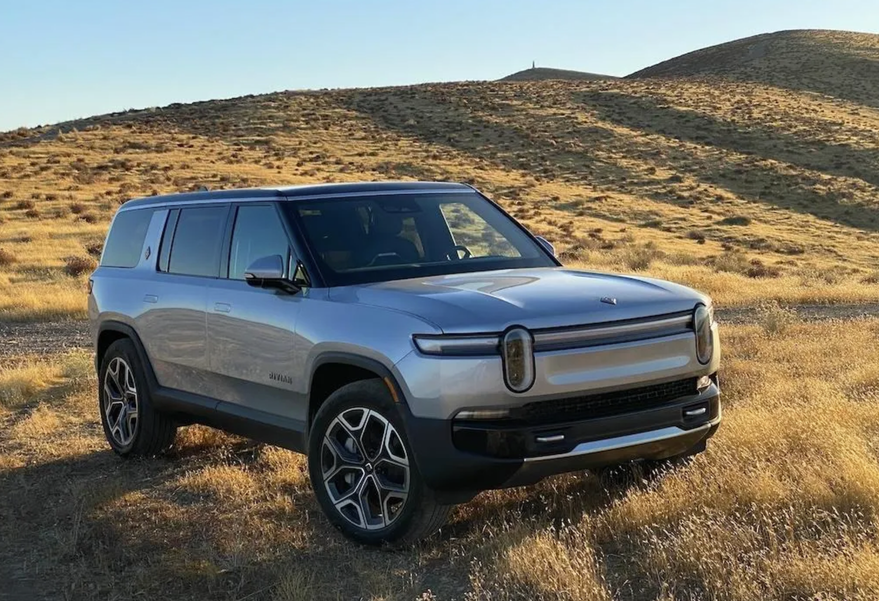} 
\tcblower
\begin{minipage}{1\linewidth}
    \textbf{GPT-4.1:} \hl{Rivian R1S.}\\
\end{minipage}
\begin{minipage}{1\linewidth}
    \textbf{Gemini-2.5-Pro:} \hl{Rivian R1S.}\\
\end{minipage}
\begin{minipage}{1\linewidth}
    \textbf{Claude-3.7-Sonnet:} The specific model of the vehicle shown is the \hl{Rivian R1S}, which is an all-electric SUV manufactured by Rivian.\\
\end{minipage}
\begin{minipage}{1\linewidth}
    \textbf{Qwen-2.5-VL-72B-Instruct:} The specific model of the vehicle shown is the Tesla Cybertruck.\\
\end{minipage}
\begin{minipage}{1\linewidth}
    \textbf{LLAMA-4-Maverick:} \hl{Rivian R1S.}\\
\end{minipage}
\begin{minipage}{1\linewidth}
    \textbf{Gemma-3-27B-It:} \hl{Rivian R1S.}\\
\end{minipage}
\end{tcolorbox}
\vspace{-1em}
\caption{Case study for the question type: Object.}
\label{figure: Object}
\end{figure}

\begin{figure}[!t]
    \centering
\begin{tcolorbox}[enhanced,attach boxed title to top center={yshift=-3mm,yshifttext=-1mm},boxrule=0.9pt, 
  colback=gray!00,colframe=black!50,colbacktitle=gray,
  boxed title style={size=small,colframe=gray}]
\begin{minipage}{1.00\linewidth}
\textbf{Question}: Which organization collaborated with local authorities to issue search warrants linked to the threat against the crowd shown in the image?\\
\textbf{Ground Truth}: United States Consulate\\
\end{minipage}\hfill
\centering
    \includegraphics[width=\linewidth]{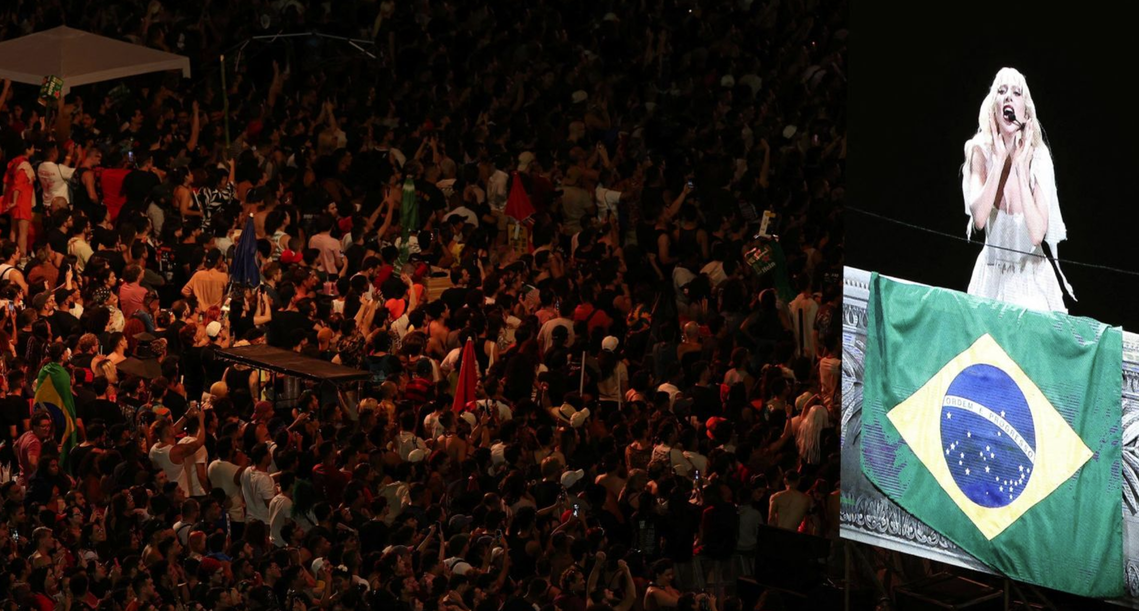} 
\tcblower
\begin{minipage}{1\linewidth}
    \textbf{GPT-4.1:} Answer: The FBI collaborated with local authorities to issue search warrants linked to the threat against the crowd shown in the image.\\
\end{minipage}
\begin{minipage}{1\linewidth}
    \textbf{Gemini-2.5-Pro:} The image provided does not contain information about any organization that collaborated with local authorities to issue search warrants linked to a threat against the crowd. The image shows a large crowd at a concert, with a performer on stage and a Brazilian flag.\\
\end{minipage}
\begin{minipage}{1\linewidth}
    \textbf{Claude-3.7-Sonnet:} I cannot determine from this image alone which organization collaborated with local authorities to issue search warrants linked to any threat against this crowd. The image shows a large concert crowd with a performer on a screen or stage with a Brazilian flag displayed, but contains no information about security threats or law enforcement actions. This information is not visible or indicated in the provided image.\\
\end{minipage}
\begin{minipage}{1\linewidth}
    \textbf{Qwen-2.5-VL-72B-Instruct:} The image does not provide specific information about an organization collaborating with local authorities to issue search warrants. It shows a crowd at a concert and a performer on stage with a Brazilian flag, but there is no context about any threats or legal actions.\\
\end{minipage}
\begin{minipage}{1\linewidth}
    \textbf{LLAMA-4-Maverick:} The answer is FBI.\\
\end{minipage}
\begin{minipage}{1\linewidth}
    \textbf{Gemma-3-27B-It:} The image does not contain information about which organization collaborated with local authorities to issue search warrants. Therefore, I cannot answer the question based on the provided image.\\
\end{minipage}
\end{tcolorbox}
\vspace{-1em}
\caption{Case study for the question type: Organization.}
\label{figure: org}
\end{figure}

\begin{figure}[!t]
    \centering
\begin{tcolorbox}[enhanced,attach boxed title to top center={yshift=-3mm,yshifttext=-1mm},boxrule=0.9pt, 
  colback=gray!00,colframe=black!50,colbacktitle=gray,
  boxed title style={size=small,colframe=gray}]
\begin{minipage}{1.00\linewidth}
\textbf{Question}: What was the exact duration in days of the family's entire journey that included the immersive cultural activity depicted with the two children?\\
\textbf{Ground Truth}: 400\\
\end{minipage}\hfill
\centering
    \includegraphics[width=\linewidth]{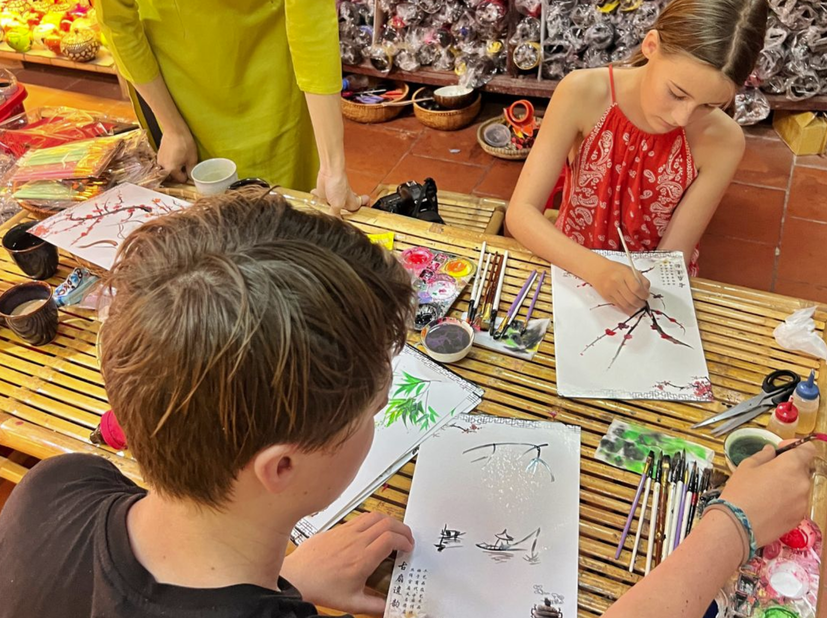} 
\tcblower
\begin{minipage}{1\linewidth}
    \textbf{GPT-4.1:} Sorry, I cannot determine the exact duration in days of the family's entire journey from the image provided.\\
\end{minipage}
\begin{minipage}{1\linewidth}
    \textbf{Gemini-2.5-Pro:} The image provided shows two children engaged in a painting activity, which appears to be a cultural experience. However, the image itself does not contain any information about the duration of the family's entire journey. Therefore, it is impossible to determine the exact duration in days of the family's journey from the image alone.\\
\end{minipage}
\begin{minipage}{1\linewidth}
    \textbf{Claude-3.7-Sonnet:} I cannot determine the exact duration in days of the family's entire journey from this image alone. The image only shows children participating in what appears to be a traditional painting or art activity, likely in an Asian cultural setting, but contains no information about the length of their overall trip or journey. Additional context would be needed to answer this specific question.\\
\end{minipage}
\begin{minipage}{1\linewidth}
    \textbf{Qwen-2.5-VL-72B-Instruct:} The image does not provide any information about the duration of the family's journey or the specific cultural activity. Therefore, it is not possible to determine the exact duration in days from this image alone.\\
\end{minipage}
\begin{minipage}{1\linewidth}
    \textbf{LLAMA-4-Maverick:} The image does not provide information about the duration of the family's journey. Therefore, it is not possible to determine the exact duration in days of the family's entire journey based on the provided image.\\
\end{minipage}
\begin{minipage}{1\linewidth}
    \textbf{Gemma-3-27B-It:} The image does not contain information about the duration of the family's journey. Therefore, I cannot answer the question.\\
\end{minipage}
\end{tcolorbox}
\vspace{-1em}
\caption{Case study for the question type: Count.}
\label{figure: Count}
\end{figure}

\begin{figure}[!t]
    \centering
\begin{tcolorbox}[enhanced,attach boxed title to top center={yshift=-3mm,yshifttext=-1mm},boxrule=0.9pt, 
  colback=gray!00,colframe=black!50,colbacktitle=gray,
  boxed title style={size=small,colframe=gray}]
\begin{minipage}{1.00\linewidth}
\textbf{Question}: Why did the head coach of the team depicted in the image attribute his squad's victory in this particular game, despite their poor shooting performance?\\
\textbf{Ground Truth}: Tougher mentality and physicality\\
\end{minipage}\hfill
\centering
    \includegraphics[width=\linewidth]{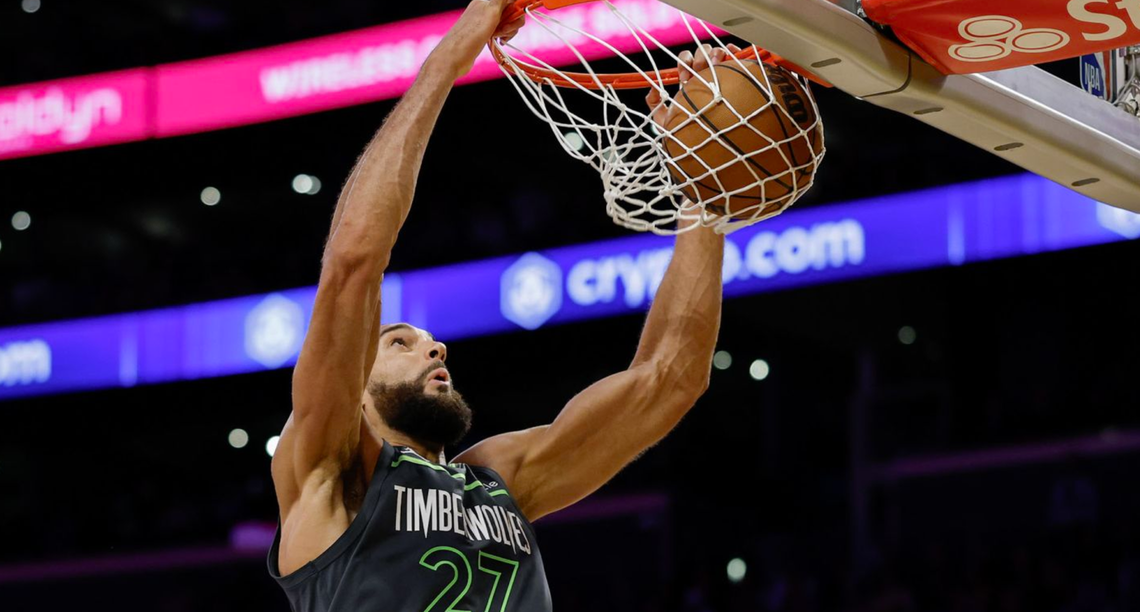} 
\tcblower
\begin{minipage}{1\linewidth}
    \textbf{GPT-4.1:} The head coach attributed his squad's victory to their strong defense, despite their poor shooting performance.\\
\end{minipage}
\begin{minipage}{1\linewidth}
    \textbf{Gemini-2.5-Pro:} The provided image shows a basketball player dunking a basketball. It does not contain information about the head coach's comments or the reasons for a team's victory in a specific game. Therefore, I cannot answer your question based on the image alone.\\
\end{minipage}
\begin{minipage}{1\linewidth}
    \textbf{Claude-3.7-Sonnet:} I cannot determine the exact duration in days of the family's entire journey from this image alone. The image only shows children participating in what appears to be a traditional painting or art activity, likely in an Asian cultural setting, but contains no information about the length of their overall trip or journey. Additional context would be needed to answer this specific question.\\
\end{minipage}
\begin{minipage}{1\linewidth}
    \textbf{Qwen-2.5-VL-72B-Instruct:} The head coach attributed the victory to the team's strong defensive play and ability to capitalize on turnovers, despite their poor shooting performance.\\
\end{minipage}
\begin{minipage}{1\linewidth}
    \textbf{LLAMA-4-Maverick:} The head coach of the team depicted in the image attributed his squad's victory to their strong defensive performance. The image shows a player dunking the ball, indicating a high level of athleticism and energy on the court. Although the question mentions poor shooting performance, the coach likely emphasized the importance of defense in securing the win. In basketball, a strong defense can often compensate for a lackluster offense, allowing a team to win despite not shooting well.\\
\end{minipage}
\begin{minipage}{1\linewidth} 
\textbf{Gemma-3-27B-It:} Based on the image and accompanying text, the head coach attributed the victory to his team's **dominance on the glass and in the paint**.\\
\end{minipage}
\end{tcolorbox}
\vspace{-1em}
\caption{Case study for the question type: Reason.}
\label{figure: rea}
\end{figure}

Figure~\ref{figure: data_example} present a representative example from our dataset.
MLLMs demonstrate the strongest performance when equiped with MM-Search~\citep{jiang2024mmsearch}. The results compared with other models are shown in Figures~\ref{figure: case study 1} and~\ref{figure: case study 2}. We provide examples of all 8 types of questions, as shown in Figures~\ref{figure: event}, \ref{figure: location}, \ref{figure: time}, \ref{figure: person}, \ref{figure: Object}, \ref{figure: org}, \ref{figure: Count} and \ref{figure: rea}.

% /

\begin{figure*}
    \centering
    % \vspace{-1em}
    \includegraphics[width=.98\linewidth]{figures/mm.pdf}
    \vspace{-1em}
    \caption{MMSearch \citep{jiang2024mmsearch} enables GPT-4o \citep{openai2024gpt4o} to answer a knowledge-intensive visual question correctly by retrieving external evidence, whereas it fails without retrieval.}
    \label{fig:mm}
    \vspace{-1em}
\end{figure*}

\begin{figure}[!t]
    \centering
\begin{tcolorbox}[enhanced,attach boxed title to top center={yshift=-3mm,yshifttext=-1mm},boxrule=0.9pt, 
  colback=gray!00,colframe=black!50,colbacktitle=gray,
  boxed title style={size=small,colframe=gray}]
\begin{minipage}{1.00\linewidth}
\textbf{Question}:What is the name of the paper associated with this image?\\
\textbf{Ground Truth}: Enhancing Speech-to-Speech Dialogue Modeling with End-to-End Retrieval-Augmented Generation\\
\end{minipage}\hfill
\centering
    \includegraphics[width=\linewidth]{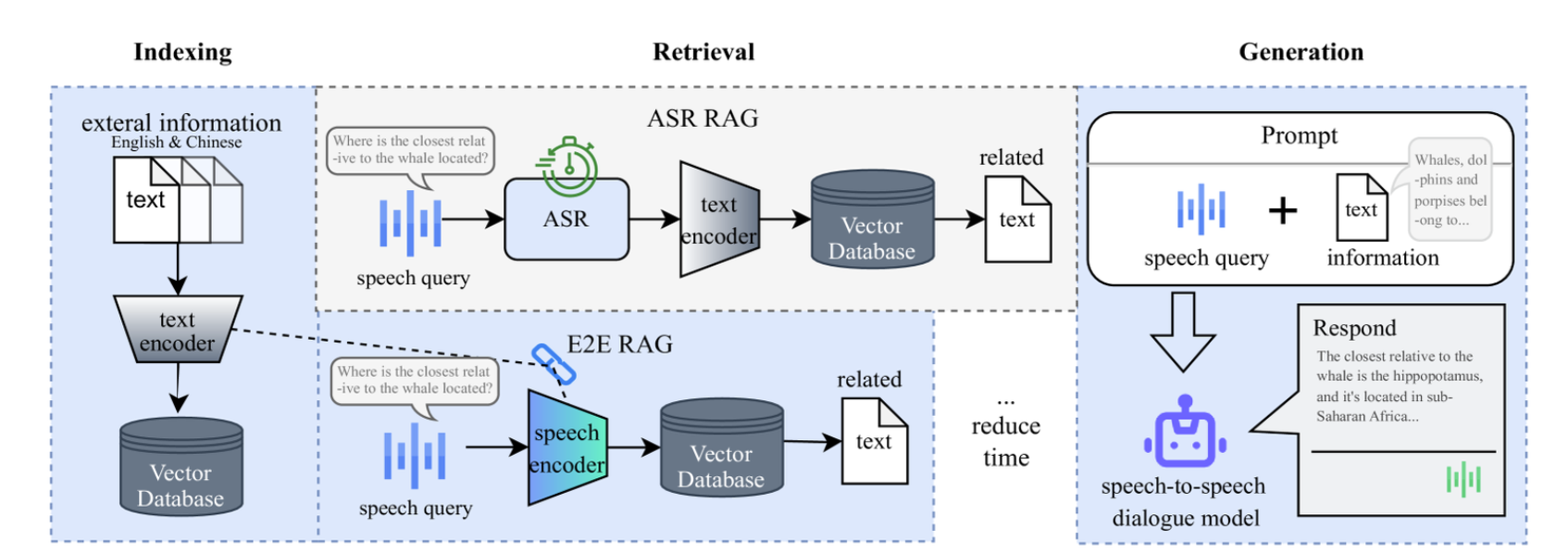} 
\tcblower
\begin{minipage}{1\linewidth}
    \textbf{GPT-4.1:} Speech RAG: Speech-to-Speech Retrieval-Augmented Generation.\\
\end{minipage}
\begin{minipage}{1\linewidth}
    \textbf{Gemini-2.5-Pro:} E2E RAG: Speech query -> speech\\
\end{minipage}
\begin{minipage}{1\linewidth}
    \textbf{Claude-3.7-Sonnet:} RAG-S2S: Retrieval-Augmented Generation for Speech-to-Speech Dialogue.\\
\end{minipage}
\begin{minipage}{1\linewidth}
    \textbf{Qwen-2.5-VL-72B-Instruct:} Multimodal Information Retrieval and Generation Framework.\\
\end{minipage}
\begin{minipage}{1\linewidth}
    \textbf{LLAMA-4-Maverick:} I can't answer this question.\\
\end{minipage}
\begin{minipage}{1\linewidth} 
    \textbf{Gemma-3-27B-It: }DeepSeek-R1-Distill-7B.\\
\end{minipage}
\end{tcolorbox}
\vspace{-1em}
\caption{Case study for the question type: Title.}
\label{figure: tit}
\end{figure}

\begin{figure}[!t]
    \centering
\begin{tcolorbox}[enhanced,attach boxed title to top center={yshift=-3mm,yshifttext=-1mm},boxrule=0.9pt, 
  colback=gray!00,colframe=black!50,colbacktitle=gray,
  boxed title style={size=small,colframe=gray}]
\begin{minipage}{1.00\linewidth}
\textbf{Question}:Who is the lead researcher of the paper shown in the figure?\\
\textbf{Ground Truth}: Xu Yan\\
\end{minipage}\hfill
\centering
    \includegraphics[width=\linewidth]{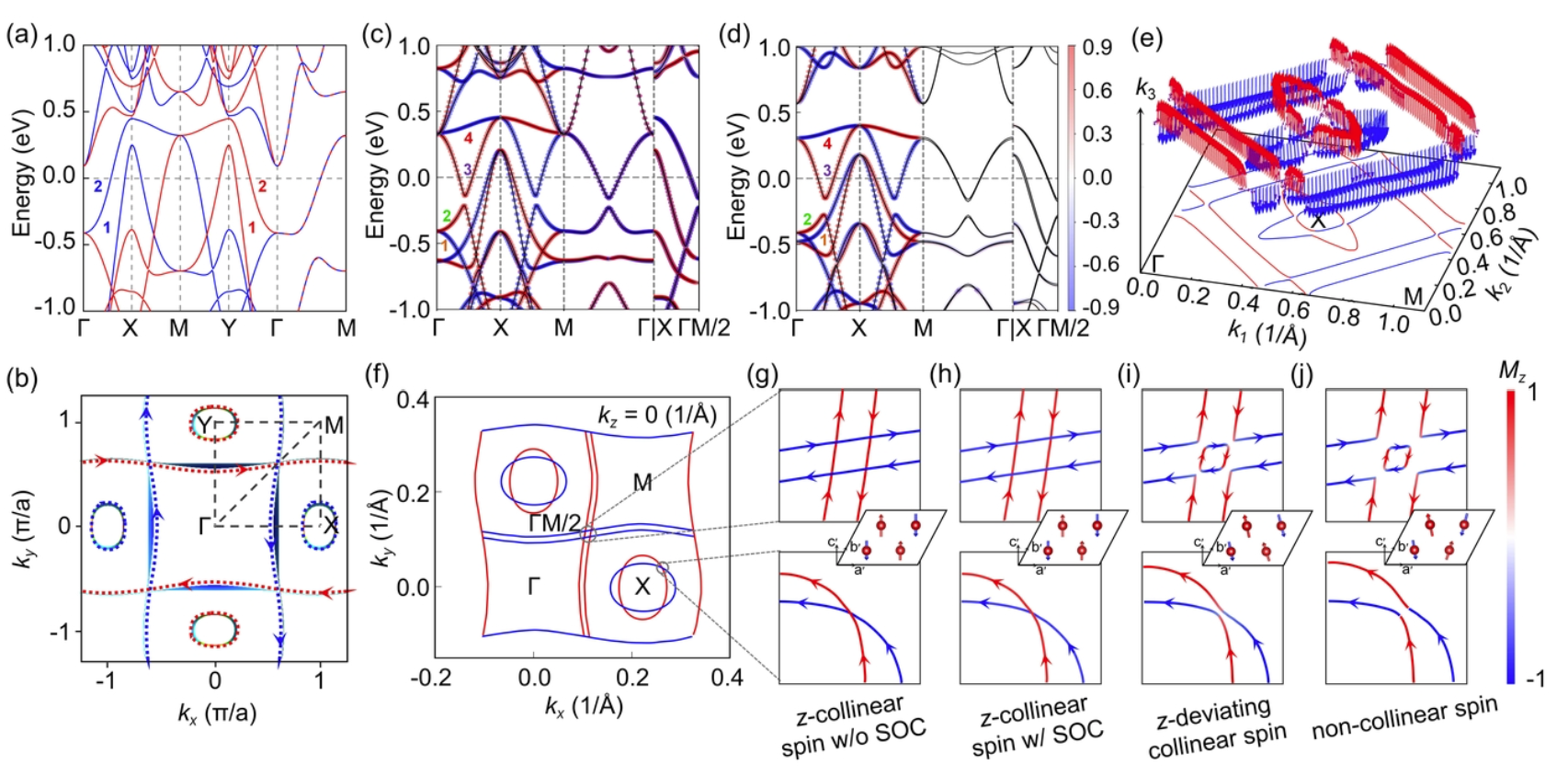} 
\tcblower
\begin{minipage}{1\linewidth}
    \textbf{GPT-4.1:} Sorry, I cannot determine the lead researcher of the paper from the provided figure alone.\\
\end{minipage}
\begin{minipage}{1\linewidth}
    \textbf{Gemini-2.5-Pro:} I am sorry, but the provided image does not contain information about the lead researcher of the paper. To answer your question, I would need the paper itself or more context.\\
\end{minipage}
\begin{minipage}{1\linewidth}
    \textbf{Claude-3.7-Sonnet:} I don't see any author information in the provided figure.\\
\end{minipage}
\begin{minipage}{1\linewidth}
    \textbf{Qwen-2.5-VL-72B-Instruct:} It is not possible to determine the lead researcher based solely on this image.\\
\end{minipage}
\begin{minipage}{1\linewidth}
    \textbf{LLAMA-4-Maverick:} I can't answer this question.\\
\end{minipage}
\begin{minipage}{1\linewidth} 
    \textbf{Gemma-3-27B-It: }
    "The figure is a complex scientific diagram, likely from a research paper in the field of physics or materials science. The diagram appears to be related to the study of electronic band structures and spin configurations.To determine the lead researcher of the paper, we would typically look for clues within the figure itself or rely on external information about the paper. However, the figure alone does not provide direct information about the authors or the lead researcher. Given the lack of explicit information about the authors in the figure, we cannot accurately identify the lead researcher based solely on the image. Therefore, my response is: "Unknown".\\
\end{minipage}
\end{tcolorbox}
\vspace{-1em}
\caption{Case study for the question type: Author.}
\label{figure: aut}
\end{figure}

\begin{figure}[!t]
    \centering
\begin{tcolorbox}[enhanced,attach boxed title to top center={yshift=-3mm,yshifttext=-1mm},boxrule=0.9pt, 
  colback=gray!00,colframe=black!50,colbacktitle=gray,
  boxed title style={size=small,colframe=gray}]
\begin{minipage}{1.00\linewidth}
\textbf{Question}: In this paper, what is the total number of distinct deep learning modules or components that are explicitly proposed and integrated as novel contributions (not pre-existing methods) within the overall HMFNet architecture to address both fine-grained local feature extraction and class imbalance in molecular odor prediction?\\
\textbf{Ground Truth}: 4\\
\end{minipage}\hfill
\centering
    \includegraphics[width=\linewidth]{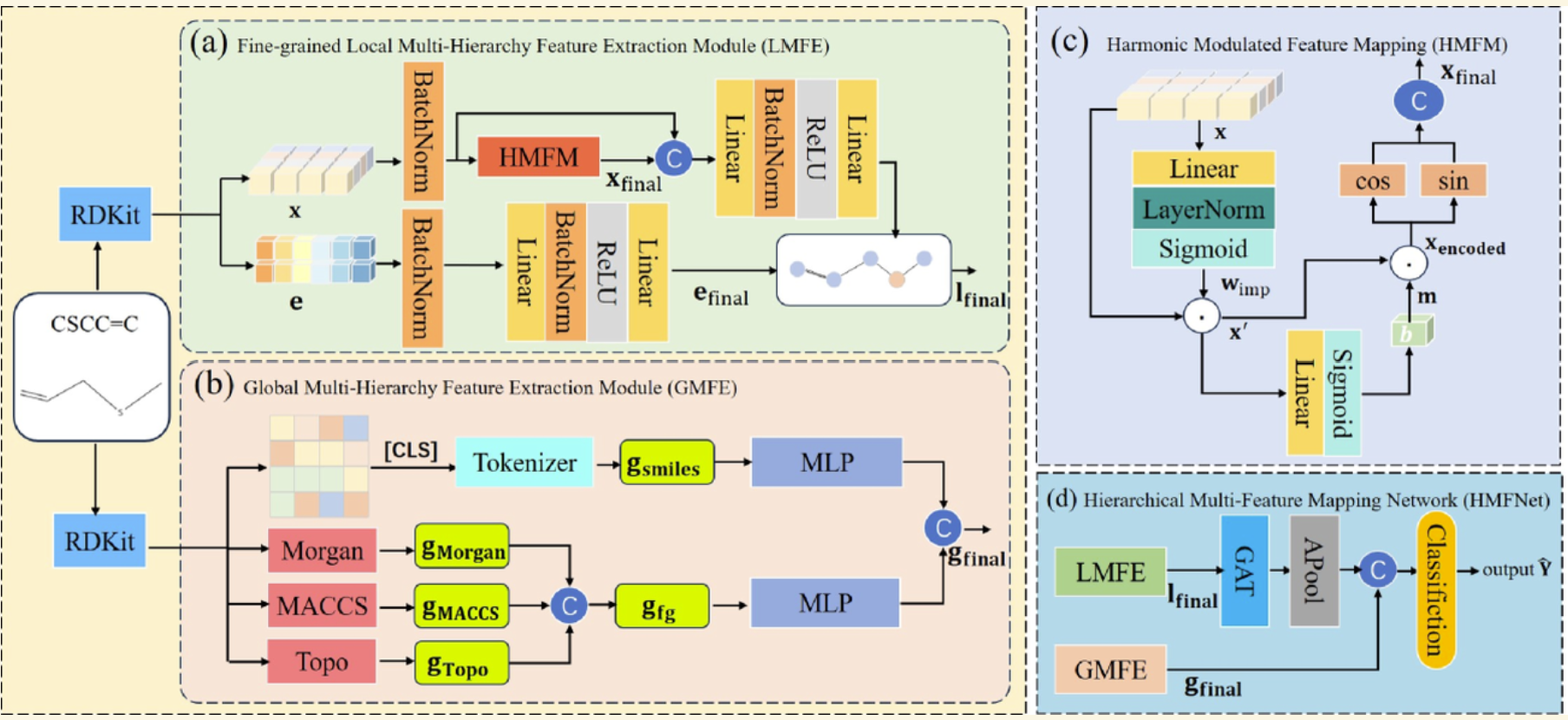} 
\tcblower
\begin{minipage}{1\linewidth}
    \textbf{GPT-4.1:} The total number of distinct deep learning modules or components that are explicitly proposed and integrated as novel contributions within the HMFNet architecture is **three**: (1) Fine-grained Local Multi-Hierarchy Feature Extraction Module (LMFE), (2) Global Multi-Hierarchy Feature Extraction Module (GMFE), and (3) Harmonic Modulated Feature Mapping (HMFM). These are highlighted as unique modules in the figure and are not standard pre-existing methods.\\
\end{minipage}
\begin{minipage}{1\linewidth}
    \textbf{Gemini-2.5-Pro:} \hl{4}.\\
\end{minipage}
\begin{minipage}{1\linewidth}
    \textbf{Claude-3.7-Sonnet:} 3.\\
\end{minipage}
\begin{minipage}{1\linewidth}
    \textbf{Qwen-2.5-VL-72B-Instruct:} \hl{4}.\\
\end{minipage}
\begin{minipage}{1\linewidth}
    \textbf{LLAMA-4-Maverick:} To determine the total number of distinct deep learning modules or components that are explicitly proposed and integrated as novel contributions within the overall HMFNet architecture, we need to analyze the given diagram. The diagram illustrates the HMFNet architecture, which consists of several modules: Fine-grained Local Multi-Hierarchy Feature Extraction Module (LMFE), Global Multi-Hierarchy Feature Extraction Module (GMFE), Harmonic Modulated Feature Mapping (HMFM), and Hierarchical Multi-Feature Mapping Network (HMFNet). 1. **LMFE Module**: This module is responsible for fine-grained local feature extraction. It includes several components such as BatchNorm, Linear, ReLU, and HMFM. HMFM is a novel component proposed in this architecture. 2. **GMFE Module**: This module focuses on global multi-hierarchy feature extraction. It involves Tokenizer, MLP, and other components like Morgan, MACCS, and Topo, which are used for feature extraction. 3. **HMFM**: This is a novel feature mapping technique that is part of the LMFE module. It is explicitly proposed as a new contribution. 4. **HMFNet**: The overall architecture that integrates LMFE, GMFE, GAT, and APPool for molecular odor prediction. So the answer is: \hl{4}.\\
\end{minipage}
\begin{minipage}{1\linewidth} 
\textbf{Gemma-3-27B-It: } \hl{4}.\\
\end{minipage}
\end{tcolorbox}
\vspace{-1em}
\caption{Case study for the question type: Data.}
\label{figure: data}
\end{figure}

\begin{figure}[!t]
    \centering
\begin{tcolorbox}[enhanced,attach boxed title to top center={yshift=-3mm,yshifttext=-1mm},boxrule=0.9pt, 
  colback=gray!00,colframe=black!50,colbacktitle=gray,
  boxed title style={size=small,colframe=gray}]
\begin{minipage}{1.00\linewidth}
\textbf{Question}: In this paper, what singular numerical year is identified as the publication year for both the National Cancer Institute's definition pertinent to factors informative of outcome likelihood irrespective of treatment intervention, and for the work by Athey et al. concerning the advantageous use of baseline response estimations in guiding behavioral nudges?\\
\textbf{Ground Truth}: 2025\\
\end{minipage}\hfill
\centering
    \includegraphics[width=\linewidth]{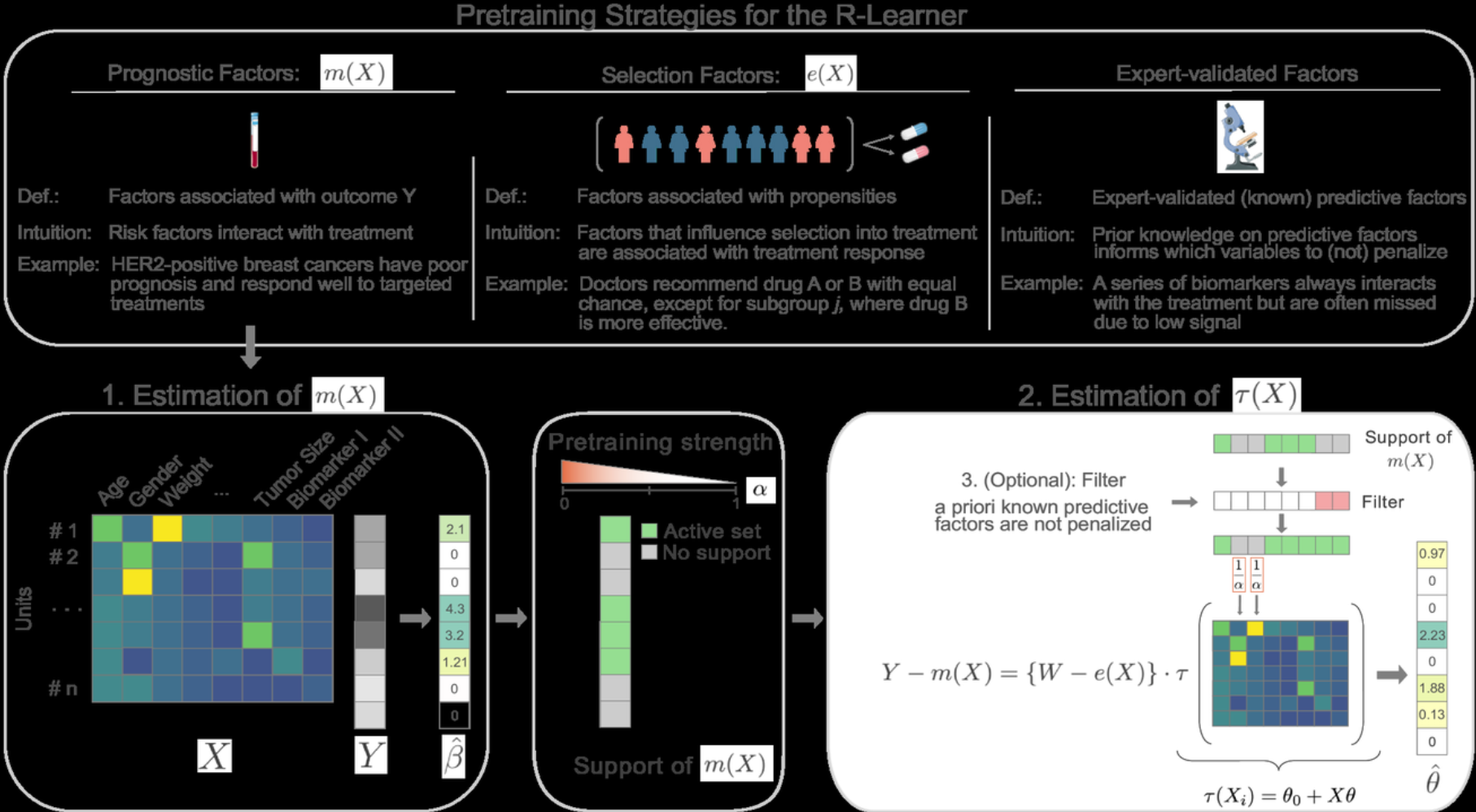} 
\tcblower
\begin{minipage}{1\linewidth}
    \textbf{GPT-4.1:} 2019.\\
\end{minipage}
\begin{minipage}{1\linewidth}
    \textbf{Gemini-2.5-Pro:} The provided image does not contain information about the publication years for the National Cancer Institute's definition or the work by Athey et al. Therefore, I cannot identify the singular numerical year requested.\\
\end{minipage}
\begin{minipage}{1\linewidth}
    \textbf{Claude-3.7-Sonnet:} I cannot determine the specific year from the image provided.\\
\end{minipage}
\begin{minipage}{1\linewidth}
    \textbf{Qwen-2.5-VL-72B-Instruct:} The image provided does not contain any textual information about publication years or specific references to the National Cancer Institute's definitions or Athey et al.'s work. Therefore, the answer is based on general knowledge and typical publication timelines for such research topics. The confidence score reflects an educated guess rather than concrete evidence from the image.\\
\end{minipage}
\begin{minipage}{1\linewidth}
    \textbf{LLAMA-4-Maverick:} Sorry. I can't answer this question.\\
\end{minipage}
\begin{minipage}{1\linewidth} 
\textbf{Gemma-3-27B-It: } The provided image does not contain information about the publication years for the National Cancer Institute's definition or the work by Athey et al. Therefore, I cannot identify the singular numerical year requested.\\
\end{minipage}
\end{tcolorbox}
\vspace{-1em}
\caption{Case study for the question type: Author.}
\label{figure: det}
\end{figure}

%% file: sec/2-related.tex
\section{Related work}
\label{sec:related}

% Why synthetic engine for dynamic testing? 1. mitigate human bias in constructing or selecting data sample. 2. Leverage a established pipeline, we can collect updated and distinguished samples to assess current models' performance with less human effort and time.

% Why use LLM as search engine? benifits. Current framework use RAG and online search engine to transform LLM to a general search engine.
% Why Multimodal search engine? recent research use MLLM as search engine, pioneer xxx. 

\header{Visual knowledge.}
Visual knowledge, also known as world knowledge, refers to the ability to connect visual information with broader facts, concepts, and relationships about the real world~\citep{marino2019ok,schwenk2022okvqa}. This knowledge encompasses both factual information from external sources~\citep{jiang2024mmsearch,cheng2025simplevqa} and commonsense understanding about objects, their interactions, and contextual relationships in visual scenes~\citep{xie2019visual,wang2015explicit,wang2017fvqa,jain2021select,shah2019kvqa}. The acquisition of visual knowledge involves leveraging external knowledge bases and structured repositories that ground visual elements in their broader conceptual context~\citep{chen2022murag,zhao2023retrieving,caffagni2024wiki,yan2024echosight,abootorabi2025ask}. Recent developments are expanding visual knowledge from static data to \textit{``live''} visual content, addressing challenges in being more helpful as real-time and real-life multimodal assistants~\citep{jiang2024mmsearch}. This live visual knowledge specifically pertains to understanding and reasoning about current news~\citep{fu2022there}, emerging events~\citep{10102595}, cultural phenomena~\citep{nayak2024benchmarking,romero2024cvqa}, and temporally relevant information that constantly evolves~\citep{du2025mmkebench,huang2024vlkeb}. Such live knowledge allows multimodal assistants to provide timely, relevant, and contextually appropriate responses to visual queries about ongoing situations and events.

\header{Synthetic data for knowledge update.} 
Knowledge Updating, also known as continual learning, focuses on rapidly injecting the latest knowledge into pretrained models so they immediately recognize emerging concepts while retaining prior competencies\citep{de2021editing,zhang2023large,huang2024vlkeb,chen2024lifelong,jovanovic2024towards, he2025knowledge}. A major challenge in this process is catastrophic forgetting, where models lose previously acquired knowledge when learning new information, necessitating carefully-constructed high-quality data and specialized updating techniques \citep{luo2023empirical,huang2024mitigating,feng2024kif}. To address these challenges, synthetic datasets have emerged as a critical solution for continuous knowledge infusion without extensive retraining \citep{thede2025understanding,abdin2024phi}. For textual knowledge, frameworks like SynthLLM generate diverse, high-quality synthetic datasets by transforming existing corpora \citep{qin2025scaling}, while techniques such as Knowledge Direct Preference Optimization (KDPO) leverage synthetic examples for targeted factual updates \citep{rozner2024knowledge}. Previous research in language and code domain has successfully build up automatic synthetic framework for code api knowledge synchronizing \citep{liu2024codeupdatearena,wang2025codesync,kumar2024live}. Our work introduces the first fully synthetic automatic engine for visual entity knowledge updating, which automatically collects new visual knowledge from online sources and leverages MLLMs to synthesize high-quality multimodal datasets. Through comprehensive experiments, we investigate the effectiveness of various fine-tuning methods, dataset formulations, and data quantity requirements, providing valuable insights for lifelong world model that must continuously incorporate emerging visual entities while preserving existing capabilities \citep{kim2023carpe}.

%% file: sec/100-checklist.tex
\section*{NeurIPS Paper Checklist}

%%% BEGIN INSTRUCTIONS %%%
The checklist is designed to encourage best practices for responsible machine learning research, addressing issues of reproducibility, transparency, research ethics, and societal impact. Do not remove the checklist: {\bf The papers not including the checklist will be desk rejected.} The checklist should follow the references and follow the (optional) supplemental material.  The checklist does NOT count towards the page
limit. 

Please read the checklist guidelines carefully for information on how to answer these questions. For each question in the checklist:
\begin{itemize}
    \item You should answer \answerYes{}, \answerNo{}, or \answerNA{}.
    \item \answerNA{} means either that the question is Not Applicable for that particular paper or the relevant information is Not Available.
    \item Please provide a short (1–2 sentence) justification right after your answer (even for NA). 
   % \item {\bf The papers not including the checklist will be desk rejected.}
\end{itemize}

{\bf The checklist answers are an integral part of your paper submission.} They are visible to the reviewers, area chairs, senior area chairs, and ethics reviewers. You will be asked to also include it (after eventual revisions) with the final version of your paper, and its final version will be published with the paper.

The reviewers of your paper will be asked to use the checklist as one of the factors in their evaluation. While "\answerYes{}" is generally preferable to "\answerNo{}", it is perfectly acceptable to answer "\answerNo{}" provided a proper justification is given (e.g., "error bars are not reported because it would be too computationally expensive" or "we were unable to find the license for the dataset we used"). In general, answering "\answerNo{}" or "\answerNA{}" is not grounds for rejection. While the questions are phrased in a binary way, we acknowledge that the true answer is often more nuanced, so please just use your best judgment and write a justification to elaborate. All supporting evidence can appear either in the main paper or the supplemental material, provided in appendix. If you answer \answerYes{} to a question, in the justification please point to the section(s) where related material for the question can be found.

IMPORTANT, please:
\begin{itemize}
    \item {\bf Delete this instruction block, but keep the section heading ``NeurIPS paper checklist"},
    \item  {\bf Keep the checklist subsection headings, questions/answers and guidelines below.}
    \item {\bf Do not modify the questions and only use the provided macros for your answers}.
\end{itemize}

%%% END INSTRUCTIONS %%%

\begin{enumerate}

\item {\bf Claims}
    \item[] Question: Do the main claims made in the abstract and introduction accurately reflect the paper's contributions and scope?
    \item[] Answer: \answerYes{} % Replace by \answerYes{}, \answerNo{}, or \answerNA{}.
    \item[] Justification: We did it in the abstract and instruction.

\item {\bf Limitations}
    \item[] Question: Does the paper discuss the limitations of the work performed by the authors?
    \item[] Answer: \answerYes{} % Replace by \answerYes{}, \answerNo{}, or \answerNA{}.
    \item[] Justification: We discuss the limitation and future work in the draft.

\item {\bf Theory Assumptions and Proofs}
    \item[] Question: For each theoretical result, does the paper provide the full set of assumptions and a complete (and correct) proof?
    \item[] Answer: \answerNA{} % Replace by \answerYes{}, \answerNo{}, or \answerNA{}.
    \item[] Justification: For necessary theory assumptions, we provide the proofs in the appendix.

    \item {\bf Experimental Result Reproducibility}
    \item[] Question: Does the paper fully disclose all the information needed to reproduce the main experimental results of the paper to the extent that it affects the main claims and/or conclusions of the paper (regardless of whether the code and data are provided or not)?
    \item[] Answer: \answerYes{} % Replace by \answerYes{}, \answerNo{}, or \answerNA{}.
    \item[] Justification: Yes, we fully disclose the details of experiments of different tasks. We also provide the source code for reproducibility.

\item {\bf Open access to data and code}
    \item[] Question: Does the paper provide open access to the data and code, with sufficient instructions to faithfully reproduce the main experimental results, as described in supplemental material?
    \item[] Answer: \answerYes{} % Replace by \answerYes{}, \answerNo{}, or \answerNA{}.
    \item[] Justification: Yes, we provide the source code and datasets in the supplementary materials.

\item {\bf Experimental Setting/Details}
    \item[] Question: Does the paper specify all the training and test details (e.g., data splits, hyperparameters, how they were chosen, type of optimizer, etc.) necessary to understand the results?
    \item[] Answer: \answerYes{} % Replace by \answerYes{}, \answerNo{}, or \answerNA{}.
    \item[] Justification: Yes, we disclose all related details in the draft.

\item {\bf Experiment Statistical Significance}
    \item[] Question: Does the paper report error bars suitably and correctly defined or other appropriate information about the statistical significance of the experiments?
    \item[] Answer: \answerNo{} % Replace by \answerYes{}, \answerNo{}, or \answerNA{}.
    \item[] Justification: Error bars are not reported because it would be too computationally expensive (especially for some advanced reasoning models).

\item {\bf Experiments Compute Resources}
    \item[] Question: For each experiment, does the paper provide sufficient information on the computer resources (type of compute workers, memory, time of execution) needed to reproduce the experiments?
    \item[] Answer: \answerYes{} % Replace by \answerYes{}, \answerNo{}, or \answerNA{}.
    \item[] Justification: All experiments are running by API or 8 $\times$ NVIDIA A800 (80G).

    
\item {\bf Code Of Ethics}
    \item[] Question: Does the research conducted in the paper conform, in every respect, with the NeurIPS Code of Ethics \url{https://neurips.cc/public/EthicsGuidelines}?
    \item[] Answer: \answerYes{} % Replace by \answerYes{}, \answerNo{}, or \answerNA{}.
    \item[] Justification: Yes, we follow the code of ethics.

\item {\bf Broader Impacts}
    \item[] Question: Does the paper discuss both potential positive societal impacts and negative societal impacts of the work performed?
    \item[] Answer: \answerYes{} % Replace by \answerYes{}, \answerNo{}, or \answerNA{}.
    \item[] Justification: We include some basic discussion of its broader impact in the introduction.
    
\item {\bf Safeguards}
    \item[] Question: Does the paper describe safeguards that have been put in place for responsible release of data or models that have a high risk for misuse (e.g., pretrained language models, image generators, or scraped datasets)?
    \item[] Answer: \answerYes{} % Replace by \answerYes{}, \answerNo{}, or \answerNA{}.
    \item[] Justification: All our data undergoes carefully filtered by advanced models and human. We release our dataset under \texttt{CC-4.0-NC-BY} license.

\item {\bf Licenses for existing assets}
    \item[] Question: Are the creators or original owners of assets (e.g., code, data, models), used in the paper, properly credited and are the license and terms of use explicitly mentioned and properly respected?
    \item[] Answer: \answerYes{} % Replace by \answerYes{}, \answerNo{}, or \answerNA{}.
    \item[] Justification: We use all assets correctly based on their licenses.

\item {\bf New Assets}
    \item[] Question: Are new assets introduced in the paper well documented and is the documentation provided alongside the assets?
    \item[] Answer: \answerYes{} % Replace by \answerYes{}, \answerNo{}, or \answerNA{}.
    \item[] Justification: We introduce the new assets and the documentation is provided in the supplementary matrials.

\item {\bf Crowdsourcing and Research with Human Subjects}
    \item[] Question: For crowdsourcing experiments and research with human subjects, does the paper include the full text of instructions given to participants and screenshots, if applicable, as well as details about compensation (if any)? 
    \item[] Answer: \answerYes{} % Replace by \answerYes{}, \answerNo{}, or \answerNA{}.
    \item[] Justification: Yes, we provide screenshots and human agreement rates (or pass rate) for every human evaluation.

\item {\bf Institutional Review Board (IRB) Approvals or Equivalent for Research with Human Subjects}
    \item[] Question: Does the paper describe potential risks incurred by study participants, whether such risks were disclosed to the subjects, and whether Institutional Review Board (IRB) approvals (or an equivalent approval/review based on the requirements of your country or institution) were obtained?
    \item[] Answer: \answerNA{} % Replace by \answerYes{}, \answerNo{}, or \answerNA{}.
    \item[] Justification: Our research does not require the IRB.

\end{enumerate}

%% file: main.bbl
\begin{thebibliography}{66}
\providecommand{\natexlab}[1]{#1}
\providecommand{\url}[1]{\texttt{#1}}
\expandafter\ifx\csname urlstyle\endcsname\relax
  \providecommand{\doi}[1]{doi: #1}\else
  \providecommand{\doi}{doi: \begingroup \urlstyle{rm}\Url}\fi

\bibitem[Decker et~al.(2023)Decker, Gross, Koebler, Lebacher, Schnitzer, and Weber]{decker2023thousand}
Thomas Decker, Ralf Gross, Alexander Koebler, Michael Lebacher, Ronald Schnitzer, and Stefan~H Weber.
\newblock The thousand faces of explainable ai along the machine learning life cycle: industrial reality and current state of research.
\newblock In \emph{International Conference on Human-Computer Interaction}, pages 184--208. Springer, 2023.

\bibitem[Li et~al.(2024{\natexlab{a}})Li, Jia, Wang, and Yan]{li2024think2drive}
Qifeng Li, Xiaosong Jia, Shaobo Wang, and Junchi Yan.
\newblock Think2drive: Efficient reinforcement learning by thinking with latent world model for autonomous driving (in carla-v2).
\newblock In \emph{European Conference on Computer Vision}, pages 142--158. Springer, 2024{\natexlab{a}}.

\bibitem[OpenAI(2025)]{openai2025gpt41}
OpenAI.
\newblock Gpt-4.1.
\newblock \url{https://openai.com/index/gpt-4-1/}, 2025.
\newblock Large language model. Accessed: 2025-05-11.

\bibitem[Citron(2025)]{citron2025deepresearch}
Dave Citron.
\newblock Deep research is now available on gemini 2.5 pro experimental, April 2025.
\newblock URL \url{https://blog.google/products/gemini/deep-research-gemini-2-5-pro-experimental/}.
\newblock Google Blog, accessed May 11, 2025.

\bibitem[{Anthropic}(2025)]{anthropic2025claude37}
{Anthropic}.
\newblock Introducing claude 3.7 sonnet and claude code, February 2025.
\newblock URL \url{https://www.anthropic.com/news/claude-3-7-sonnet-and-claude-code}.
\newblock Accessed: 2025-05-11.

\bibitem[Antol et~al.(2015)Antol, Agrawal, Lu, Mitchell, Batra, Zitnick, and Parikh]{antol2015vqa}
Stanislaw Antol, Aishwarya Agrawal, Jiasen Lu, Margaret Mitchell, Dhruv Batra, C~Lawrence Zitnick, and Devi Parikh.
\newblock Vqa: Visual question answering.
\newblock In \emph{Proceedings of the IEEE international conference on computer vision}, pages 2425--2433, 2015.

\bibitem[Hudson and Manning(2019)]{hudson2019gqa}
Drew~A Hudson and Christopher~D Manning.
\newblock Gqa: A new dataset for real-world visual reasoning and compositional question answering.
\newblock In \emph{Proceedings of the IEEE/CVF conference on computer vision and pattern recognition}, pages 6700--6709, 2019.

\bibitem[Nguyen et~al.(2025)Nguyen, Nguyen, Nguyen, Luong, Dang, and Lai]{nguyen2025owlviz}
Thuy Nguyen, Dang Nguyen, Hoang Nguyen, Thuan Luong, Long~Hoang Dang, and Viet~Dac Lai.
\newblock Owlviz: An open-world benchmark for visual question answering.
\newblock \emph{arXiv preprint arXiv:2503.07631}, 2025.

\bibitem[Ishfaq et~al.(2023)Ishfaq, Lan, Xu, Mahmood, Precup, Anandkumar, and Azizzadenesheli]{ishfaq2023provable}
Haque Ishfaq, Qingfeng Lan, Pan Xu, A~Rupam Mahmood, Doina Precup, Anima Anandkumar, and Kamyar Azizzadenesheli.
\newblock Provable and practical: Efficient exploration in reinforcement learning via langevin monte carlo.
\newblock \emph{arXiv preprint arXiv:2305.18246}, 2023.

\bibitem[Shabtay et~al.(2024)Shabtay, Polo, Doveh, Lin, Mirza, Chosen, Yurochkin, Sun, Arbelle, Karlinsky, et~al.]{shabtay2024livexiv}
Nimrod Shabtay, Felipe~Maia Polo, Sivan Doveh, Wei Lin, M~Jehanzeb Mirza, Leshem Chosen, Mikhail Yurochkin, Yuekai Sun, Assaf Arbelle, Leonid Karlinsky, et~al.
\newblock Livexiv--a multi-modal live benchmark based on arxiv papers content.
\newblock \emph{arXiv preprint arXiv:2410.10783}, 2024.

\bibitem[Li et~al.(2024{\natexlab{b}})Li, Wang, Xu, Wang, Feng, Kong, and Liu]{li2024multimodal}
Lei Li, Yuqi Wang, Runxin Xu, Peiyi Wang, Xiachong Feng, Lingpeng Kong, and Qi~Liu.
\newblock Multimodal arxiv: A dataset for improving scientific comprehension of large vision-language models.
\newblock \emph{arXiv preprint arXiv:2403.00231}, 2024{\natexlab{b}}.

\bibitem[Jiang et~al.(2024)Jiang, Zhang, Guo, Wu, Lei, Qiu, Lu, Chen, Fu, Song, et~al.]{jiang2024mmsearch}
Dongzhi Jiang, Renrui Zhang, Ziyu Guo, Yanmin Wu, Jiayi Lei, Pengshuo Qiu, Pan Lu, Zehui Chen, Chaoyou Fu, Guanglu Song, et~al.
\newblock Mmsearch: Benchmarking the potential of large models as multi-modal search engines.
\newblock \emph{arXiv preprint arXiv:2409.12959}, 2024.

\bibitem[Wei et~al.(2022)Wei, Wang, Schuurmans, Bosma, Xia, Chi, Le, Zhou, et~al.]{wei2022chain}
Jason Wei, Xuezhi Wang, Dale Schuurmans, Maarten Bosma, Fei Xia, Ed~Chi, Quoc~V Le, Denny Zhou, et~al.
\newblock Chain-of-thought prompting elicits reasoning in large language models.
\newblock \emph{Advances in neural information processing systems}, 35:\penalty0 24824--24837, 2022.

\bibitem[Wei et~al.(2024)Wei, Karina, Chung, Jiao, Papay, Glaese, Schulman, and Fedus]{wei2024measuring}
Jason Wei, Nguyen Karina, Hyung~Won Chung, Yunxin~Joy Jiao, Spencer Papay, Amelia Glaese, John Schulman, and William Fedus.
\newblock Measuring short-form factuality in large language models.
\newblock \emph{arXiv preprint arXiv:2411.04368}, 2024.

\bibitem[Google(2025)]{google2025gemini2.5pro}
Google.
\newblock Gemini 2.5 pro.
\newblock \url{https://blog.google/technology/google-deepmind/gemini-model-thinking-updates-march-2025/}, 2025.

\bibitem[Yang et~al.(2024)Yang, Yang, Zhang, Hui, Zheng, Yu, Li, Liu, Huang, Wei, et~al.]{yang2024qwen2}
An~Yang, Baosong Yang, Beichen Zhang, Binyuan Hui, Bo~Zheng, Bowen Yu, Chengyuan Li, Dayiheng Liu, Fei Huang, Haoran Wei, et~al.
\newblock Qwen2. 5 technical report.
\newblock \emph{arXiv preprint arXiv:2412.15115}, 2024.

\bibitem[Team et~al.(2025)Team, Kamath, Ferret, Pathak, Vieillard, Merhej, Perrin, Matejovicova, Ram{\'e}, Rivi{\`e}re, et~al.]{team2025gemma}
Gemma Team, Aishwarya Kamath, Johan Ferret, Shreya Pathak, Nino Vieillard, Ramona Merhej, Sarah Perrin, Tatiana Matejovicova, Alexandre Ram{\'e}, Morgane Rivi{\`e}re, et~al.
\newblock Gemma 3 technical report.
\newblock \emph{arXiv preprint arXiv:2503.19786}, 2025.

\bibitem[{Meta}(2025)]{2025llama4}
{Meta}.
\newblock The llama 4 herd: The beginning of a new era of natively multimodal ai innovation, 2025.
\newblock URL \url{https://ai.meta.com/blog/llama-4-multimodal-intelligence/}.

\bibitem[{OpenAI}(2025)]{openai2025gpto3}
{OpenAI}.
\newblock Introducing openai o3 and o4-mini, 2025.
\newblock URL \url{https://openai.com/index/introducing-o3-and-o4-mini/}.

\bibitem[Phan et~al.(2025)Phan, Gatti, Han, Li, Hu, Zhang, Zhang, Shaaban, Ling, Shi, et~al.]{phan2025humanity}
Long Phan, Alice Gatti, Ziwen Han, Nathaniel Li, Josephina Hu, Hugh Zhang, Chen Bo~Calvin Zhang, Mohamed Shaaban, John Ling, Sean Shi, et~al.
\newblock Humanity's last exam.
\newblock \emph{arXiv preprint arXiv:2501.14249}, 2025.

\bibitem[Panickssery et~al.(2024)Panickssery, Bowman, and Feng]{panickssery2024llm}
Arjun Panickssery, Samuel Bowman, and Shi Feng.
\newblock Llm evaluators recognize and favor their own generations.
\newblock \emph{Advances in Neural Information Processing Systems}, 37:\penalty0 68772--68802, 2024.

\bibitem[Gao et~al.(2024)Gao, Zhang, Chen, Huang, Wu, Fu, Wan, Zhang, and Sun]{gao2024best}
Chujie Gao, Qihui Zhang, Dongping Chen, Yue Huang, Siyuan Wu, Zhengyan Fu, Yao Wan, Xiangliang Zhang, and Lichao Sun.
\newblock The best of both worlds: Toward an honest and helpful large language model.
\newblock \emph{arXiv preprint arXiv:2406.00380}, 2024.

\bibitem[Ravaut et~al.(2024)Ravaut, Ding, Jiao, Chen, Li, Zhao, Qin, Xiong, and Joty]{ravaut2024much}
Mathieu Ravaut, Bosheng Ding, Fangkai Jiao, Hailin Chen, Xingxuan Li, Ruochen Zhao, Chengwei Qin, Caiming Xiong, and Shafiq Joty.
\newblock How much are large language models contaminated? a comprehensive survey and the llmsanitize library.
\newblock \emph{arXiv preprint arXiv:2404.00699}, 2024.

\bibitem[Zeng et~al.(2024)Zeng, Gu, Yang, Duan, Shi, and Wang]{zeng2024visual}
Zhen Zeng, Leijiang Gu, Xun Yang, Zhangling Duan, Zenglin Shi, and Meng Wang.
\newblock Visual-oriented fine-grained knowledge editing for multimodal large language models.
\newblock \emph{arXiv preprint arXiv:2411.12790}, 2024.

\bibitem[Chen et~al.(2024{\natexlab{a}})Chen, Wang, Wang, Zhang, Li, and He]{chen2024lifelong}
Qizhou Chen, Chengyu Wang, Dakan Wang, Taolin Zhang, Wangyue Li, and Xiaofeng He.
\newblock Lifelong knowledge editing for vision language models with low-rank mixture-of-experts.
\newblock \emph{arXiv preprint arXiv:2411.15432}, 2024{\natexlab{a}}.

\bibitem[Hu et~al.(2022)Hu, Shen, Wallis, Allen-Zhu, Li, Wang, Wang, Chen, et~al.]{hu2022lora}
Edward~J Hu, Yelong Shen, Phillip Wallis, Zeyuan Allen-Zhu, Yuanzhi Li, Shean Wang, Lu~Wang, Weizhu Chen, et~al.
\newblock Lora: Low-rank adaptation of large language models.
\newblock \emph{ICLR}, 1\penalty0 (2):\penalty0 3, 2022.

\bibitem[Liu et~al.(2024{\natexlab{a}})Liu, Wang, Yin, Molchanov, Wang, Cheng, and Chen]{liu2024dora}
Shih-Yang Liu, Chien-Yi Wang, Hongxu Yin, Pavlo Molchanov, Yu-Chiang~Frank Wang, Kwang-Ting Cheng, and Min-Hung Chen.
\newblock Dora: Weight-decomposed low-rank adaptation.
\newblock In \emph{Forty-first International Conference on Machine Learning}, 2024{\natexlab{a}}.

\bibitem[Wang et~al.(2025)Wang, Chu, Cheng, Yang, Qiu, Wan, Zhao, Shi, and Chen]{wang2025codesync}
Chenlong Wang, Zhaoyang Chu, Zhengxiang Cheng, Xuyi Yang, Kaiyue Qiu, Yao Wan, Zhou Zhao, Xuanhua Shi, and Dongping Chen.
\newblock Codesync: Synchronizing large language models with dynamic code evolution at scale.
\newblock \emph{arXiv preprint arXiv:2502.16645}, 2025.

\bibitem[Yue et~al.(2024)Yue, Ni, Zhang, Zheng, Liu, Zhang, Stevens, Jiang, Ren, Sun, et~al.]{yue2024mmmu}
Xiang Yue, Yuansheng Ni, Kai Zhang, Tianyu Zheng, Ruoqi Liu, Ge~Zhang, Samuel Stevens, Dongfu Jiang, Weiming Ren, Yuxuan Sun, et~al.
\newblock Mmmu: A massive multi-discipline multimodal understanding and reasoning benchmark for expert agi.
\newblock In \emph{Proceedings of the IEEE/CVF Conference on Computer Vision and Pattern Recognition}, pages 9556--9567, 2024.

\bibitem[Chen et~al.(2024{\natexlab{b}})Chen, Li, Dong, Zhang, Zang, Chen, Duan, Wang, Qiao, Lin, et~al.]{chen2024we}
Lin Chen, Jinsong Li, Xiaoyi Dong, Pan Zhang, Yuhang Zang, Zehui Chen, Haodong Duan, Jiaqi Wang, Yu~Qiao, Dahua Lin, et~al.
\newblock Are we on the right way for evaluating large vision-language models?
\newblock \emph{arXiv preprint arXiv:2403.20330}, 2024{\natexlab{b}}.

\bibitem[Team et~al.(2023)Team, Anil, Borgeaud, Alayrac, Yu, Soricut, Schalkwyk, Dai, Hauth, Millican, et~al.]{team2023gemini}
Gemini Team, Rohan Anil, Sebastian Borgeaud, Jean-Baptiste Alayrac, Jiahui Yu, Radu Soricut, Johan Schalkwyk, Andrew~M Dai, Anja Hauth, Katie Millican, et~al.
\newblock Gemini: a family of highly capable multimodal models.
\newblock \emph{arXiv preprint arXiv:2312.11805}, 2023.

\bibitem[{Jinming Wu}(2025)]{wu2025mmsearchr1}
{Jinming Wu}.
\newblock Multimodal-search-r1: Incentivizing lmms to search, 2025.
\newblock URL \url{https://github.com/EvolvingLMMs-Lab/multimodal-search-r1}.

\bibitem[Marino et~al.(2019)Marino, Rastegari, Farhadi, and Mottaghi]{marino2019ok}
Kenneth Marino, Mohammad Rastegari, Ali Farhadi, and Roozbeh Mottaghi.
\newblock Ok-vqa: A visual question answering benchmark requiring external knowledge.
\newblock In \emph{Proceedings of the IEEE/cvf conference on computer vision and pattern recognition}, pages 3195--3204, 2019.

\bibitem[Schwenk et~al.(2022)Schwenk, Khandelwal, Clark, Marino, and Mottaghi]{schwenk2022okvqa}
Dustin Schwenk, Apoorv Khandelwal, Christopher Clark, Kenneth Marino, and Roozbeh Mottaghi.
\newblock A-okvqa: A benchmark for visual question answering using world knowledge.
\newblock In \emph{European conference on computer vision}, pages 146--162. Springer, 2022.

\bibitem[Cheng et~al.(2025)Cheng, Zhang, Zhang, Yang, Guan, Wu, Li, Zhang, Liu, Mai, et~al.]{cheng2025simplevqa}
Xianfu Cheng, Wei Zhang, Shiwei Zhang, Jian Yang, Xiangyuan Guan, Xianjie Wu, Xiang Li, Ge~Zhang, Jiaheng Liu, Yuying Mai, et~al.
\newblock Simplevqa: Multimodal factuality evaluation for multimodal large language models.
\newblock \emph{arXiv preprint arXiv:2502.13059}, 2025.

\bibitem[Xie et~al.(2019)Xie, Lai, Doran, and Kadav]{xie2019visual}
Ning Xie, Farley Lai, Derek Doran, and Asim Kadav.
\newblock Visual entailment: A novel task for fine-grained image understanding.
\newblock \emph{arXiv preprint arXiv:1901.06706}, 2019.

\bibitem[Wang et~al.(2015)Wang, Wu, Shen, Hengel, and Dick]{wang2015explicit}
Peng Wang, Qi~Wu, Chunhua Shen, Anton van~den Hengel, and Anthony Dick.
\newblock Explicit knowledge-based reasoning for visual question answering.
\newblock \emph{arXiv preprint arXiv:1511.02570}, 2015.

\bibitem[Wang et~al.(2017)Wang, Wu, Shen, Dick, and Van Den~Hengel]{wang2017fvqa}
Peng Wang, Qi~Wu, Chunhua Shen, Anthony Dick, and Anton Van Den~Hengel.
\newblock Fvqa: Fact-based visual question answering.
\newblock \emph{IEEE transactions on pattern analysis and machine intelligence}, 40\penalty0 (10):\penalty0 2413--2427, 2017.

\bibitem[Jain et~al.(2021)Jain, Kothyari, Kumar, Jyothi, Ramakrishnan, and Chakrabarti]{jain2021select}
Aman Jain, Mayank Kothyari, Vishwajeet Kumar, Preethi Jyothi, Ganesh Ramakrishnan, and Soumen Chakrabarti.
\newblock Select, substitute, search: A new benchmark for knowledge-augmented visual question answering.
\newblock In \emph{Proceedings of the 44th International ACM SIGIR Conference on Research and Development in Information Retrieval}, pages 2491--2498, 2021.

\bibitem[Shah et~al.(2019)Shah, Mishra, Yadati, and Talukdar]{shah2019kvqa}
Sanket Shah, Anand Mishra, Naganand Yadati, and Partha~Pratim Talukdar.
\newblock Kvqa: Knowledge-aware visual question answering.
\newblock In \emph{Proceedings of the AAAI conference on artificial intelligence}, volume~33, pages 8876--8884, 2019.

\bibitem[Chen et~al.(2022)Chen, Hu, Chen, Verga, and Cohen]{chen2022murag}
Wenhu Chen, Hexiang Hu, Xi~Chen, Pat Verga, and William~W Cohen.
\newblock Murag: Multimodal retrieval-augmented generator for open question answering over images and text.
\newblock \emph{arXiv preprint arXiv:2210.02928}, 2022.

\bibitem[Zhao et~al.(2023)Zhao, Chen, Wang, Jiao, Do, Qin, Ding, Guo, Li, Li, et~al.]{zhao2023retrieving}
Ruochen Zhao, Hailin Chen, Weishi Wang, Fangkai Jiao, Xuan~Long Do, Chengwei Qin, Bosheng Ding, Xiaobao Guo, Minzhi Li, Xingxuan Li, et~al.
\newblock Retrieving multimodal information for augmented generation: A survey.
\newblock \emph{arXiv preprint arXiv:2303.10868}, 2023.

\bibitem[Caffagni et~al.(2024)Caffagni, Cocchi, Moratelli, Sarto, Cornia, Baraldi, and Cucchiara]{caffagni2024wiki}
Davide Caffagni, Federico Cocchi, Nicholas Moratelli, Sara Sarto, Marcella Cornia, Lorenzo Baraldi, and Rita Cucchiara.
\newblock Wiki-llava: Hierarchical retrieval-augmented generation for multimodal llms.
\newblock In \emph{Proceedings of the IEEE/CVF Conference on Computer Vision and Pattern Recognition}, pages 1818--1826, 2024.

\bibitem[Yan and Xie(2024)]{yan2024echosight}
Yibin Yan and Weidi Xie.
\newblock Echosight: Advancing visual-language models with wiki knowledge.
\newblock \emph{arXiv preprint arXiv:2407.12735}, 2024.

\bibitem[Abootorabi et~al.(2025)Abootorabi, Zobeiri, Dehghani, Mohammadkhani, Mohammadi, Ghahroodi, Baghshah, and Asgari]{abootorabi2025ask}
Mohammad~Mahdi Abootorabi, Amirhosein Zobeiri, Mahdi Dehghani, Mohammadali Mohammadkhani, Bardia Mohammadi, Omid Ghahroodi, Mahdieh~Soleymani Baghshah, and Ehsaneddin Asgari.
\newblock Ask in any modality: A comprehensive survey on multimodal retrieval-augmented generation.
\newblock \emph{arXiv preprint arXiv:2502.08826}, 2025.

\bibitem[Fu et~al.(2022)Fu, Zhou, Chandratreya, Vondrick, and Roth]{fu2022there}
Xingyu Fu, Ben Zhou, Ishaan~Preetam Chandratreya, Carl Vondrick, and Dan Roth.
\newblock There is a time and place for reasoning beyond the image.
\newblock \emph{arXiv preprint arXiv:2203.00758}, 2022.

\bibitem[Yang et~al.(2023)Yang, Xiang, You, Li, and Liu]{10102595}
Zhenguo Yang, Jiale Xiang, Jiuxiang You, Qing Li, and Wenyin Liu.
\newblock Event-oriented visual question answering: The e-vqa dataset and benchmark.
\newblock \emph{IEEE Transactions on Knowledge and Data Engineering}, 35\penalty0 (10):\penalty0 10210--10223, 2023.
\newblock \doi{10.1109/TKDE.2023.3267036}.

\bibitem[Nayak et~al.(2024)Nayak, Jain, Awal, Reddy, Van~Steenkiste, Hendricks, Agrawal, et~al.]{nayak2024benchmarking}
Shravan Nayak, Kanishk Jain, Rabiul Awal, Siva Reddy, Sjoerd Van~Steenkiste, Lisa~Anne Hendricks, Aishwarya Agrawal, et~al.
\newblock Benchmarking vision language models for cultural understanding.
\newblock \emph{arXiv preprint arXiv:2407.10920}, 2024.

\bibitem[Romero et~al.(2024)Romero, Lyu, Wibowo, Lynn, Hamed, Kishore, Mandal, Dragonetti, Abzaliev, Tonja, et~al.]{romero2024cvqa}
David Romero, Chenyang Lyu, Haryo~Akbarianto Wibowo, Teresa Lynn, Injy Hamed, Aditya~Nanda Kishore, Aishik Mandal, Alina Dragonetti, Artem Abzaliev, Atnafu~Lambebo Tonja, et~al.
\newblock Cvqa: Culturally-diverse multilingual visual question answering benchmark.
\newblock \emph{arXiv preprint arXiv:2406.05967}, 2024.

\bibitem[Du. et~al.(2025)Du., Jiang, Gao, Shi, Zheng, Qi, and Li]{du2025mmkebench}
Yuntao Du., Kailin Jiang, Zhi Gao, Chenrui Shi, Zilong Zheng, Siyuan Qi, and Qing Li.
\newblock {MMKE}-bench: A multimodal editing benchmark for diverse visual knowledge.
\newblock In \emph{The Thirteenth International Conference on Learning Representations}, 2025.
\newblock URL \url{https://openreview.net/forum?id=v8qABSeeKO}.

\bibitem[Huang et~al.(2024{\natexlab{a}})Huang, Zhong, Yu, Liu, Wu, Wang, and Tan]{huang2024vlkeb}
Han Huang, Haitian Zhong, Tao Yu, Qiang Liu, Shu Wu, Liang Wang, and Tieniu Tan.
\newblock Vlkeb: A large vision-language model knowledge editing benchmark.
\newblock \emph{arXiv preprint arXiv:2403.07350}, 2024{\natexlab{a}}.

\bibitem[De~Cao et~al.(2021)De~Cao, Aziz, and Titov]{de2021editing}
Nicola De~Cao, Wilker Aziz, and Ivan Titov.
\newblock Editing factual knowledge in language models.
\newblock \emph{arXiv preprint arXiv:2104.08164}, 2021.

\bibitem[Zhang et~al.(2023)Zhang, Fang, Chen, Namazi-Rad, and Wang]{zhang2023large}
Zihan Zhang, Meng Fang, Ling Chen, Mohammad-Reza Namazi-Rad, and Jun Wang.
\newblock How do large language models capture the ever-changing world knowledge? a review of recent advances.
\newblock \emph{arXiv preprint arXiv:2310.07343}, 2023.

\bibitem[Jovanovic and Voss(2024)]{jovanovic2024towards}
Mladjan Jovanovic and Peter Voss.
\newblock Towards incremental learning in large language models: A critical review.
\newblock \emph{arXiv preprint arXiv:2404.18311}, 2024.

\bibitem[He et~al.(2025)He, Song, and Sun]{he2025knowledge}
Guoxiu He, Xin Song, and Aixin Sun.
\newblock Knowledge updating? no more model editing! just selective contextual reasoning.
\newblock \emph{arXiv preprint arXiv:2503.05212}, 2025.

\bibitem[Luo et~al.(2023)Luo, Yang, Meng, Li, Zhou, and Zhang]{luo2023empirical}
Yun Luo, Zhen Yang, Fandong Meng, Yafu Li, Jie Zhou, and Yue Zhang.
\newblock An empirical study of catastrophic forgetting in large language models during continual fine-tuning.
\newblock \emph{arXiv preprint arXiv:2308.08747}, 2023.

\bibitem[Huang et~al.(2024{\natexlab{b}})Huang, Cui, Wang, Yang, Liao, Song, Yao, and Su]{huang2024mitigating}
Jianheng Huang, Leyang Cui, Ante Wang, Chengyi Yang, Xinting Liao, Linfeng Song, Junfeng Yao, and Jinsong Su.
\newblock Mitigating catastrophic forgetting in large language models with self-synthesized rehearsal.
\newblock \emph{arXiv preprint arXiv:2403.01244}, 2024{\natexlab{b}}.

\bibitem[Feng et~al.(2024)Feng, Chu, Xu, Lu, Liu, Yu, and Wu]{feng2024kif}
Yujie Feng, Xu~Chu, Yongxin Xu, Zexin Lu, Bo~Liu, Philip~S Yu, and Xiao-Ming Wu.
\newblock Kif: Knowledge identification and fusion for language model continual learning.
\newblock \emph{arXiv preprint arXiv:2408.05200}, 2024.

\bibitem[Thede et~al.(2025)Thede, Roth, Bethge, Akata, and Hartvigsen]{thede2025understanding}
Lukas Thede, Karsten Roth, Matthias Bethge, Zeynep Akata, and Tom Hartvigsen.
\newblock Understanding the limits of lifelong knowledge editing in llms.
\newblock \emph{arXiv preprint arXiv:2503.05683}, 2025.

\bibitem[Abdin et~al.(2024)Abdin, Aneja, Behl, Bubeck, Eldan, Gunasekar, Harrison, Hewett, Javaheripi, Kauffmann, et~al.]{abdin2024phi}
Marah Abdin, Jyoti Aneja, Harkirat Behl, S{\'e}bastien Bubeck, Ronen Eldan, Suriya Gunasekar, Michael Harrison, Russell~J Hewett, Mojan Javaheripi, Piero Kauffmann, et~al.
\newblock Phi-4 technical report.
\newblock \emph{arXiv preprint arXiv:2412.08905}, 2024.

\bibitem[Qin et~al.(2025)Qin, Dong, Zhang, Dong, Huang, Yang, Khademi, Zhang, Awadalla, Fung, et~al.]{qin2025scaling}
Zeyu Qin, Qingxiu Dong, Xingxing Zhang, Li~Dong, Xiaolong Huang, Ziyi Yang, Mahmoud Khademi, Dongdong Zhang, Hany~Hassan Awadalla, Yi~R Fung, et~al.
\newblock Scaling laws of synthetic data for language models.
\newblock \emph{arXiv preprint arXiv:2503.19551}, 2025.

\bibitem[Rozner et~al.(2024)Rozner, Battash, Wolf, and Lindenbaum]{rozner2024knowledge}
Amit Rozner, Barak Battash, Lior Wolf, and Ofir Lindenbaum.
\newblock Knowledge editing in language models via adapted direct preference optimization.
\newblock \emph{arXiv preprint arXiv:2406.09920}, 2024.

\bibitem[Liu et~al.(2024{\natexlab{b}})Liu, Pandit, Ye, Choi, and Durrett]{liu2024codeupdatearena}
Zeyu~Leo Liu, Shrey Pandit, Xi~Ye, Eunsol Choi, and Greg Durrett.
\newblock Codeupdatearena: Benchmarking knowledge editing on api updates.
\newblock \emph{arXiv preprint arXiv:2407.06249}, 2024{\natexlab{b}}.

\bibitem[Kumar and Kaur(2024)]{kumar2024live}
Aditya Kumar and Manjinder Kaur.
\newblock Live code sync.
\newblock In \emph{Advances in AI for Biomedical Instrumentation, Electronics and Computing}, pages 554--559. CRC Press, 2024.

\bibitem[Kim et~al.(2023)Kim, Yoon, Ye, Bae, Ho, Hwang, and Yun]{kim2023carpe}
Yujin Kim, Jaehong Yoon, Seonghyeon Ye, Sangmin Bae, Namgyu Ho, Sung~Ju Hwang, and Se-Young Yun.
\newblock Carpe diem: On the evaluation of world knowledge in lifelong language models.
\newblock \emph{arXiv preprint arXiv:2311.08106}, 2023.

\bibitem[{OpenAI}(2024)]{openai2024gpt4o}
{OpenAI}.
\newblock Gpt-4o, 2024.
\newblock URL \url{https://openai.com/gpt-4o}.
\newblock Accessed: 2025-05-01.

\end{thebibliography}
